\documentclass[journal]{IEEEtran}
\newcommand{\ignore}[1]{}
\usepackage[utf8]{inputenc}
\usepackage[T1]{fontenc}
\usepackage{amsmath,epsfig}
\usepackage{graphicx}
\usepackage{bm}
\usepackage{amssymb}
\usepackage{cite}
\usepackage{array}
\usepackage{extarrows}
\usepackage{url}

\usepackage{color}

\usepackage{multirow}
\usepackage{booktabs}
\usepackage{blkarray}
\usepackage{multirow}
\usepackage{array}
\usepackage{color}
\usepackage{pmat}

\begin{document}
%
\title{Salient Object Detection with Lossless Feature Reflection and Weighted Structural Loss}
\author{Pingping~Zhang,
        Wei~Liu,
        Huchuan~Lu
        and~Chunhua Shen
\thanks{
%
PP. Zhang and HC. Lu are with School of Information and Communication Engineering, Dalian University of Technology, Dalian, 116024, China. (Email: jssxzhpp@mail.dlut.edu.cn; lhchuan@dlut.edu.cn) The corresponding author is Prof. Huchuan Lu.

W. Liu is with Key Laboratory of Ministry of Education for System Control and Information Processing, Shanghai Jiao Tong University, Shanghai, 200240, China.(Email: liuwei.1989@sjtu.edu.cn)

CH. Shen is with School of Computer Science, University of Adelaide, Adelaide, SA 50005, Australia.(Email: chunhua.shen@adelaide.edu.au)

This work is supported in part by the National Natural Science Foundation of China (NNSFC), No. 61725202, No. 61751212 and No. 61771088. This work was done when the first two authors were visiting the University of Adelaide, supported by the China Scholarship Council (CSC) program.
}
}
\maketitle
\begin{abstract}
Salient object detection (SOD), which aims to identify and locate the most salient pixels or regions in images, has been attracting more and more interest due to its various real-world applications.
However, this vision task is quite challenging, especially under complex image scenes.
Inspired by the intrinsic reflection of natural images, in this paper we propose a novel feature learning framework for large-scale salient object detection.
Specifically, we design a symmetrical fully convolutional network (SFCN) to effectively learn complementary saliency features under the guidance of lossless feature reflection.
The location information, together with contextual and semantic information, of salient objects are jointly utilized to supervise the proposed network for more accurate saliency predictions.
In addition, to overcome the blurry boundary problem, we propose a new weighted structural loss function to ensure clear object boundaries and spatially consistent saliency.
The coarse prediction results are effectively refined by these structural information for performance improvements.
Extensive experiments on seven saliency detection datasets demonstrate that our approach achieves consistently superior performance and outperforms the very recent state-of-the-art methods with a large margin.
\end{abstract}
\IEEEpeerreviewmaketitle
\section{Introduction}
\IEEEPARstart{S}{alient} object detection (SOD) is a fundamental yet challenging task in computer vision.
It aims to identify and locate distinctive objects or regions which attract human attention in natural images.
In general, SOD is regarded as a prerequisite step to narrow down subsequent object-related vision tasks, such as sematic segmentation, visual tracking and person re-identification.

In the past two decades, a large number of SOD methods have been proposed.
Most of them have been well summarized in~\cite{borji2015salient}.
According to these works, conventional SOD methods focus on extracting discriminative local and global handcrafted features from pixels or regions to represent their visual properties.
With several heuristic priors such as center surrounding and visual contrast, these methods predict salient scores according to the extracted features for saliency detection.
Although great success has been made, there still exist many important problems which need to be solved.
For example, the low-level handcrafted features suffer from limited representation capability and robustness, and are difficult to capture the semantic and structural information of objects in images, which is very important for more accurate SOD.
What's more, to further extract powerful and robust visual features manually is a tough mission for performance improvement, especially in complex image scenes, such as cluttered backgrounds and low-contrast imaging patterns.

With the recent prevalence of deep architectures, many remarkable progresses have been achieved in a wide range of computer vision tasks, \emph{e.g.}, image classification, object detection and semantic segmentation.
Thus, many researchers start to make their efforts to utilize deep convolutional neural networks (CNNs) for SOD and have achieved outstanding performance, since CNNs have strong ability to automatically extract high-level feature representations, successfully avoiding the drawbacks of handcrafted features.
However, most of state-of-the-art SOD methods still require large-scale pre-trained CNNs, which usually employ the strided convolution and pooling operations.
These downsampling methods largely increase the receptive field of CNNs, helping to extract high-level semantic features, nevertheless they inevitably drop the location information and fine details of objects, leading to unclear boundary predictions.
Furthermore, the lack of structural supervision also makes SOD an extremely challenging problem in complex image scenes.

In order to utilize the semantic and structural information derived from deep pre-trained CNNs, in this paper we propose to solve both tasks of complementary feature extraction and saliency region classification with an unified framework which is learned in the end-to-end manner.
More specifically, we design a novel symmetrical fully convolutional network (SFCN) architecture which consists of two sibling branches and one fusing branch.
The two sibling branches take reciprocal image pairs as inputs and share weights for learning complementary visual features under the guidance of lossless feature reflection.
The fusing branch integrates the multi-level complementary features in a hierarchical manner for SOD.
More importantly, to effectively train our network, we propose a novel weighted loss function which incorporates rich structural information and supervises the three branches during the training process.
In this manner, our proposed model can sufficiently capture the boundaries and spatial contexts of salient objects, hence significantly boosts the performance of SOD.

In summary, \textbf{our contributions} are three folds:
\begin{itemize}
\item
We present a novel deep network architecture, \emph{i.e.}, SFCN, which is symmetrically designed to learn complementary visual features and predict accurate saliency maps under the guidance of lossless feature reflection.
\item
We propose a new structural loss function to learn clear object boundaries and spatially consistent saliency.
This novel loss function is able to utilize the location, contextual and semantic information of salient objects to supervise the proposed SFCN for performance improvements.
\item
Extensive experimental results on seven public large-scale saliency benchmarks demonstrate that the proposed approach achieves superior performance and outperforms the very recent state-of-the-art methods by a large margin.
\end{itemize}

A preliminary version of this work appeared at IJCAI 2018~\cite{zhang2018salient}.
Compared to our previous work, this paper employs an analysis for our reflection method and extends our original adaptive batch normalization (AdaBN) into layer-wise AdaBN, which achieves better performance than original methods.
Secondly, more related works are reviewed, especially for two-stream networks.
We add more comparisons and discussions with respect to recent deep learning based
SOD methods that came after our original paper.
Moreover, more experiments are conducted to show the influence of different configurations.
\section{Related Work}
In this section, we first briefly review existing deep learning based SOD methods.
Then, we discuss the representative works on the intrinsic reflection of natural images and two-stream deep networks, which clarifies our main motivations.
\vspace{-6mm}
{\flushleft\textbf{Salient Object Detection.}} Recent years, deep learning based methods have achieved solid performance improvements in SOD.
For example, Wang \emph{et al.}~\cite{wang2015deep} integrate both local pixel estimation and global proposal search for SOD by training two deep neural networks.
Zhao \emph{et al.}~\cite{zhao2015saliency} propose a multi-context deep CNN framework to benefit from the local context and global context of salient objects.
Li \emph{et al.}~\cite{li2015visual} employ multiple deep CNNs to extract multi-scale features for saliency prediction.
Then they propose a deep contrast network to combine a pixel-level stream and segment-wise stream for saliency estimation~\cite{li2016deep}.
Inspired by the great success of fully convolutional networks (FCNs)~\cite{long2015fully}, Wang \emph{et al.}~\cite{wang2016saliency} develop a recurrent FCN to incorporate saliency priors for more accurate saliency map inference.
Liu \emph{et al.}~\cite{liu2016dhsnet} also design a deep hierarchical network to learn a coarse global estimation and then refine the saliency map hierarchically and progressively.
Then, Hou \emph{et al.}~\cite{hou2017deeply} introduce dense short connections within the holistically-nested edge detection (HED) architecture~\cite{xie2015holistically} to get rich multi-scale features for SOD.
Zhang \emph{et al.}~\cite{zhang2017amulet} propose a bidirectional learning framework to aggregate multi-level convolutional features for SOD.
And they also develop a novel dropout to learn the deep uncertain convolutional features to enhance the robustness and accuracy of saliency detection~\cite{zhang2017learning}.
To enhance the boundary information, Wang \emph{et al.}~\cite{wang2017stagewise} and Zhuge \emph{et al.}~\cite{zhuge2018boundary} provide stage-wise refinement frameworks to gradually get accurate saliency detection results.
Hu \emph{et al.}~\cite{hu2017deep} provide a deep level set framework to refine the saliency detection results.
Luo \emph{et al.}~\cite{luo2017non} integrate non-local deep features into global predictions for boundary refinements.
Besides, Zhu \emph{et al.}~\cite{zhu2017saliency} formulate the saliency pattern detection by ranking structured trees.
Zhang \emph{et al.}~\cite{zhang2018bi} propose a bi-directional message passing model for saliency detection.
Zhang \emph{et al.}~\cite{zhang2018progressive} propose a recurrent network with progressive attentions for regional locations.
To speed up the inference of SOD, Zhang \emph{et al.}~\cite{zhang2018agile} introduce a contextual attention
module that can rapidly highlight most salient objects or regions with contextual pyramids.
Wang \emph{et al.}~\cite{wang2018detect} propose to detect objects globally and refine them locally.
Liu \emph{et al.}~\cite{liu2018picanet} extend their previous work~\cite{liu2016dhsnet} with pixel-wise contextual attention.

Despite aforementioned approaches employ powerful CNNs and make remarkable success in SOD, there still exist some obvious problems.
For example, the strategies of multi-stage training significantly reduces the efficiency for model deployment.
And the explicit pixel-wise loss functions used by these methods for model training cannot well reflect the structural information of salient objects.
Hence, there is still a large space for performance improvements in accurate SOD.
While our model is trained in end-to-end manner and supervised by a structural loss.
Thus, our model does not need any multi-stage training and complex post-processing.
\vspace{-2mm}
{\flushleft\textbf{Image Intrinsic Reflection.}} Image intrinsic reflection is a classical topic in computer vision field~\cite{land1971lightness,klinker1990physical}.
It can be used to segment and analyze surfaces with image color variations due to highlights and shading~\cite{klinker1990physical}.
Most of existing intrinsic reflection methods are based on the Retinex model~\cite{land1971lightness}, which captures image information for Mondrian images: images of a planar canvas that is covered by samll patches of constant reflectance and illuminated by multiple light sources.
Recent years, researchers have augmented the basic Retinex model with non-local texture cues~\cite{zhao2012closed} and image sparsity priors~\cite{shen2011intrinsic}.
Sophisticated techniques that recover reflectance and shading along with a shape estimate have also been proposed~\cite{barron2012color}.
Inspired by these works, we construct a reciprocal image pair based on the input image (see Section III. A).
However, there are three obvious differences between our proposed method and previous intrinsic reflection methods:
1) the task objective is different. The aim of previous methods is explaining an input RGB image by estimating albedo and shading fields. Our aim is to learn complementary visual features for SOD.
2) the resulting image pair is different. Image intrinsic reflection methods usually factory an input RGB image into a reflectance image and a shading image. For every pixel, the reflectance image encodes the albedo of depicted surfaces, while the shading image encodes the incident illumination at corresponding points in the scene.
In our proposed method, a reciprocal image pair, that captures the characteristics of planar reflection, is generated as the input of deep networks.
3) the source of reflection is different. The source of previous intrinsic reflection methods is the albedo of depicted
surfaces, while our reflection is originated from deep features in CNNs. Therefore, our reflection is feature-level not image-level.
In the experimental section, we will show that our method is very useful for SOD.
\begin{figure*}
\begin{center}
\includegraphics[width=0.9\linewidth,height=8.5cm]{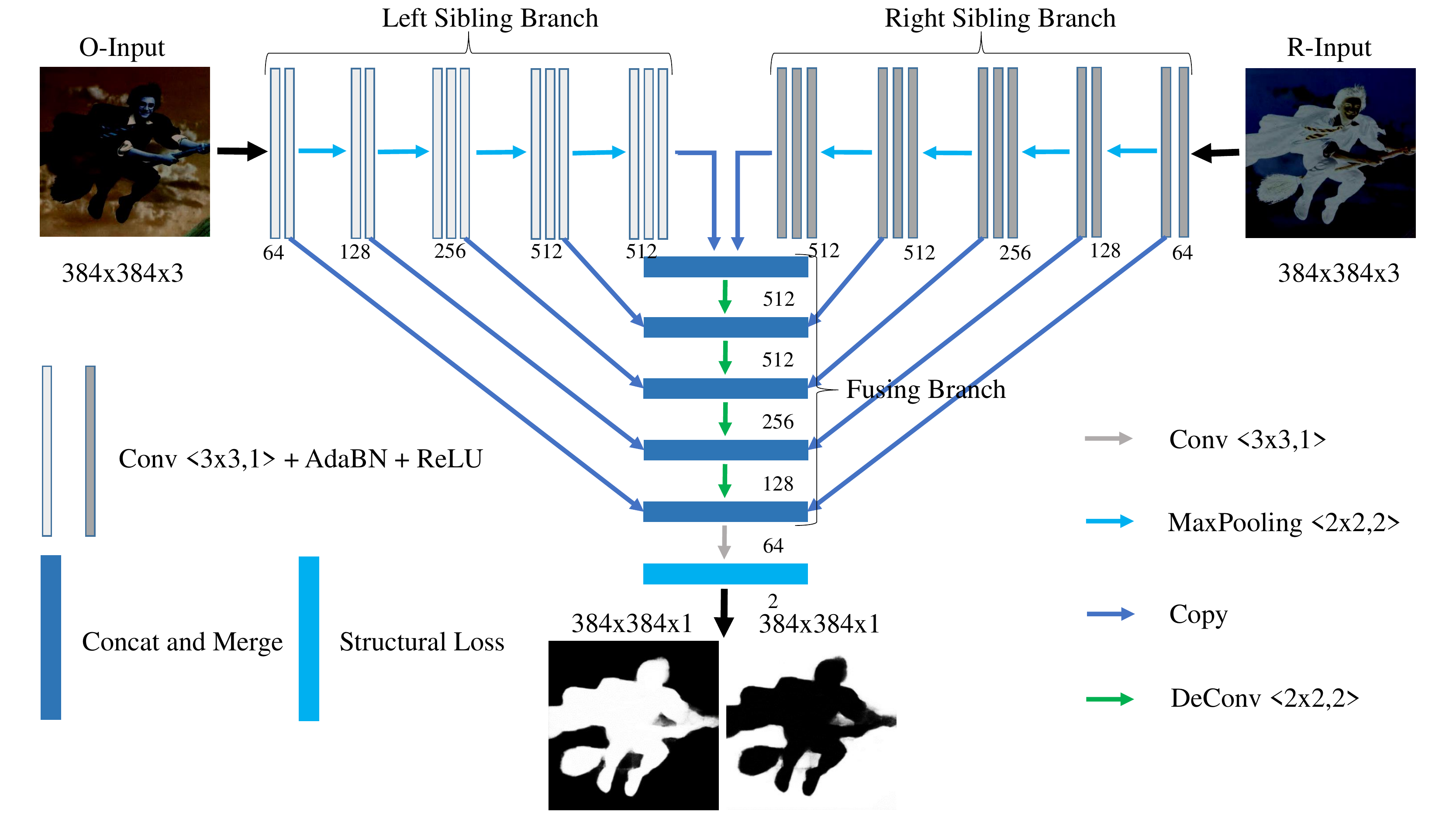}
\end{center}
\vspace{-4mm}
\caption{The semantic overview of our proposed approach. We first convert an input RGB image into a reciprocal image pair, including the origin image (O-Input) and the reflection image (R-Input).
Then the image pair is fed into two sibling branches, extracting multi-level deep features.
Each sibling branch is based on the VGG-16 model~\cite{simonyan2014very}, which has 13 convolutional layers (kernel size = $3\times 3$, stride size = 1) and 4 max pooling layers (pooling size = $2\times 2$, stride = 2).
To achieve the lossless reflection features, the two sibling branches are designed to share weights in convolutional layers, but with an adaptive batch normalization (AdaBN).
The different colors of the sibling branches show the differences in corresponding convolutional layers.
Afterwards, the fusing branch hierarchically integrates the complementary features for the saliency prediction with the proposed structural loss.
The numbers illustrate the detailed filter setting in each convolutional layer.
More details can be found in the main text.
}
\label{fig:framework}
\vspace{-4mm}
\end{figure*}
\vspace{-2mm}
{\flushleft\textbf{Two-stream Deep Networks.}} Due to their power in exploring feature fusion, two-stream deep networks~\cite{simonyan2014two} have been popular in various multi-modal computer vision tasks, such as action recognition, face detection, person re-identification, etc.
Our proposed SFCN is encouraged by their strengths.
In two-stream deep networks, the information from different sources is typically combined with the input fusion, early fusion and late fusion stage at a specific fusion position.
However, prior works using two-stream networks train both streams separately, which prevents the network from exploiting regularities between the two streams.
To enhance such fusion, several multi-level ad-hoc fusion methods~\cite{chung2017two,tran2017two} have also been introduced by considering the relationship between different scales and levels of deep features.
Different from previous methods, we argue that dense fusion positions (shown in Fig.\ref{fig:framework}) can be applied to
the feature fusion problem to enhance the fusion process, while few works take this fact into account.
Moreover, human vision system perceives a scene in a coarse-to-fine manner~\cite{allman1985stimulus}, which includes coarse understanding for identifying the location and shape of the target object, and fine capturing for exploring its detailed parts.
Thus, in this paper we fuse the multi-scale features in a coarse-to-fine manner.
Our fusion method ensures to highlight the salient objects from a global perspective and obtain
clear object boundaries in a local view.
\section{The Proposed Approach}
Fig.~\ref{fig:framework} illustrates the semantic overview of our proposed approach.
In our method, we first convert an input RGB image into a reciprocal image pair, including the origin image (O-Input) and the reflection image (R-Input), by utilizing the ImageNet mean~\cite{deng2009imagenet} and a pixel-wise negation operator.
Then the image pair is fed into the sibling branches of our proposed SFCN, extracting multi-level deep features.
Afterwards, the fusing branch hierarchically integrates the complementary features into the same resolution of input images.
Finally, the saliency map is predicted by exploiting integrated features and the proposed structural loss.
In the following subsections, we will elaborate the proposed SFCN architecture, weighted structural loss and saliency inference in detail.
\subsection{Symmetrical FCN}
The proposed SFCN is an end-to-end fully convolutional network.
It consists of three main branches with a paired reciprocal image input to achieve lossless feature reflection learning.
We describe each of them as follows.
\vspace{-2mm}
{\flushleft\textbf{Reciprocal Image Input.}}
Essentially, computer vision system understands complex environments from planar perception.
Image intrinsic reflection plays an important role in this perception process~\cite{levin2007user}.
It aims to separate an observed image into a transmitted scene and a reflected scene.
When two scenes are separated adequately, existing computer vision algorithms can better understand each scene since an interference of the other one is decreased.
Motivated by the complementary of image reflection information~\cite{land1971lightness}, we first convert the given RGB image $X\in \mathbb{R}^{W\times H\times 3}$ into a reciprocal image pair by exploiting the following planar reflection function,
\begin{equation}
\begin{aligned}
Ref(X,k) &= (X-M,\phi(k,M-X))\\
&= (X-M,-k(X-M))\\
&= (X_O, X^{k}_R).
\end{aligned}
\label{equ:equ1}
\end{equation}
where $k$ is a hyperparameter to control the reflection scale, $\phi$ is a content-persevering transform, and $M\in \mathbb{R}^{W\times H\times 3}$ is the mean of an image or image dataset.
From above equations, one can see that the converted image pair, \emph{i.e.}, $X_O$ and $X^{k}_R$, is reciprocal with a reflection plane.
In detail, the reflection scheme is a pixel-wise negation operator, allowing the given images to be reflected in both positive and negative directions while maintaining the same objects of images, as shown in Fig.~\ref{fig:framework}.
In the proposed reflection, we adopt the multiplicative operator $k$ to measure the reflection scale, but it is not the only feasible method.
For example, this reflection can be combined with other non-linear operator $\phi$, such as quadratic form and exponential transform, to add more diversity.
Besides, changing $k$ will result into different reflection planes. We perform ablation experiments to analyze the effects in Section IV. D.
To reduce the computation burden, in this paper we mainly use the image planar reflection ($k=1$) and the mean of the ImageNet dataset~\cite{deng2009imagenet}, which is pre-computed from large-scale natural image datasets and well-known in current computer vision community.
\vspace{-2mm}
{\flushleft\textbf{Sibling Branches with Layer-wise AdaBN.}}
Based on the reciprocal image pair, we propose two sibling branches to extract complementary reflection features.
More specifically, we follow the two-stream structure~\cite{simonyan2014two} and build each sibling branch based on the VGG-16 model~\cite{simonyan2014very}.
Each sibling branch has 13 convolutional layers (kernel size = $3\times 3$, stride size = 1) and 4 max pooling layers (pooling size = $2\times 2$, stride = 2).
To achieve the lossless reflection features, the two sibling branches are designed to share weights in each convolutional layer, but with an adaptive batch normalization (AdaBN).
The main reason of AdaBN design is that after the reflection transform, the reciprocal images have different image domains.
Domain related knowledge heavily affects the statistics of BN layers.
In order to learn domain invariant features, it's beneficial for each domain to keep its own BN statistics in each layer.
For the implementation, we keep the weights of corresponding convolutional layers of the two sibling branches the same, while use different learnable BN operators between the convolution and ReLU operators~\cite{zhang2017amulet}.
\begin{figure}
\begin{center}
\begin{tabular}{@{}c@{}c@{}c@{}c}
\includegraphics[width=\linewidth,height=4cm]{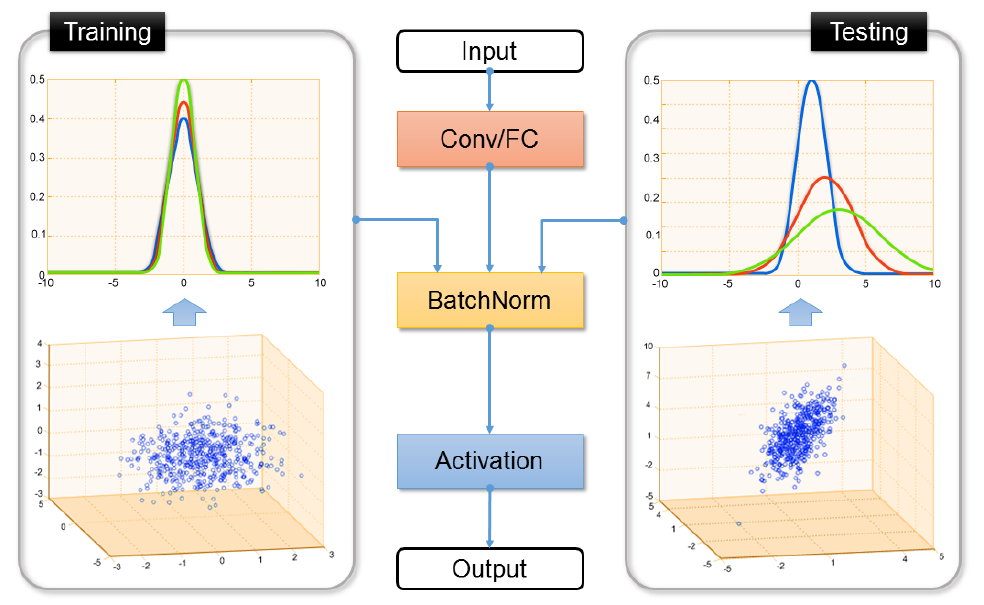} \ \\
{\small(a)}\\
\includegraphics[width=\linewidth,height=4cm]{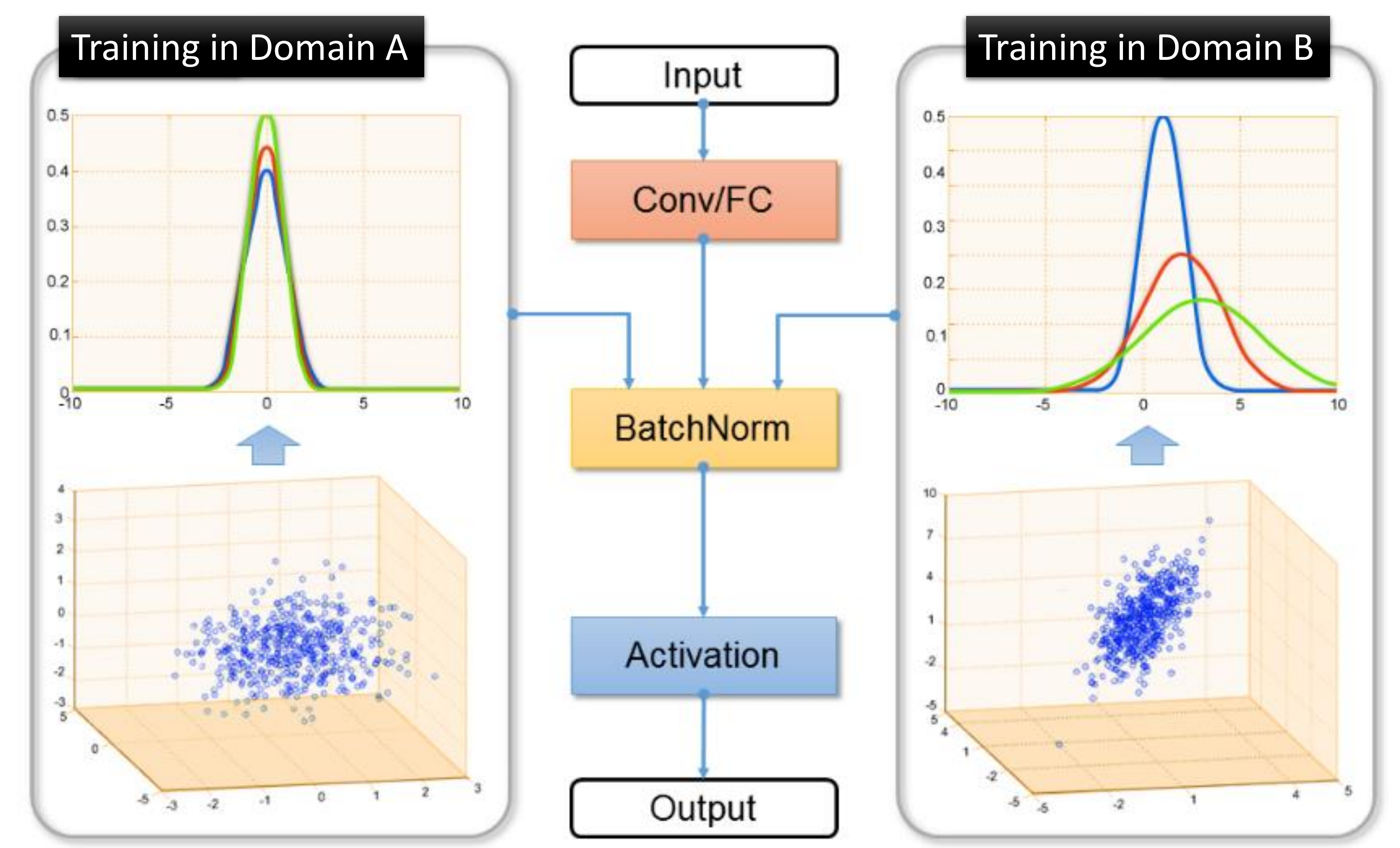} \ \\
{\small(b)}\\
\end{tabular}
\caption{Visual comparison of a) regular batch normalization (BN)~\cite{ioffe2015batch} and b) our layer-wise adaptive batch normalization (AdaBN). The pink block are convolutional/fully connected layers with shared weights. The blue block are activation functions. The scatter points are feature distributions from the source domain (left) and target domain (left). For the regular BN, there is a large domain shift for the testing, which may decrease the performance. In contrast, our layer-wise AdaBN jointly learns the domain-dependent statistics, which automatically capture the characteristic of different domains.}
\label{fig:adabn}
\end{center}
\vspace{-8mm}
\end{figure}

Besides, both shallow layers and deep layers of CNNs are influenced by domain shift, thus domain
adaptation by manipulating specific layer alone is not enough.
Therefore, we extend the AdaBN~\cite{zhang2018salient,li2016revisiting} to the layer-wise AdaBN, which computes the mean and variance in each corresponding layer from different domains.
Fig.~\ref{fig:adabn} shows the key difference between the regular BN~\cite{ioffe2015batch} and our layer-wise AdaBN.
The regular BN layer is originally designed to alleviate the issue of internal covariate shifting -- a common problem when training a very deep neural network.
It first standardizes each feature in a mini-batch, and then learns a common slope and bias for each mini-batch.
Formally, given the input to a BN layer $X\in\mathbb{R}^{n\times d}$, where $n$ denotes the batch size, and $d$ is the feature dimension, the BN layer transforms the input into:
\begin{align}
\hat{x}_i &= \frac{x_i-\mathbb{E}(X_{.i})}{\sqrt{Var(X_{.i})}}, \\
\quad \bar{x}_i &= \alpha_i \hat{x}_i + \beta_i,
  \label{equ:equ5}
\end{align}
where $x_i$ and $\bar{x}_i$ ($i\in {1, ... , d}$) are the input/output scalars of one neuron response in one data sample; $X_{.i}$ denotes the $i$-th column of the input data; and $\alpha_i$ and $\beta_i$ are parameters to be learned.
This transformation guarantees that the input distribution of each layer remains unchanged across different mini-batches.
During the testing phase, the global statistics of all training samples is used to normalize every mini-batch of test data, as shown in Fig.~\ref{fig:adabn} (a). Thus, there is a large domain shift for the testing, which may decrease the performance for the target domain.
Different from the regular BN, we adopt different learnable BN operators between each convolution and ReLU operator~\cite{zhang2017amulet} for each domain.
The intuition behind our method is straightforward: The standardization of each layer by AdaBN
ensures that each layer receives data from a similar distribution, no matter it comes from the source
domain or the target domain.
During training AdaBN, the domain-dependent statistics are jointly calculated from corresponding layers, which automatically capture the characteristic of different domains.
During the testing phase, our layer-wise AdaBN can retain the statistics of different domains for all testing samples.
Given two corresponding convolutional responses of the two sibling branches (treated as two independent domains, \emph{i.e.}, source domain and target domain), our layer-wise AdaBN algorithm is described in Algorithm 1.
In practice, following previous work~\cite{li2016revisiting}, we adopt an online algorithm to efficiently estimate the mean and variance.
\begin{table}[htbp]
\begin{tabular}{p{0.95\linewidth}}
    \toprule
    \textbf{Algorithm 1:} Layer-wise Adaptive Batch Normalization (AdaBN)\\
    \midrule
    \textbf{Input: } Neuron responses $x_{i}$ in $l$-th convolutional layer.\\
    \textbf{Output: } Neuron responses $\bar{x}_{i}$ in $l$-th convolutional layer.\\
    \textbf{1: }  \textbf{for} each neuron $i$ in SFCN \textbf{do}\\
    \textbf{2: }  \quad Concatenate neuron responses on all images in a minibatch of target domain $t$ (the other sibling branch): $\textbf{x}_{i}=[..., x_{i}(m), ...] $\\
    \textbf{3: }  \quad Compute the mean and variance of the target domain: \\$\mu^{t}_{i}=\mathbb{E}(\textbf{x}^{t}_{i}), \sigma^{t}_{i}=\sqrt{Var(\textbf{x}^{t}_{i})}$.\\
    \textbf{4: }   \textbf{for} each neuron $i$ in SFCN, testing image $m$ in target domain \textbf{do}\\
    \textbf{5: }   \quad Compute BN output $\bar{x}_{i}(m)= \alpha_i \frac{x_{i}(m)-\mu^{t}_{i}}{\sigma^{t}_{i}} + \beta_i$.\\
    \textbf{6: }  \textbf{end}\\
    \bottomrule
\end{tabular}
\label{alg:Alg1}
\vspace{-4mm}
\end{table}
\vspace{-2mm}
{\flushleft\textbf{Hierarchical Feature Fusion.}}
After extracting multi-level reflection features from the two sibling branches, we adhere an additional fusing branch to integrate them for the final saliency prediction.
In order to preserve the spatial structure of input images and enhance the contextual information, we integrate the multi-level reflection features in a hierarchical manner.
Formally, the fusing function is defined by
\begin{equation}
  \label{equ:equ6}
f_{l}(X)=
\left\{
\begin{aligned}
&h([g_{l}(X_O), f_{l+1}(X), g^{*}_{l}(X^{k}_R)]),l<L\\
&h([g_{l}(X_O), g^{*}_{l}(X^{k}_R)]), l = L
\end{aligned}
\right.
\end{equation}
where $h$ denotes the integration operator, which is a $1\times 1$ convolutional layer followed by a deconvolutional layer to ensure the same resolution.
$[\cdot]$ is the concatenation operator in channel-wise.
$g_{l}$ and $g^{*}_{l}$ are the reflection features of the $l$-th convolutional layer in the two sibling branches, respectively.
We note that though the structures of $g_{l}$ and $g^{*}_{l}$ are exactly the same, the resulting reflection features are essentially different.
The main reason is that they are extracted from different inputs and by using different statistics of AdaBN.
As shown in the experimental section, they are very complementary for improving the saliency detection.
We also find that there are other fusion methods, such as hierarchical recurrent convolutions~\cite{liu2016dhsnet}, multi-level refinement~\cite{pinheiro2016learning} and progressive attention recurrent network~\cite{zhang2018progressive}.
These feature refinement modules can further improve the performances and are complementary to our work.
However, they are more complex than ours.
Our work mainly aims to learn lossless reflection features and incorporate the structural loss. So we keep the fusion module simple.

In the end of the fusion, we add a convolutional layer with two $3\times3$ filters for the saliency prediction.
The numbers of Fig.~\ref{fig:framework} illustrate the detailed filter setting in each convolutional layer.
More details can be found in Section IV. B.
\vspace{-2mm}
\subsection{Weighted Structural Loss}
Given the SOD training dataset $ S=\{(X_n,Y_n)\}^{N}_{n=1}$ with $N$ training pairs, where $X_n =
\{x^n_j, j = 1,...,T\}$ and $Y_n = \{y^n_j, j = 1,...,T\}$ are the input image and the binary ground-truth image with $T$ pixels, respectively.
$y^n_j = 1$ denotes the foreground pixel and $y^n_j = 0$ denotes the background pixel.
For notional simplicity, we subsequently drop the subscript $n$ and consider each image independently.
In most of existing deep learning based SOD methods, the loss function used to train the network is the standard pixel-wise binary cross-entropy (BCE) loss:
\begin{equation}
  \label{equ:equ7}
\begin{aligned}
  \mathcal{L}_{bce}= -\sum_{j\in Y_{+}} \text{log~Pr}(y_{j}=1|X;\theta)\\-\sum_{j\in Y_{-}} \text{log~Pr}(y_{j}=0|X;\theta).
\end{aligned}
\end{equation}
where $\theta$ is the parameter of the network.
Pr$(y_j =1|X;\theta)\in [0,1]$ is the confidence score of the network prediction that measures how likely the pixel belong to the foreground.
$Y_{+}$ and $Y_{-}$ denote the foreground and background pixel set in an image, respectively.

However, for a typical natural image, the class distribution of salient/non-salient pixels is heavily imbalanced: most of the pixels in the ground truth are non-salient.
To automatically balance the loss between positive/negative classes, we introduce a class-balancing weight $\rho$ on a per-pixel term basis, following~\cite{xie2015holistically,zhang2017amulet}.
Specifically, we define the weighted binary cross-entropy (WBCE) loss function as,
\begin{equation}
  \label{equ:equ8}
\begin{aligned}
  \mathcal{L}_{wbce}= - \rho \sum_{j\in Y_{+}} \text{log~Pr}(y_{j}=1|X;\theta)\\
  -(1-\rho)\sum_{j\in Y_{-}} \text{log~Pr}(y_{j}=0|X;\theta).
\end{aligned}
\end{equation}
The loss weight $\rho = |Y_{+}|/|Y|$, and $|Y_{+}|$ and $|Y_{-}|$ denote the foreground and background pixel number, respectively.

For saliency detection, it is also crucial to preserve the overall spatial structure and semantic content.
Thus, rather than only encouraging the pixels of the output prediction to match with the ground-truth using the above pixel-wise loss, we also minimize the differences between their multi-level features by a deep convolutional network~\cite{johnson2016perceptual}.
The main intuition behind this operator is that minimizing the difference between multi-level features, which encode low-level fine details and high-level coarse semantics, helps to retain the spatial structure and semantic content of predictions.
Formally, let $\phi_{\tilde{l}}$ denotes the output of the $\tilde{l}$-th convolutional layer in a CNN, our semantic content (SC) loss is defined as
\begin{align}
\mathcal{L}_{sc} = \sum_{\tilde{l}=1}^{\tilde{L}}\lambda_{\tilde{l}}||\phi_{\tilde{l}}(Y;w)-\phi_{\tilde{l}}(\hat{Y};w)||_2,
  \label{equ:equ9}
\end{align}
where $\hat{Y}$ is the overall prediction, $w$ is the parameter of a pre-trained CNN and $\lambda_{\tilde{l}}$ is the trade-off parameter, controlling the influence of the loss in the $\tilde{l}$-th layer.
In our work, we use the light CNN-9 model~\cite{wu2015lightened} pre-trained on the ImageNet dataset as the semantic CNN.
Each layer of the network will provide a cascade loss between the ground-truth and the prediction.
More specifically, the semantic content loss feeds the prediction $\hat{Y}$ and ground truth $Y$ to the pre-trained CNN-9 model and compares the intermediate features.
It's believed that the cascade loss from the higher layers control the global structure; the loss from the lower layers control the local detail during training.
We verify the loss with different layers of the light CNN-9 model for feature reconstruction loss by setting value of $\tilde{l}$ to one of 9.
Besides, following~\cite{johnson2016perceptual}, parameters of the CNN-9 model is fixed during the training.
The cascade loss can provide the ability to measure the similarity between the ground-truth and the prediction under $\tilde{L}$ different scales and enhance the ability of the saliency classifier.

To overcome the blurry boundary problem~\cite{li2016deepsaliency}, we also introduce an edge-persevering loss which encourages to keep the details of boundaries of salient objects.
In fact, the edge-preserving $L_1$ loss is frequently used in low-level vision tasks such as image super-resolution, denoising and deblurring.
However, the exact $L_1$ loss is not continuously differentiable, so we use the soomth $L_1$ loss, \emph{i.e.}, modified Huber Loss, for our task and ensure the gradient learning.
Specifically, the smooth $L_1$ loss function is defined as
\begin{equation}
  \label{equ:equ10}
\mathcal{L}_{s1}=
\left\{
\begin{aligned}
&\frac{1}{2\epsilon}||D||^2_2+\frac{1}{2}\epsilon,&||D||_1<\epsilon\\
&||D||_1, &otherwise
\end{aligned}
\right.
\end{equation}
where $D=Y-\hat{Y}$ and $\epsilon$ is a predefined threshold. We set $\epsilon=0.1$, following the practice in~\cite{wang2016robust}.
This training loss also helps to minimize pixel-level differences between the overall prediction and the ground-truth.

By taking all above loss functions together, we define our final loss function as
\begin{align}
\mathcal{L} = \text{arg min}~\mathcal{L}_{wbce}+\mu \mathcal{L}_{sc}+\gamma \mathcal{L}_{s1},
  \label{equ:equ11}
\end{align}
where $\mu$ and $\gamma$ are hyperparameters to balance the specific terms.
All the above losses are continuously differentiable, so we can use the standard stochastic gradient descent (SGD) method~\cite{ruder2016overview} to obtain the optimal parameters.
\subsection{Saliency Inference}
For saliency inference, different from the methods used sigmoid classifiers in~\cite{liu2016dhsnet,xie2015holistically}, we use the following softmax classifier to evaluate the prediction scores:
\begin{align}
\text{Pr}(y_{j}=1|X;\theta,w) = \frac{e^{z_1}}{e^{z_0}+e^{z_1}},
  \label{equ:equ8}
\end{align}
\begin{align}
\text{Pr}(y_{j}=0|X;\theta,w) = \frac{e^{z_0}}{e^{z_0}+e^{z_1}},
  \label{equ:equ9}
\end{align}
where $z_0$ and $z_1$ are the score of each label of training data. In this way, the prediction of the SFCN is composed of a foreground excitation map ($\textbf{M}^{fe}$) and a background excitation map ($\textbf{M}^{be}$), as illustrated at the bottom of Fig.~\ref{fig:framework}.
We utilize $\textbf{M}^{fe}$ and $\textbf{M}^{be}$ to generate the final result.
Formally, the final saliency map can be obtained by
\begin{align}
  \textbf{S} & = \sigma(\textbf{M}^{fe}-\textbf{M}^{be}) = max(0, \textbf{M}^{fe}-\textbf{M}^{be}),
  \label{equ:equ10}
\end{align}
where $\sigma$ is the ReLU activation function for clipping the negative values.
This strategy not only increases the pixel-level discrimination but also captures context contrast information.
\section{Experimental Results}
\subsection{Datasets and Evaluation Metrics}
To train our model, we adopt the \textbf{MSRA10K}~\cite{borji2015salient} dataset, which has 10,000 training images with high quality pixel-wise saliency annotations.
Most of images in this dataset have a single salient object.
To combat overfitting, we augment this dataset by random cropping and mirror reflection~\cite{zhang2017amulet,zhang2018salient}.

For the SOD performance evaluation, we adopt seven public saliency detection datasets described as follows:
\textbf{DUT-OMRON}~\cite{yang2013saliency} dataset has 5,168 high quality natural images.
Each image in this dataset has one or more objects with relatively complex image background.
\textbf{DUTS-TE} dataset is the test set of currently largest saliency detection benchmark (DUTS)~\cite{wang2017learning}.
It contains 5,019 images with high quality pixel-wise annotations.
\textbf{ECSSD}~\cite{shi2016hierarchical} dataset contains 1,000 natural images, in which many semantically meaningful and complex structures are included.
\textbf{HKU-IS-TE}~\cite{li2015visual} dataset has 1,447 images with pixel-wise annotations.
Images of this dataset are well chosen to include multiple disconnected objects or objects touching the image boundary.
\textbf{PASCAL-S}~\cite{li2014secrets} dataset is generated from the PASCAL VOC~\cite{Everingham2010ThePV} dataset and contains 850 natural images with segmentation-based masks.
\textbf{SED}~\cite{borj2015salient} dataset has two non-overlapped subsets, \emph{i.e.}, SED1 and SED2.
SED1 has 100 images each containing only one salient object, while SED2 has 100 images each containing two salient objects.
\textbf{SOD}~\cite{jiang2013salient} dataset has 300 images, in which many images contain multiple objects either with low contrast or touching the image boundary.

To evaluate our method with other algorithms, we follow the works in~\cite{zhang2017amulet,zhang2017learning,zhang2018salient} and use four main metrics as the standard evaluation score (with typical parameter settings), including the widely used precision-recall (PR) curves, F-measure ($\eta=0.3$), mean absolute error (MAE) and S-measure ($\lambda=0.5$).
More details can be found in~\cite{borji2015salient,fan2017structure}.
\subsection{Implementation Details}
We implement our proposed model based on the public Caffe toolbox~\cite{jia2014caffe} with the MATLAB 2016 platform.
We train and test our method in a quad-core PC machine with an NVIDIA GeForce GTX 1070 GPU (with 8G memory) and an i5-6600 CPU.
We perform training with the augmented training images from the MSRA10K dataset.
Following~\cite{zhang2017amulet,zhang2017learning}, we do not use validation set and train the model until its training loss converges.
The input image from different datasets is uniformly resized into $384\times384\times3$ pixels. Then we perform the reflection transform by subtracting the ImageNet mean~\cite{deng2009imagenet}.
The weights of two sibling branches are initialized from the VGG-16 model~\cite{simonyan2014very}.
For the fusing branch, we initialize the weights by the ``msra'' method~\cite{he2015delving}.
For the hyper-parameters ($\lambda_{\tilde{l}}$, $\mu$, $\gamma$), we mainly following the practice in previous works~\cite{johnson2016perceptual,xiao2018interactive}.
The authors in~\cite{johnson2016perceptual} proved that setting equal $\lambda_{\tilde{l}}=1$ can capture better results to utilize the multi-level semantic features.
We also adopt this setting and observe the same fact.
Besides, we follow~\cite{xiao2018interactive} to adopt $\mu=0.01$ and $\gamma =20$. We optimize the final loss function for our experiments without further tuning.
Indeed, better hyper-parameters can further improve the performances.
During the training, we use standard SGD method~\cite{ruder2016overview} with a batch size 12, momentum 0.9 and weight decay 0.0005.
We set the base learning rate to 1e$^{-8}$ and decrease the learning rate by 10\% when training loss reaches a flat.
The training process converges after 150k iterations.
When testing, our proposed SOD algorithm runs at about \textbf{12 fps}.
The source code is made publicly available at http://ice.dlut.edu.cn/lu/.
\subsection{Comparison with the State-of-the-arts}
To fully evaluate the detection performance, we compare our proposed method with other 22 state-of-the-art ones, including 18 deep learning based algorithms (\textbf{AMU}~\cite{zhang2017amulet}, \textbf{BDMP}~\cite{zhang2018bi}, \textbf{DCL}~\cite{li2016deep}, \textbf{DGRL}~\cite{wang2018detect}, \textbf{DHS}~\cite{liu2016dhsnet}, \textbf{DS}~\cite{li2016deepsaliency}, \textbf{DSS}~\cite{hou2017deeply}, \textbf{ELD}~\cite{lee2016deep}, \textbf{LEGS}~\cite{wang2015deep}, \textbf{MCDL}~\cite{zhao2015saliency}, \textbf{MDF}~\cite{li2015visual}, \textbf{NLDF}~\cite{luo2017non}, \textbf{PAGRN}~\cite{zhang2018progressive}, \textbf{PICA}~\cite{liu2018picanet}, \textbf{RFCN}~\cite{wang2016saliency}, \textbf{RST}~\cite{zhu2017saliency}, \textbf{SRM}~\cite{wang2017stagewise}, \textbf{UCF}~\cite{zhang2017learning}) and 4 conventional algorithms (\textbf{BL}~\cite{tong2015salient}, \textbf{BSCA}~\cite{qin2015saliency}, \textbf{DRFI}~\cite{jiang2013salient}, \textbf{DSR}~\cite{li2013saliency}).
For fair comparison, we use either the implementations with recommended parameter settings or the saliency maps provided by the authors.
\begin{table*}
\begin{center}
\doublerulesep=0.1pt
\caption{Quantitative comparison with 24 methods on 4 large-scale datasets. The best three results are shown in \textcolor[rgb]{1,0,0}{red},~\textcolor[rgb]{0,1,0}{green} and \textcolor[rgb]{0,0,1}{blue}, respectively. $^{*}$ means methods using multi-stage training or post-processing, such as CRF or superpixel refinements. }
\label{table:fauc1}
\resizebox{0.82\textwidth}{!}
{
\begin{tabular}{|c|c|c|c|c|c|c|c|c|c|c|c|c|c|c|c|c|c|c|c|c|c|c|c|c|||c|c|c|c|c|c|c|c|||}
\hline
\multicolumn{4}{|c|}{}
&\multicolumn{6}{|c|}{\textbf{DUT-OMRON}}
&\multicolumn{6}{|c|}{\textbf{DUTS-TE}}
&\multicolumn{6}{|c|}{\textbf{ECSSD}}
&\multicolumn{6}{|c|}{\textbf{HKU-IS-TE}}
\\
\hline
\hline
\multicolumn{4}{|c|}{Methods}
&\multicolumn{2}{|c|}{$F_\eta\uparrow$}&\multicolumn{2}{|c|}{$MAE\downarrow$}&\multicolumn{2}{|c|}{$S_\lambda\uparrow$}
&\multicolumn{2}{|c|}{$F_\eta\uparrow$}&\multicolumn{2}{|c|}{$MAE\downarrow$}&\multicolumn{2}{|c|}{$S_\lambda\uparrow$}
&\multicolumn{2}{|c|}{$F_\eta\uparrow$}&\multicolumn{2}{|c|}{$MAE\downarrow$}&\multicolumn{2}{|c|}{$S_\lambda\uparrow$}
&\multicolumn{2}{|c|}{$F_\eta\uparrow$}&\multicolumn{2}{|c|}{$MAE\downarrow$}&\multicolumn{2}{|c|}{$S_\lambda\uparrow$}
\\
\hline
\hline
\multicolumn{4}{|c|}{\textbf{Ours+}}
&\multicolumn{2}{|c|}{\textcolor[rgb]{0,1,0}{0.718}}&\multicolumn{2}{|c|}{\textcolor[rgb]{0,0,1}{0.0643}}&\multicolumn{2}{|c|}{\textcolor[rgb]{1,0,0}{0.811}}
&\multicolumn{2}{|c|}{0.742}&\multicolumn{2}{|c|}{0.0622}&\multicolumn{2}{|c|}{\textcolor[rgb]{1,0,0}{0.853}}
&\multicolumn{2}{|c|}{\textcolor[rgb]{1,0,0}{0.911}}&\multicolumn{2}{|c|}{\textcolor[rgb]{1,0,0}{0.0421}}&\multicolumn{2}{|c|}{\textcolor[rgb]{1,0,0}{0.916}}
&\multicolumn{2}{|c|}{\textcolor[rgb]{1,0,0}{0.906}}&\multicolumn{2}{|c|}{\textcolor[rgb]{1,0,0}{0.0357}}&\multicolumn{2}{|c|}{\textcolor[rgb]{1,0,0}{0.914}}
\\
\multicolumn{4}{|c|}{\textbf{Ours}~\cite{zhang2018salient}}
&\multicolumn{2}{|c|}{0.696}&\multicolumn{2}{|c|}{0.0862}&\multicolumn{2}{|c|}{0.774}
&\multicolumn{2}{|c|}{0.716}&\multicolumn{2}{|c|}{0.0834}&\multicolumn{2}{|c|}{0.799}
&\multicolumn{2}{|c|}{0.880}&\multicolumn{2}{|c|}{0.0523}&\multicolumn{2}{|c|}{0.897}
&\multicolumn{2}{|c|}{0.875}&\multicolumn{2}{|c|}{\textcolor[rgb]{0,0,1}{0.0397}}&\multicolumn{2}{|c|}{\textcolor[rgb]{0,0,1}{0.905}}
\\
\hline
\hline
\multicolumn{4}{|c|}{\textbf{AMU}~\cite{zhang2017amulet}}
&\multicolumn{2}{|c|}{0.647}&\multicolumn{2}{|c|}{0.0976}&\multicolumn{2}{|c|}{0.771}
&\multicolumn{2}{|c|}{0.682}&\multicolumn{2}{|c|}{0.0846}&\multicolumn{2}{|c|}{0.796}
&\multicolumn{2}{|c|}{0.868}&\multicolumn{2}{|c|}{0.0587}&\multicolumn{2}{|c|}{0.894}
&\multicolumn{2}{|c|}{0.843}&\multicolumn{2}{|c|}{0.0501}&\multicolumn{2}{|c|}{0.886}
\\
\multicolumn{4}{|c|}{\textbf{BDMP}~\cite{zhang2018bi}}
&\multicolumn{2}{|c|}{0.692}&\multicolumn{2}{|c|}{\textcolor[rgb]{0,1,0}{0.0636}}&\multicolumn{2}{|c|}{0.789}
&\multicolumn{2}{|c|}{0.751}&\multicolumn{2}{|c|}{\textcolor[rgb]{1,0,0}{0.0490}}&\multicolumn{2}{|c|}{\textcolor[rgb]{0,1,0}{0.851}}
&\multicolumn{2}{|c|}{0.869}&\multicolumn{2}{|c|}{\textcolor[rgb]{0,0,1}{0.0445}}&\multicolumn{2}{|c|}{\textcolor[rgb]{0,0,1}{0.911}}
&\multicolumn{2}{|c|}{0.871}&\multicolumn{2}{|c|}{0.0390}&\multicolumn{2}{|c|}{\textcolor[rgb]{0,1,0}{0.907}}
\\
\multicolumn{4}{|c|}{\textbf{DCL}$^{*}$~\cite{li2016deep}}
&\multicolumn{2}{|c|}{0.684}&\multicolumn{2}{|c|}{0.1573}&\multicolumn{2}{|c|}{0.743}
&\multicolumn{2}{|c|}{0.714}&\multicolumn{2}{|c|}{0.1493}&\multicolumn{2}{|c|}{0.785}
&\multicolumn{2}{|c|}{0.829}&\multicolumn{2}{|c|}{0.1495}&\multicolumn{2}{|c|}{0.863}
&\multicolumn{2}{|c|}{0.853}&\multicolumn{2}{|c|}{0.1363}&\multicolumn{2}{|c|}{0.859}
\\
\multicolumn{4}{|c|}{\textbf{DGRL}~\cite{wang2018detect}}
&\multicolumn{2}{|c|}{0.709}&\multicolumn{2}{|c|}{\textcolor[rgb]{1,0,0}{0.0633}}&\multicolumn{2}{|c|}{0.791}
&\multicolumn{2}{|c|}{\textcolor[rgb]{0,0,1}{0.768}}&\multicolumn{2}{|c|}{\textcolor[rgb]{0,1,0}{0.0512}}&\multicolumn{2}{|c|}{0.836}
&\multicolumn{2}{|c|}{0.880}&\multicolumn{2}{|c|}{\textcolor[rgb]{0,1,0}{0.0424}}&\multicolumn{2}{|c|}{0.906}
&\multicolumn{2}{|c|}{0.882}&\multicolumn{2}{|c|}{\textcolor[rgb]{0,1,0}{0.0378}}&\multicolumn{2}{|c|}{0.896}
\\
\multicolumn{4}{|c|}{\textbf{DHS}~\cite{liu2016dhsnet}}
&\multicolumn{2}{|c|}{--}&\multicolumn{2}{|c|}{--}&\multicolumn{2}{|c|}{--}
&\multicolumn{2}{|c|}{0.724}&\multicolumn{2}{|c|}{0.0673}&\multicolumn{2}{|c|}{0.809}
&\multicolumn{2}{|c|}{0.872}&\multicolumn{2}{|c|}{0.0601}&\multicolumn{2}{|c|}{0.884}
&\multicolumn{2}{|c|}{0.854}&\multicolumn{2}{|c|}{0.0532}&\multicolumn{2}{|c|}{0.869}
\\
\multicolumn{4}{|c|}{\textbf{DS}$^{*}$~\cite{li2016deepsaliency}}
&\multicolumn{2}{|c|}{0.603}&\multicolumn{2}{|c|}{0.1204}&\multicolumn{2}{|c|}{0.741}
&\multicolumn{2}{|c|}{0.632}&\multicolumn{2}{|c|}{0.0907}&\multicolumn{2}{|c|}{0.790}
&\multicolumn{2}{|c|}{0.826}&\multicolumn{2}{|c|}{0.1216}&\multicolumn{2}{|c|}{0.821}
&\multicolumn{2}{|c|}{0.787}&\multicolumn{2}{|c|}{0.0774}&\multicolumn{2}{|c|}{0.854}
\\
\multicolumn{4}{|c|}{\textbf{DSS}$^{*}$~\cite{hou2017deeply}}
&\multicolumn{2}{|c|}{\textcolor[rgb]{1,0,0}{0.729}}&\multicolumn{2}{|c|}{0.0656}&\multicolumn{2}{|c|}{0.765}
&\multicolumn{2}{|c|}{\textcolor[rgb]{1,0,0}{0.791}}&\multicolumn{2}{|c|}{0.0564}&\multicolumn{2}{|c|}{0.812}
&\multicolumn{2}{|c|}{\textcolor[rgb]{0,1,0}{0.904}}&\multicolumn{2}{|c|}{0.0517}&\multicolumn{2}{|c|}{0.882}
&\multicolumn{2}{|c|}{\textcolor[rgb]{0,1,0}{0.902}}&\multicolumn{2}{|c|}{0.0401}&\multicolumn{2}{|c|}{0.878}
\\
\multicolumn{4}{|c|}{\textbf{ELD}$^{*}$~\cite{lee2016deep}}
&\multicolumn{2}{|c|}{0.611}&\multicolumn{2}{|c|}{0.0924}&\multicolumn{2}{|c|}{0.743}
&\multicolumn{2}{|c|}{0.628}&\multicolumn{2}{|c|}{0.0925}&\multicolumn{2}{|c|}{0.749}
&\multicolumn{2}{|c|}{0.810}&\multicolumn{2}{|c|}{0.0795}&\multicolumn{2}{|c|}{0.839}
&\multicolumn{2}{|c|}{0.776}&\multicolumn{2}{|c|}{0.0724}&\multicolumn{2}{|c|}{0.823}
\\
\multicolumn{4}{|c|}{\textbf{LEGS}$^{*}$~\cite{wang2015deep}}
&\multicolumn{2}{|c|}{0.592}&\multicolumn{2}{|c|}{0.1334}&\multicolumn{2}{|c|}{0.701}
&\multicolumn{2}{|c|}{0.585}&\multicolumn{2}{|c|}{0.1379}&\multicolumn{2}{|c|}{0.687}
&\multicolumn{2}{|c|}{0.785}&\multicolumn{2}{|c|}{0.1180}&\multicolumn{2}{|c|}{0.787}
&\multicolumn{2}{|c|}{0.732}&\multicolumn{2}{|c|}{0.1182}&\multicolumn{2}{|c|}{0.745}
\\
\multicolumn{4}{|c|}{\textbf{MCDL}$^{*}$~\cite{zhao2015saliency}}
&\multicolumn{2}{|c|}{0.625}&\multicolumn{2}{|c|}{0.0890}&\multicolumn{2}{|c|}{0.739}
&\multicolumn{2}{|c|}{0.594}&\multicolumn{2}{|c|}{0.1050}&\multicolumn{2}{|c|}{0.706}
&\multicolumn{2}{|c|}{0.796}&\multicolumn{2}{|c|}{0.1007}&\multicolumn{2}{|c|}{0.803}
&\multicolumn{2}{|c|}{0.760}&\multicolumn{2}{|c|}{0.0908}&\multicolumn{2}{|c|}{0.786}
\\
\multicolumn{4}{|c|}{\textbf{MDF}$^{*}$~\cite{li2015visual}}
&\multicolumn{2}{|c|}{0.644}&\multicolumn{2}{|c|}{0.0916}&\multicolumn{2}{|c|}{0.703}
&\multicolumn{2}{|c|}{0.673}&\multicolumn{2}{|c|}{0.0999}&\multicolumn{2}{|c|}{0.723}
&\multicolumn{2}{|c|}{0.807}&\multicolumn{2}{|c|}{0.1049}&\multicolumn{2}{|c|}{0.776}
&\multicolumn{2}{|c|}{0.802}&\multicolumn{2}{|c|}{0.0946}&\multicolumn{2}{|c|}{0.779}
\\
\multicolumn{4}{|c|}{\textbf{NLDF}~\cite{luo2017non}}
&\multicolumn{2}{|c|}{0.684}&\multicolumn{2}{|c|}{0.0796}&\multicolumn{2}{|c|}{0.750}
&\multicolumn{2}{|c|}{0.743}&\multicolumn{2}{|c|}{0.0652}&\multicolumn{2}{|c|}{0.805}
&\multicolumn{2}{|c|}{0.878}&\multicolumn{2}{|c|}{0.0627}&\multicolumn{2}{|c|}{0.875}
&\multicolumn{2}{|c|}{0.872}&\multicolumn{2}{|c|}{0.0480}&\multicolumn{2}{|c|}{0.878}
\\
\multicolumn{4}{|c|}{\textbf{PAGRN}~\cite{zhang2018progressive}}
&\multicolumn{2}{|c|}{\textcolor[rgb]{0,0,1}{0.711}}&\multicolumn{2}{|c|}{0.0709}&\multicolumn{2}{|c|}{0.751}
&\multicolumn{2}{|c|}{\textcolor[rgb]{0,1,0}{0.788}}&\multicolumn{2}{|c|}{0.0555}&\multicolumn{2}{|c|}{0.825}
&\multicolumn{2}{|c|}{\textcolor[rgb]{0,0,1}{0.894}}&\multicolumn{2}{|c|}{0.0610}&\multicolumn{2}{|c|}{0.889}
&\multicolumn{2}{|c|}{\textcolor[rgb]{0,0,1}{0.887}}&\multicolumn{2}{|c|}{0.0475}&\multicolumn{2}{|c|}{0.889}
\\
\multicolumn{4}{|c|}{\textbf{PICA}$^{*}$~\cite{liu2018picanet}}
&\multicolumn{2}{|c|}{0.710}&\multicolumn{2}{|c|}{0.0676}&\multicolumn{2}{|c|}{\textcolor[rgb]{0,1,0}{0.808}}
&\multicolumn{2}{|c|}{0.755}&\multicolumn{2}{|c|}{\textcolor[rgb]{0,0,1}{0.0539}}&\multicolumn{2}{|c|}{\textcolor[rgb]{0,0,1}{0.850}}
&\multicolumn{2}{|c|}{0.884}&\multicolumn{2}{|c|}{0.0466}&\multicolumn{2}{|c|}{\textcolor[rgb]{0,1,0}{0.914}}
&\multicolumn{2}{|c|}{0.870}&\multicolumn{2}{|c|}{0.0415}&\multicolumn{2}{|c|}{0.905}
\\
\multicolumn{4}{|c|}{\textbf{RFCN}$^{*}$~\cite{wang2016saliency}}
&\multicolumn{2}{|c|}{0.627}&\multicolumn{2}{|c|}{0.1105}&\multicolumn{2}{|c|}{0.752}
&\multicolumn{2}{|c|}{0.712}&\multicolumn{2}{|c|}{0.0900}&\multicolumn{2}{|c|}{0.784}
&\multicolumn{2}{|c|}{0.834}&\multicolumn{2}{|c|}{0.1069}&\multicolumn{2}{|c|}{0.852}
&\multicolumn{2}{|c|}{0.838}&\multicolumn{2}{|c|}{0.0882}&\multicolumn{2}{|c|}{0.860}
\\
\multicolumn{4}{|c|}{\textbf{RST}$^{*}$~\cite{zhu2017saliency}}
&\multicolumn{2}{|c|}{0.707}&\multicolumn{2}{|c|}{0.0753}&\multicolumn{2}{|c|}{\textcolor[rgb]{0,0,1}{0.792}}
&\multicolumn{2}{|c|}{--}&\multicolumn{2}{|c|}{--}&\multicolumn{2}{|c|}{--}
&\multicolumn{2}{|c|}{0.884}&\multicolumn{2}{|c|}{0.0678}&\multicolumn{2}{|c|}{0.899}
&\multicolumn{2}{|c|}{--}&\multicolumn{2}{|c|}{--}&\multicolumn{2}{|c|}{--}
\\
\multicolumn{4}{|c|}{\textbf{SRM}$^{*}$~\cite{wang2017stagewise}}
&\multicolumn{2}{|c|}{0.707}&\multicolumn{2}{|c|}{0.0694}&\multicolumn{2}{|c|}{0.777}
&\multicolumn{2}{|c|}{0.757}&\multicolumn{2}{|c|}{0.0588}&\multicolumn{2}{|c|}{0.824}
&\multicolumn{2}{|c|}{0.892}&\multicolumn{2}{|c|}{0.0543}&\multicolumn{2}{|c|}{0.895}
&\multicolumn{2}{|c|}{0.873}&\multicolumn{2}{|c|}{0.0458}&\multicolumn{2}{|c|}{0.887}
\\
\multicolumn{4}{|c|}{\textbf{UCF}~\cite{zhang2017learning}}
&\multicolumn{2}{|c|}{0.621}&\multicolumn{2}{|c|}{0.1203}&\multicolumn{2}{|c|}{0.748}
&\multicolumn{2}{|c|}{0.635}&\multicolumn{2}{|c|}{0.1119}&\multicolumn{2}{|c|}{0.777}
&\multicolumn{2}{|c|}{0.844}&\multicolumn{2}{|c|}{0.0689}&\multicolumn{2}{|c|}{0.884}
&\multicolumn{2}{|c|}{0.823}&\multicolumn{2}{|c|}{0.0612}&\multicolumn{2}{|c|}{0.874}
\\
\hline
\hline
\multicolumn{4}{|c|}{\textbf{BL}~\cite{tong2015salient}}
&\multicolumn{2}{|c|}{0.499}&\multicolumn{2}{|c|}{0.2388}&\multicolumn{2}{|c|}{0.625}
&\multicolumn{2}{|c|}{0.490}&\multicolumn{2}{|c|}{0.2379}&\multicolumn{2}{|c|}{0.615}
&\multicolumn{2}{|c|}{0.684}&\multicolumn{2}{|c|}{0.2159}&\multicolumn{2}{|c|}{0.714}
&\multicolumn{2}{|c|}{0.666}&\multicolumn{2}{|c|}{0.2048}&\multicolumn{2}{|c|}{0.702}
\\
\multicolumn{4}{|c|}{\textbf{BSCA}~\cite{qin2015saliency}}
&\multicolumn{2}{|c|}{0.509}&\multicolumn{2}{|c|}{0.1902}&\multicolumn{2}{|c|}{0.648}
&\multicolumn{2}{|c|}{0.500}&\multicolumn{2}{|c|}{0.1961}&\multicolumn{2}{|c|}{0.633}
&\multicolumn{2}{|c|}{0.705}&\multicolumn{2}{|c|}{0.1821}&\multicolumn{2}{|c|}{0.725}
&\multicolumn{2}{|c|}{0.658}&\multicolumn{2}{|c|}{0.1738}&\multicolumn{2}{|c|}{0.705}
\\
\multicolumn{4}{|c|}{\textbf{DRFI}~\cite{jiang2013salient}}
&\multicolumn{2}{|c|}{0.550}&\multicolumn{2}{|c|}{0.1378}&\multicolumn{2}{|c|}{0.688}
&\multicolumn{2}{|c|}{0.541}&\multicolumn{2}{|c|}{0.1746}&\multicolumn{2}{|c|}{0.662}
&\multicolumn{2}{|c|}{0.733}&\multicolumn{2}{|c|}{0.1642}&\multicolumn{2}{|c|}{0.752}
&\multicolumn{2}{|c|}{0.726}&\multicolumn{2}{|c|}{0.1431}&\multicolumn{2}{|c|}{0.743}
\\
\multicolumn{4}{|c|}{\textbf{DSR}~\cite{li2013saliency}}
&\multicolumn{2}{|c|}{0.524}&\multicolumn{2}{|c|}{0.1389}&\multicolumn{2}{|c|}{0.660}
&\multicolumn{2}{|c|}{0.518}&\multicolumn{2}{|c|}{0.1455}&\multicolumn{2}{|c|}{0.646}
&\multicolumn{2}{|c|}{0.662}&\multicolumn{2}{|c|}{0.1784}&\multicolumn{2}{|c|}{0.731}
&\multicolumn{2}{|c|}{0.682}&\multicolumn{2}{|c|}{0.1405}&\multicolumn{2}{|c|}{0.701}
\\
\hline
\end{tabular}
}
\vspace{-2mm}
\end{center}
\vspace{-2mm}
\end{table*}
\begin{table*}
\begin{center}
\doublerulesep=0.1pt
\caption{Quantitative comparison with 24 methods on 4 complex structure datasets. The best three results are shown in \textcolor[rgb]{1,0,0}{red},~\textcolor[rgb]{0,1,0}{green} and \textcolor[rgb]{0,0,1}{blue}, respectively. $^{*}$ means methods using multi-stage training or post-processing, such as CRF or superpixel refinements.}
\label{table:fauc2}
\resizebox{0.82\textwidth}{!}
{
\begin{tabular}{|c|c|c|c|c|c|c|c|c|c|c|c|c|c|c|c|c|c|c|c|c|c|c|c|c|||c|c|c|c|c|c|c|c|||}
\hline
\multicolumn{4}{|c|}{}
&\multicolumn{6}{|c|}{\textbf{PASCAL-S}}
&\multicolumn{6}{|c|}{\textbf{SED1}}
&\multicolumn{6}{|c|}{\textbf{SED2}}
&\multicolumn{6}{|c|}{\textbf{SOD}}
\\
\hline
\hline
\multicolumn{4}{|c|}{Methods}
&\multicolumn{2}{|c|}{$F_\eta\uparrow$}&\multicolumn{2}{|c|}{$MAE\downarrow$}&\multicolumn{2}{|c|}{$S_\lambda\uparrow$}
&\multicolumn{2}{|c|}{$F_\eta\uparrow$}&\multicolumn{2}{|c|}{$MAE\downarrow$}&\multicolumn{2}{|c|}{$S_\lambda\uparrow$}
&\multicolumn{2}{|c|}{$F_\eta\uparrow$}&\multicolumn{2}{|c|}{$MAE\downarrow$}&\multicolumn{2}{|c|}{$S_\lambda\uparrow$}
&\multicolumn{2}{|c|}{$F_\eta\uparrow$}&\multicolumn{2}{|c|}{$MAE\downarrow$}&\multicolumn{2}{|c|}{$S_\lambda\uparrow$}
\\
\hline
\hline
\multicolumn{4}{|c|}{\textbf{Ours+}}
&\multicolumn{2}{|c|}{\textcolor[rgb]{1,0,0}{0.813}}&\multicolumn{2}{|c|}{\textcolor[rgb]{1,0,0}{0.0732}}&\multicolumn{2}{|c|}{\textcolor[rgb]{1,0,0}{0.851}}
&\multicolumn{2}{|c|}{\textcolor[rgb]{1,0,0}{0.925}}&\multicolumn{2}{|c|}{\textcolor[rgb]{1,0,0}{0.0442}}&\multicolumn{2}{|c|}{\textcolor[rgb]{1,0,0}{0.913}}
&\multicolumn{2}{|c|}{\textcolor[rgb]{1,0,0}{0.886}}&\multicolumn{2}{|c|}{\textcolor[rgb]{1,0,0}{0.0447}}&\multicolumn{2}{|c|}{\textcolor[rgb]{1,0,0}{0.888}}
&\multicolumn{2}{|c|}{\textcolor[rgb]{1,0,0}{0.822}}&\multicolumn{2}{|c|}{\textcolor[rgb]{1,0,0}{0.1012}}&\multicolumn{2}{|c|}{\textcolor[rgb]{1,0,0}{0.804}}
\\
\multicolumn{4}{|c|}{\textbf{Ours}~\cite{zhang2018salient}}
&\multicolumn{2}{|c|}{0.772}&\multicolumn{2}{|c|}{0.1044}&\multicolumn{2}{|c|}{0.809}
&\multicolumn{2}{|c|}{\textcolor[rgb]{0,1,0}{0.913}}&\multicolumn{2}{|c|}{\textcolor[rgb]{0,1,0}{0.0484}}&\multicolumn{2}{|c|}{\textcolor[rgb]{0,1,0}{0.905}}
&\multicolumn{2}{|c|}{\textcolor[rgb]{0,1,0}{0.871}}&\multicolumn{2}{|c|}{\textcolor[rgb]{0,1,0}{0.0478}}&\multicolumn{2}{|c|}{\textcolor[rgb]{0,1,0}{0.870}}
&\multicolumn{2}{|c|}{0.789}&\multicolumn{2}{|c|}{0.1233}&\multicolumn{2}{|c|}{0.772}
\\
\hline
\hline
\multicolumn{4}{|c|}{\textbf{AMU}~\cite{zhang2017amulet}}
&\multicolumn{2}{|c|}{0.768}&\multicolumn{2}{|c|}{0.0982}&\multicolumn{2}{|c|}{0.820}
&\multicolumn{2}{|c|}{0.892}&\multicolumn{2}{|c|}{0.0602}&\multicolumn{2}{|c|}{0.893}
&\multicolumn{2}{|c|}{\textcolor[rgb]{0,0,1}{0.830}}&\multicolumn{2}{|c|}{\textcolor[rgb]{0,0,1}{0.0620}}&\multicolumn{2}{|c|}{\textcolor[rgb]{0,0,1}{0.852}}
&\multicolumn{2}{|c|}{0.743}&\multicolumn{2}{|c|}{0.1440}&\multicolumn{2}{|c|}{0.753}
\\
\multicolumn{4}{|c|}{\textbf{BDMP}~\cite{zhang2018bi}}
&\multicolumn{2}{|c|}{0.770}&\multicolumn{2}{|c|}{\textcolor[rgb]{0,1,0}{0.0740}}&\multicolumn{2}{|c|}{\textcolor[rgb]{0,0,1}{0.845}}
&\multicolumn{2}{|c|}{0.867}&\multicolumn{2}{|c|}{0.0607}&\multicolumn{2}{|c|}{0.879}
&\multicolumn{2}{|c|}{0.789}&\multicolumn{2}{|c|}{0.0790}&\multicolumn{2}{|c|}{0.819}
&\multicolumn{2}{|c|}{0.763}&\multicolumn{2}{|c|}{0.1068}&\multicolumn{2}{|c|}{\textcolor[rgb]{0,0,1}{0.787}}
\\
\multicolumn{4}{|c|}{\textbf{DCL}$^{*}$~\cite{li2016deep}}
&\multicolumn{2}{|c|}{0.714}&\multicolumn{2}{|c|}{0.1807}&\multicolumn{2}{|c|}{0.791}
&\multicolumn{2}{|c|}{0.855}&\multicolumn{2}{|c|}{0.1513}&\multicolumn{2}{|c|}{0.845}
&\multicolumn{2}{|c|}{0.795}&\multicolumn{2}{|c|}{0.1565}&\multicolumn{2}{|c|}{0.760}
&\multicolumn{2}{|c|}{0.741}&\multicolumn{2}{|c|}{0.1942}&\multicolumn{2}{|c|}{0.748}
\\
\multicolumn{4}{|c|}{\textbf{DGRL}~\cite{wang2018detect}}
&\multicolumn{2}{|c|}{\textcolor[rgb]{0,1,0}{0.811}}&\multicolumn{2}{|c|}{\textcolor[rgb]{0,0,1}{0.0744}}&\multicolumn{2}{|c|}{0.839}
&\multicolumn{2}{|c|}{\textcolor[rgb]{0,0,1}{0.904}}&\multicolumn{2}{|c|}{\textcolor[rgb]{0,0,1}{0.0527}}&\multicolumn{2}{|c|}{0.891}
&\multicolumn{2}{|c|}{0.806}&\multicolumn{2}{|c|}{0.0735}&\multicolumn{2}{|c|}{0.799}
&\multicolumn{2}{|c|}{\textcolor[rgb]{0,1,0}{0.800}}&\multicolumn{2}{|c|}{\textcolor[rgb]{0,0,1}{0.1050}}&\multicolumn{2}{|c|}{0.774}
\\
\multicolumn{4}{|c|}{\textbf{DHS}~\cite{liu2016dhsnet}}
&\multicolumn{2}{|c|}{0.777}&\multicolumn{2}{|c|}{0.0950}&\multicolumn{2}{|c|}{0.807}
&\multicolumn{2}{|c|}{0.888}&\multicolumn{2}{|c|}{0.0552}&\multicolumn{2}{|c|}{0.894}
&\multicolumn{2}{|c|}{0.822}&\multicolumn{2}{|c|}{0.0798}&\multicolumn{2}{|c|}{0.796}
&\multicolumn{2}{|c|}{0.775}&\multicolumn{2}{|c|}{0.1288}&\multicolumn{2}{|c|}{0.750}
\\
\multicolumn{4}{|c|}{\textbf{DS}$^{*}$~\cite{li2016deepsaliency}}
&\multicolumn{2}{|c|}{0.659}&\multicolumn{2}{|c|}{0.1760}&\multicolumn{2}{|c|}{0.739}
&\multicolumn{2}{|c|}{0.845}&\multicolumn{2}{|c|}{0.0931}&\multicolumn{2}{|c|}{0.859}
&\multicolumn{2}{|c|}{0.754}&\multicolumn{2}{|c|}{0.1233}&\multicolumn{2}{|c|}{0.776}
&\multicolumn{2}{|c|}{0.698}&\multicolumn{2}{|c|}{0.1896}&\multicolumn{2}{|c|}{0.712}
\\
\multicolumn{4}{|c|}{\textbf{DSS}$^{*}$~\cite{hou2017deeply}}
&\multicolumn{2}{|c|}{\textcolor[rgb]{0,0,1}{0.810}}&\multicolumn{2}{|c|}{0.0957}&\multicolumn{2}{|c|}{0.796}
&\multicolumn{2}{|c|}{-}&\multicolumn{2}{|c|}{-}&\multicolumn{2}{|c|}{-}
&\multicolumn{2}{|c|}{-}&\multicolumn{2}{|c|}{-}&\multicolumn{2}{|c|}{-}
&\multicolumn{2}{|c|}{0.788}&\multicolumn{2}{|c|}{0.1229}&\multicolumn{2}{|c|}{0.744}
\\
\multicolumn{4}{|c|}{\textbf{ELD}$^{*}$~\cite{lee2016deep}}
&\multicolumn{2}{|c|}{0.718}&\multicolumn{2}{|c|}{0.1232}&\multicolumn{2}{|c|}{0.757}
&\multicolumn{2}{|c|}{0.872}&\multicolumn{2}{|c|}{0.0670}&\multicolumn{2}{|c|}{0.864}
&\multicolumn{2}{|c|}{0.759}&\multicolumn{2}{|c|}{0.1028}&\multicolumn{2}{|c|}{0.769}
&\multicolumn{2}{|c|}{0.712}&\multicolumn{2}{|c|}{0.1550}&\multicolumn{2}{|c|}{0.705}
\\
\multicolumn{4}{|c|}{\textbf{LEGS}$^{*}$~\cite{wang2015deep}}
&\multicolumn{2}{|c|}{--}&\multicolumn{2}{|c|}{--}&\multicolumn{2}{|c|}{--}
&\multicolumn{2}{|c|}{0.854}&\multicolumn{2}{|c|}{0.1034}&\multicolumn{2}{|c|}{0.828}
&\multicolumn{2}{|c|}{0.736}&\multicolumn{2}{|c|}{0.1236}&\multicolumn{2}{|c|}{0.716}
&\multicolumn{2}{|c|}{0.683}&\multicolumn{2}{|c|}{0.1962}&\multicolumn{2}{|c|}{0.657}
\\
\multicolumn{4}{|c|}{\textbf{MCDL}$^{*}$~\cite{zhao2015saliency}}
&\multicolumn{2}{|c|}{0.691}&\multicolumn{2}{|c|}{0.1453}&\multicolumn{2}{|c|}{0.719}
&\multicolumn{2}{|c|}{0.878}&\multicolumn{2}{|c|}{0.0773}&\multicolumn{2}{|c|}{0.855}
&\multicolumn{2}{|c|}{0.757}&\multicolumn{2}{|c|}{0.1163}&\multicolumn{2}{|c|}{0.742}
&\multicolumn{2}{|c|}{0.677}&\multicolumn{2}{|c|}{0.1808}&\multicolumn{2}{|c|}{0.650}
\\
\multicolumn{4}{|c|}{\textbf{MDF}$^{*}$~\cite{li2015visual}}
&\multicolumn{2}{|c|}{0.709}&\multicolumn{2}{|c|}{0.1458}&\multicolumn{2}{|c|}{0.692}
&\multicolumn{2}{|c|}{0.842}&\multicolumn{2}{|c|}{0.0989}&\multicolumn{2}{|c|}{0.833}
&\multicolumn{2}{|c|}{0.800}&\multicolumn{2}{|c|}{0.1014}&\multicolumn{2}{|c|}{0.772}
&\multicolumn{2}{|c|}{0.721}&\multicolumn{2}{|c|}{0.1647}&\multicolumn{2}{|c|}{0.674}
\\
\multicolumn{4}{|c|}{\textbf{NLDF}~\cite{luo2017non}}
&\multicolumn{2}{|c|}{0.779}&\multicolumn{2}{|c|}{0.0991}&\multicolumn{2}{|c|}{0.803}
&\multicolumn{2}{|c|}{--}&\multicolumn{2}{|c|}{--}&\multicolumn{2}{|c|}{--}
&\multicolumn{2}{|c|}{--}&\multicolumn{2}{|c|}{--}&\multicolumn{2}{|c|}{--}
&\multicolumn{2}{|c|}{0.791}&\multicolumn{2}{|c|}{0.1242}&\multicolumn{2}{|c|}{0.756}
\\
\multicolumn{4}{|c|}{\textbf{PAGRN}~\cite{zhang2018progressive}}
&\multicolumn{2}{|c|}{0.807}&\multicolumn{2}{|c|}{0.0926}&\multicolumn{2}{|c|}{0.818}
&\multicolumn{2}{|c|}{0.861}&\multicolumn{2}{|c|}{0.0948}&\multicolumn{2}{|c|}{0.822}
&\multicolumn{2}{|c|}{0.794}&\multicolumn{2}{|c|}{0.1027}&\multicolumn{2}{|c|}{0.757}
&\multicolumn{2}{|c|}{0.771}&\multicolumn{2}{|c|}{0.1458}&\multicolumn{2}{|c|}{0.716}
\\
\multicolumn{4}{|c|}{\textbf{PICA}$^{*}$~\cite{liu2018picanet}}
&\multicolumn{2}{|c|}{0.801}&\multicolumn{2}{|c|}{0.0768}&\multicolumn{2}{|c|}{\textcolor[rgb]{0,1,0}{0.850}}
&\multicolumn{2}{|c|}{--}&\multicolumn{2}{|c|}{--}&\multicolumn{2}{|c|}{--}
&\multicolumn{2}{|c|}{--}&\multicolumn{2}{|c|}{--}&\multicolumn{2}{|c|}{--}
&\multicolumn{2}{|c|}{0.791}&\multicolumn{2}{|c|}{\textcolor[rgb]{0,1,0}{0.1025}}&\multicolumn{2}{|c|}{\textcolor[rgb]{0,1,0}{0.790}}
\\
\multicolumn{4}{|c|}{\textbf{RFCN}$^{*}$~\cite{wang2016saliency}}
&\multicolumn{2}{|c|}{0.751}&\multicolumn{2}{|c|}{0.1324}&\multicolumn{2}{|c|}{0.799}
&\multicolumn{2}{|c|}{0.850}&\multicolumn{2}{|c|}{0.1166}&\multicolumn{2}{|c|}{0.832}
&\multicolumn{2}{|c|}{0.767}&\multicolumn{2}{|c|}{0.1131}&\multicolumn{2}{|c|}{0.771}
&\multicolumn{2}{|c|}{0.743}&\multicolumn{2}{|c|}{0.1697}&\multicolumn{2}{|c|}{0.730}
\\
\multicolumn{4}{|c|}{\textbf{RST}$^{*}$~\cite{zhu2017saliency}}
&\multicolumn{2}{|c|}{0.771}&\multicolumn{2}{|c|}{0.1127}&\multicolumn{2}{|c|}{0.796}
&\multicolumn{2}{|c|}{--}&\multicolumn{2}{|c|}{--}&\multicolumn{2}{|c|}{--}
&\multicolumn{2}{|c|}{--}&\multicolumn{2}{|c|}{--}&\multicolumn{2}{|c|}{--}
&\multicolumn{2}{|c|}{0.762}&\multicolumn{2}{|c|}{0.1441}&\multicolumn{2}{|c|}{0.747}
\\
\multicolumn{4}{|c|}{\textbf{SRM}$^{*}$~\cite{wang2017stagewise}}
&\multicolumn{2}{|c|}{0.801}&\multicolumn{2}{|c|}{0.0850}&\multicolumn{2}{|c|}{0.832}
&\multicolumn{2}{|c|}{0.885}&\multicolumn{2}{|c|}{0.0753}&\multicolumn{2}{|c|}{0.854}
&\multicolumn{2}{|c|}{0.817}&\multicolumn{2}{|c|}{0.0914}&\multicolumn{2}{|c|}{0.761}
&\multicolumn{2}{|c|}{\textcolor[rgb]{0,1,0}{0.800}}&\multicolumn{2}{|c|}{0.1265}&\multicolumn{2}{|c|}{0.742}
\\
\multicolumn{4}{|c|}{\textbf{UCF}~\cite{zhang2017learning}}
&\multicolumn{2}{|c|}{0.735}&\multicolumn{2}{|c|}{0.1149}&\multicolumn{2}{|c|}{0.806}
&\multicolumn{2}{|c|}{0.865}&\multicolumn{2}{|c|}{0.0631}&\multicolumn{2}{|c|}{\textcolor[rgb]{0,0,1}{0.896}}
&\multicolumn{2}{|c|}{0.810}&\multicolumn{2}{|c|}{0.0680}&\multicolumn{2}{|c|}{0.846}
&\multicolumn{2}{|c|}{0.738}&\multicolumn{2}{|c|}{0.1478}&\multicolumn{2}{|c|}{0.762}
\\
\hline
\hline
\multicolumn{4}{|c|}{\textbf{BL}~\cite{tong2015salient}}
&\multicolumn{2}{|c|}{0.574}&\multicolumn{2}{|c|}{0.2487}&\multicolumn{2}{|c|}{0.647}
&\multicolumn{2}{|c|}{0.780}&\multicolumn{2}{|c|}{0.1849}&\multicolumn{2}{|c|}{0.783}
&\multicolumn{2}{|c|}{0.713}&\multicolumn{2}{|c|}{0.1856}&\multicolumn{2}{|c|}{0.705}
&\multicolumn{2}{|c|}{0.580}&\multicolumn{2}{|c|}{0.2670}&\multicolumn{2}{|c|}{0.625}
\\
\multicolumn{4}{|c|}{\textbf{BSCA}~\cite{qin2015saliency}}
&\multicolumn{2}{|c|}{0.601}&\multicolumn{2}{|c|}{0.2229}&\multicolumn{2}{|c|}{0.652}
&\multicolumn{2}{|c|}{0.805}&\multicolumn{2}{|c|}{0.1535}&\multicolumn{2}{|c|}{0.785}
&\multicolumn{2}{|c|}{0.706}&\multicolumn{2}{|c|}{0.1578}&\multicolumn{2}{|c|}{0.714}
&\multicolumn{2}{|c|}{0.584}&\multicolumn{2}{|c|}{0.2516}&\multicolumn{2}{|c|}{0.621}
\\
\multicolumn{4}{|c|}{\textbf{DRFI}~\cite{jiang2013salient}}
&\multicolumn{2}{|c|}{0.618}&\multicolumn{2}{|c|}{0.2065}&\multicolumn{2}{|c|}{0.670}
&\multicolumn{2}{|c|}{0.807}&\multicolumn{2}{|c|}{0.1480}&\multicolumn{2}{|c|}{0.797}
&\multicolumn{2}{|c|}{0.745}&\multicolumn{2}{|c|}{0.1334}&\multicolumn{2}{|c|}{0.751}
&\multicolumn{2}{|c|}{0.634}&\multicolumn{2}{|c|}{0.2240}&\multicolumn{2}{|c|}{0.624}
\\
\multicolumn{4}{|c|}{\textbf{DSR}~\cite{li2013saliency}}
&\multicolumn{2}{|c|}{0.558}&\multicolumn{2}{|c|}{0.2149}&\multicolumn{2}{|c|}{0.594}
&\multicolumn{2}{|c|}{0.791}&\multicolumn{2}{|c|}{0.1579}&\multicolumn{2}{|c|}{0.736}
&\multicolumn{2}{|c|}{0.712}&\multicolumn{2}{|c|}{0.1406}&\multicolumn{2}{|c|}{0.715}
&\multicolumn{2}{|c|}{0.596}&\multicolumn{2}{|c|}{0.2345}&\multicolumn{2}{|c|}{0.596}
\\
\hline
\end{tabular}
}
\vspace{-2mm}
\end{center}
\vspace{-2mm}
\end{table*}
\begin{figure*}
\begin{center}
\resizebox{0.8\textwidth}{!}
{
\begin{tabular}{@{}c@{}c@{}c@{}c}
\includegraphics[width=0.25\linewidth,height=4cm]{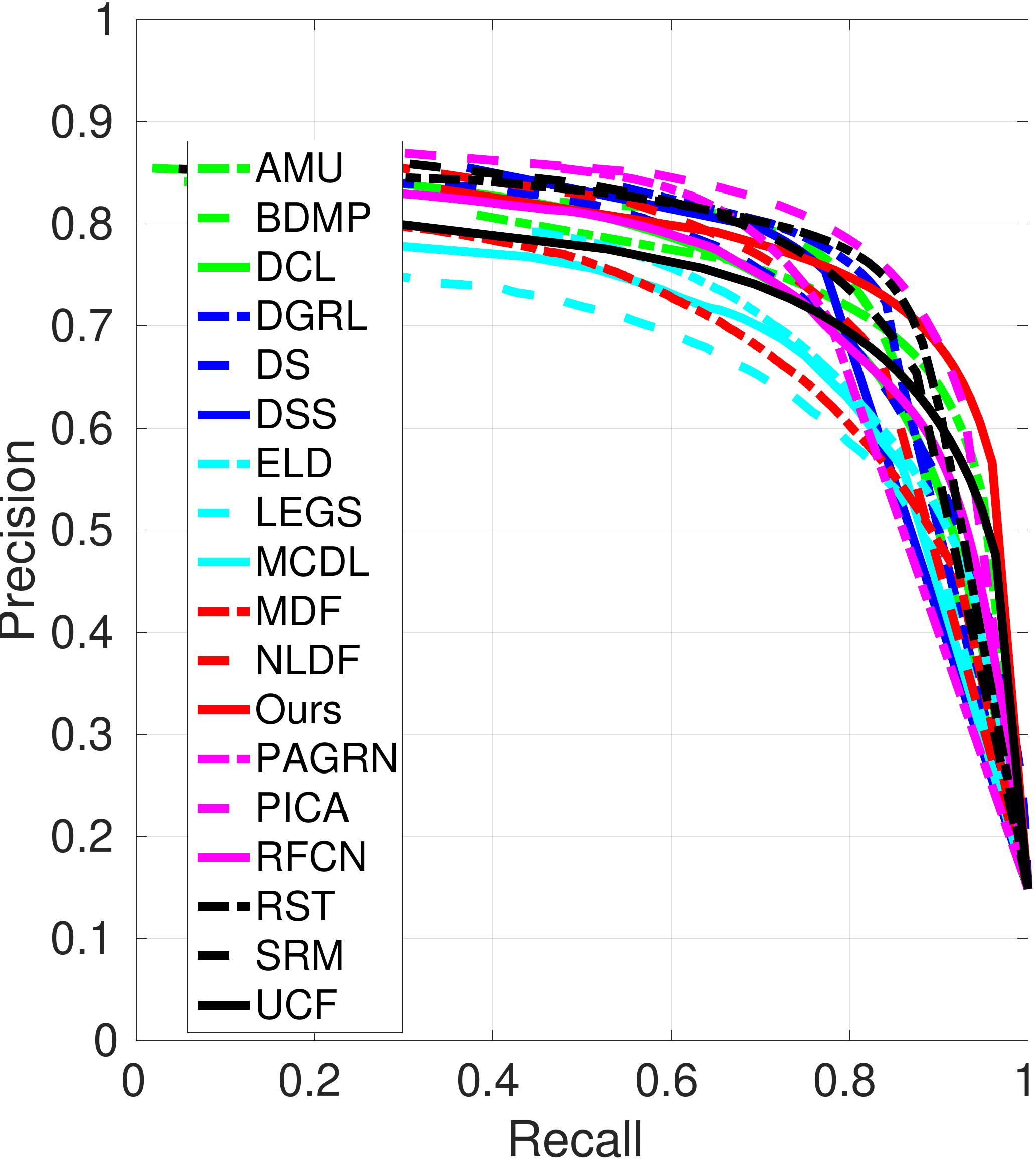} \ &
\includegraphics[width=0.25\linewidth,height=4cm]{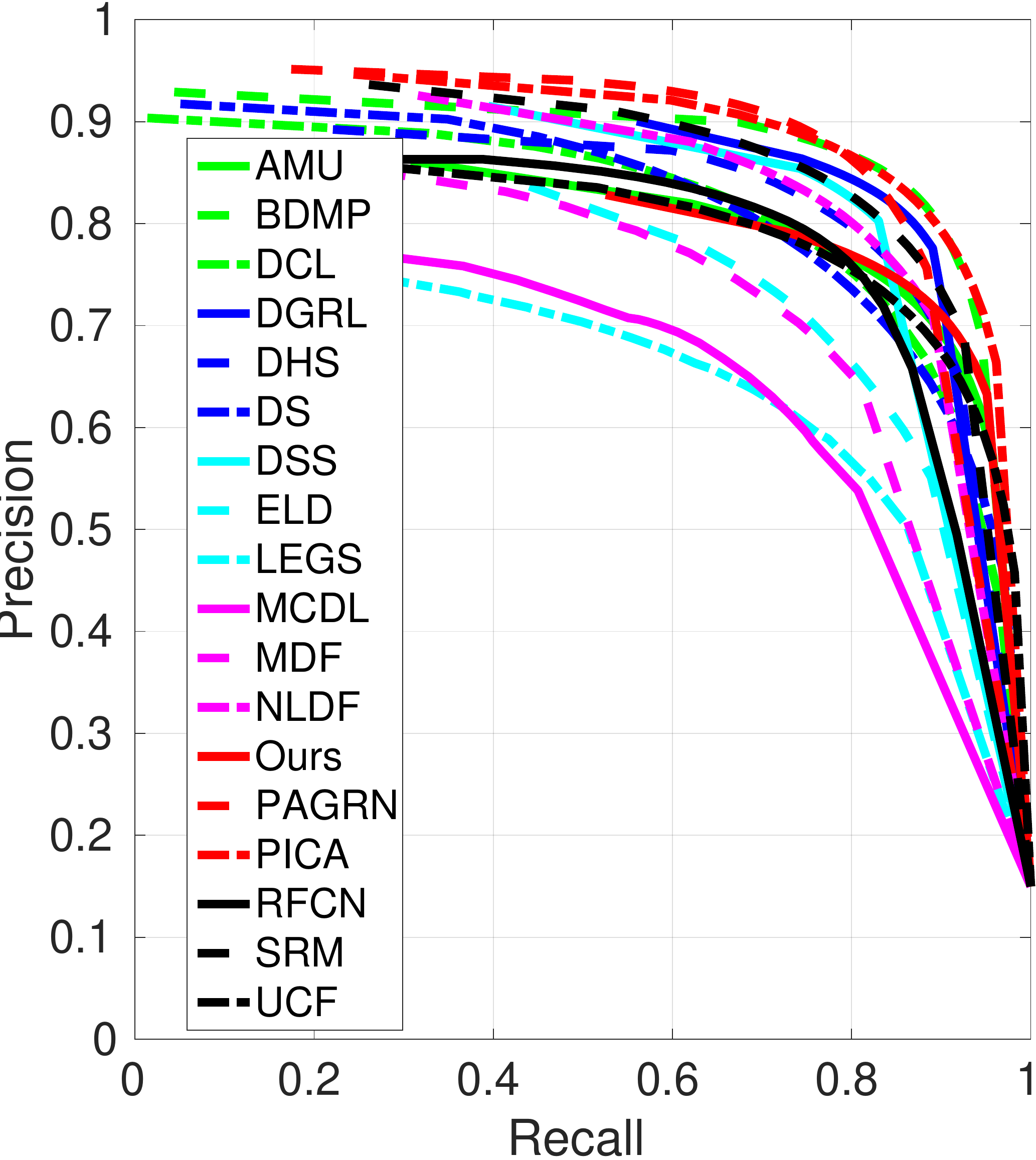} \ &
\includegraphics[width=0.25\linewidth,height=4cm]{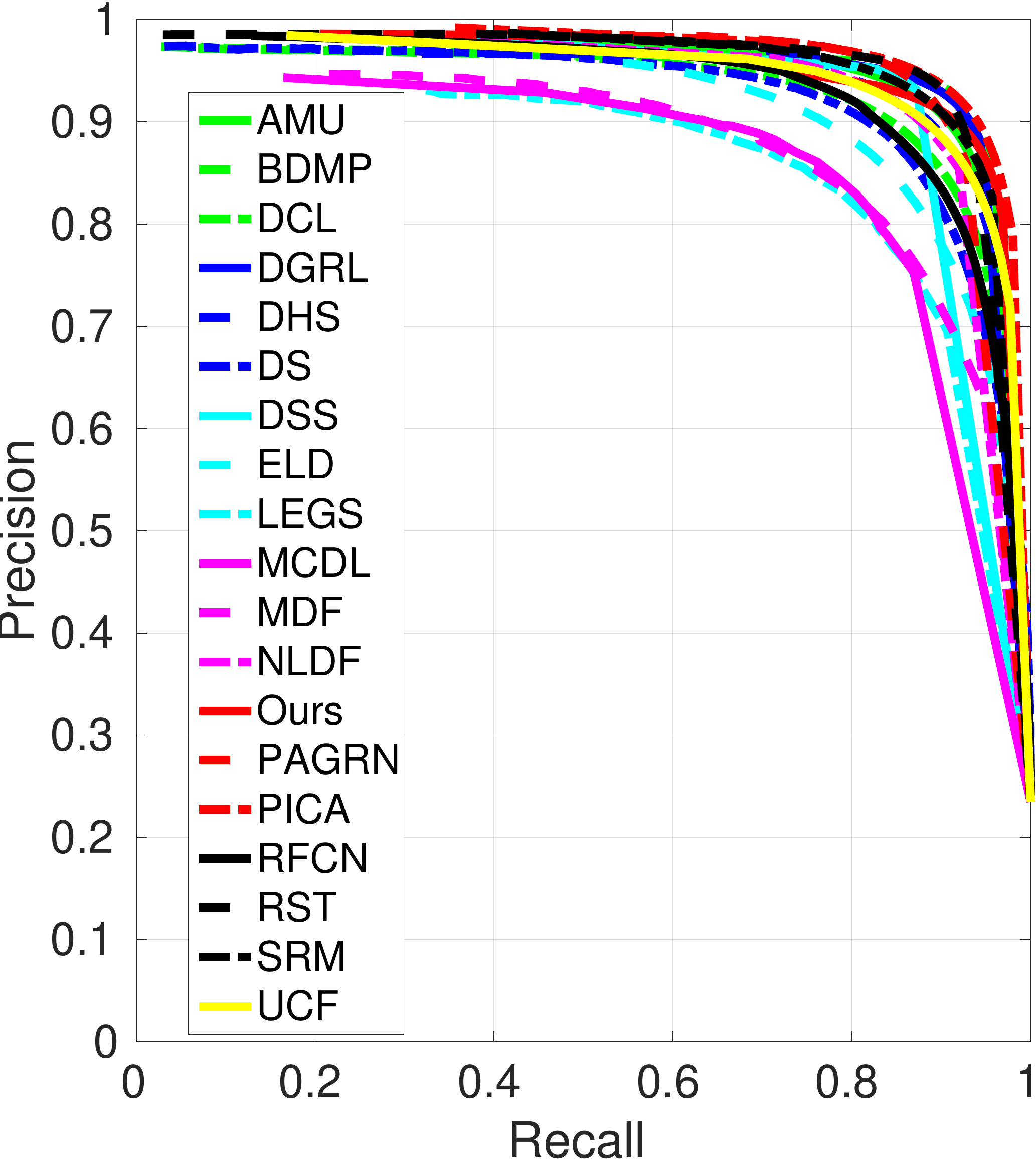} \ &
\includegraphics[width=0.25\linewidth,height=4cm]{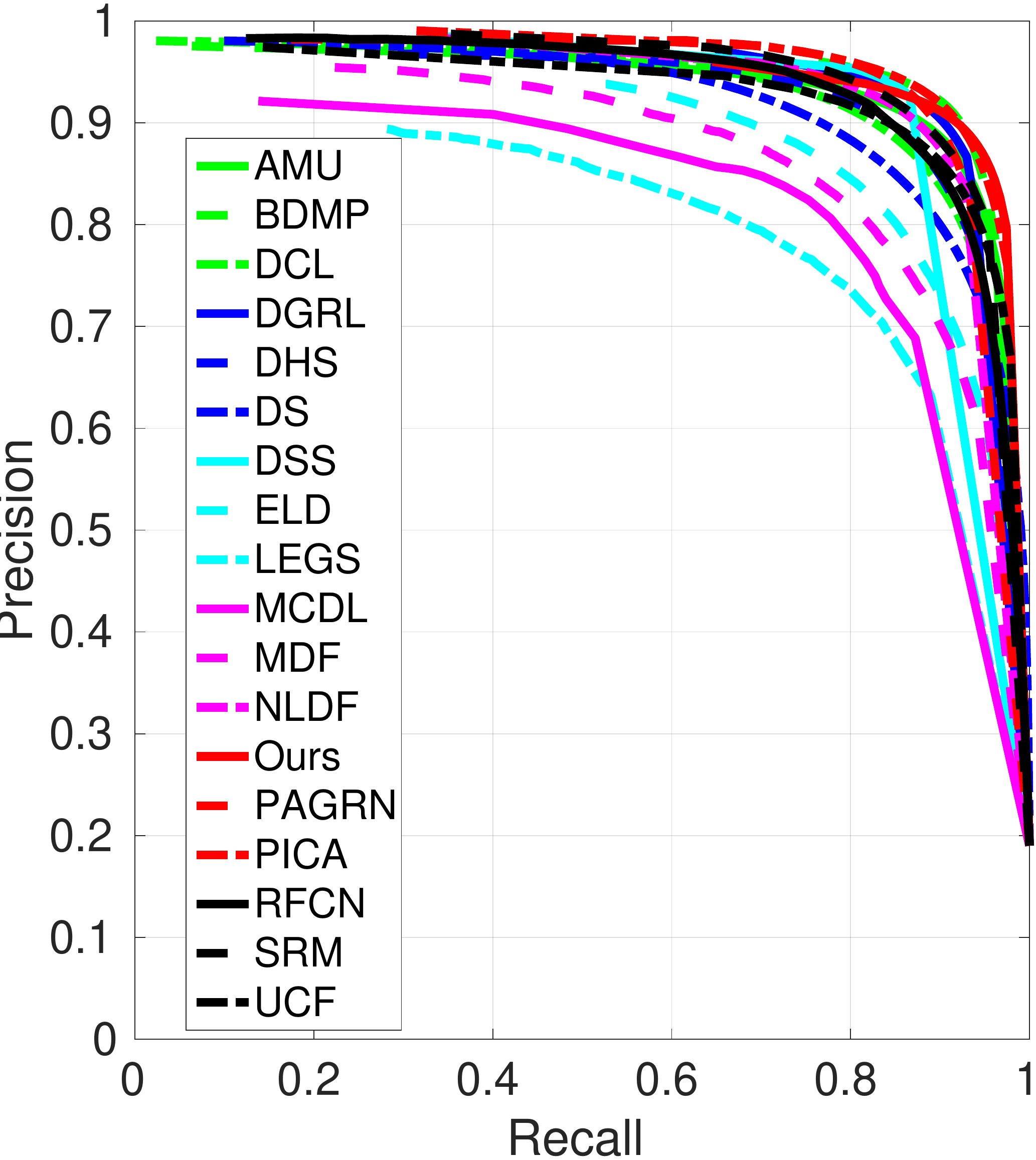} \ \\
{\small(a) \textbf{DUT-OMRON}} & {\small(b) \textbf{DUTS-TE}} & {\small(c) \textbf{ECSSD}} & {\small(d) \textbf{HKU-IS-TE}}\\
\includegraphics[width=0.25\linewidth,height=4cm]{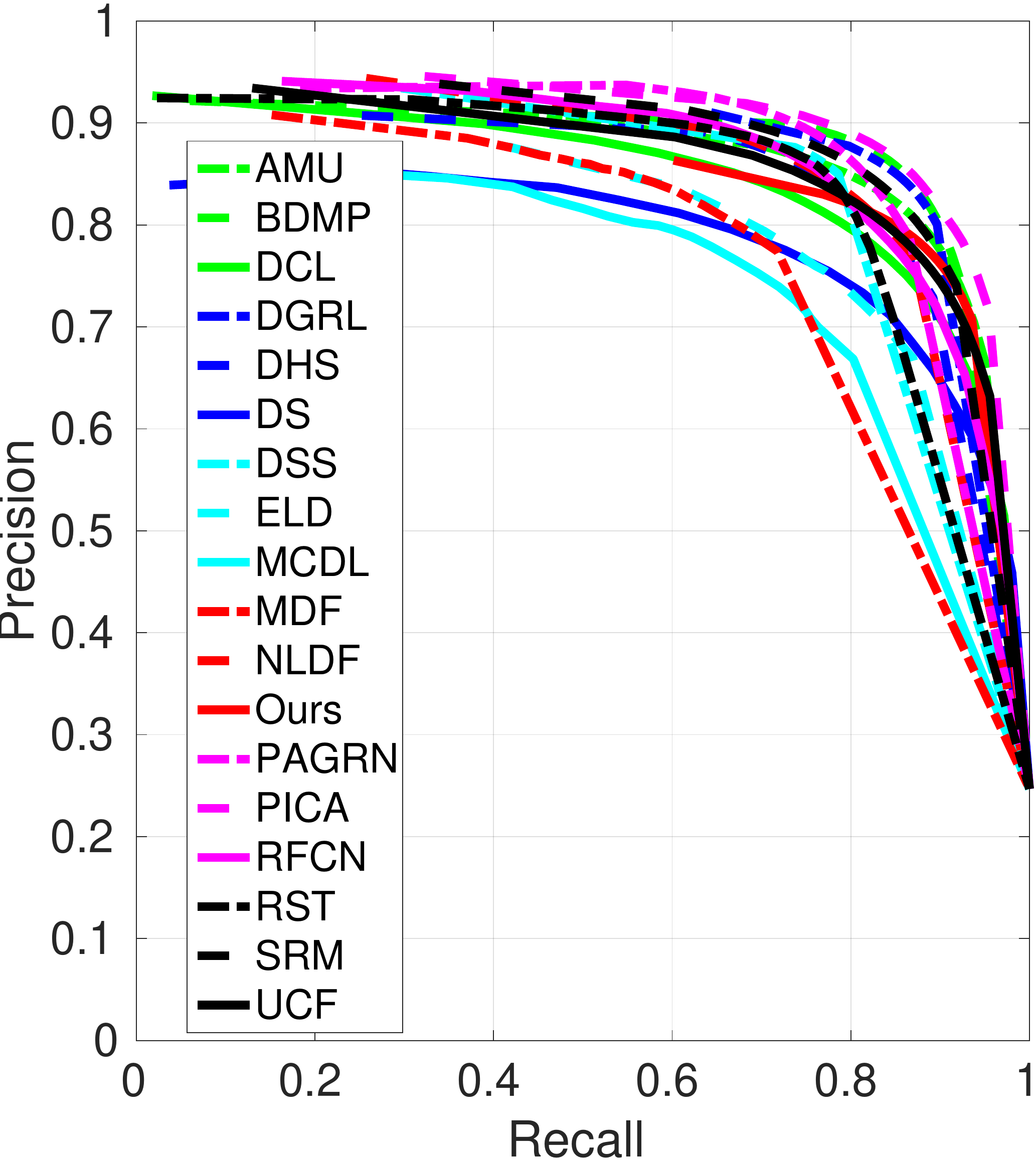} \ &
\includegraphics[width=0.25\linewidth,height=4cm]{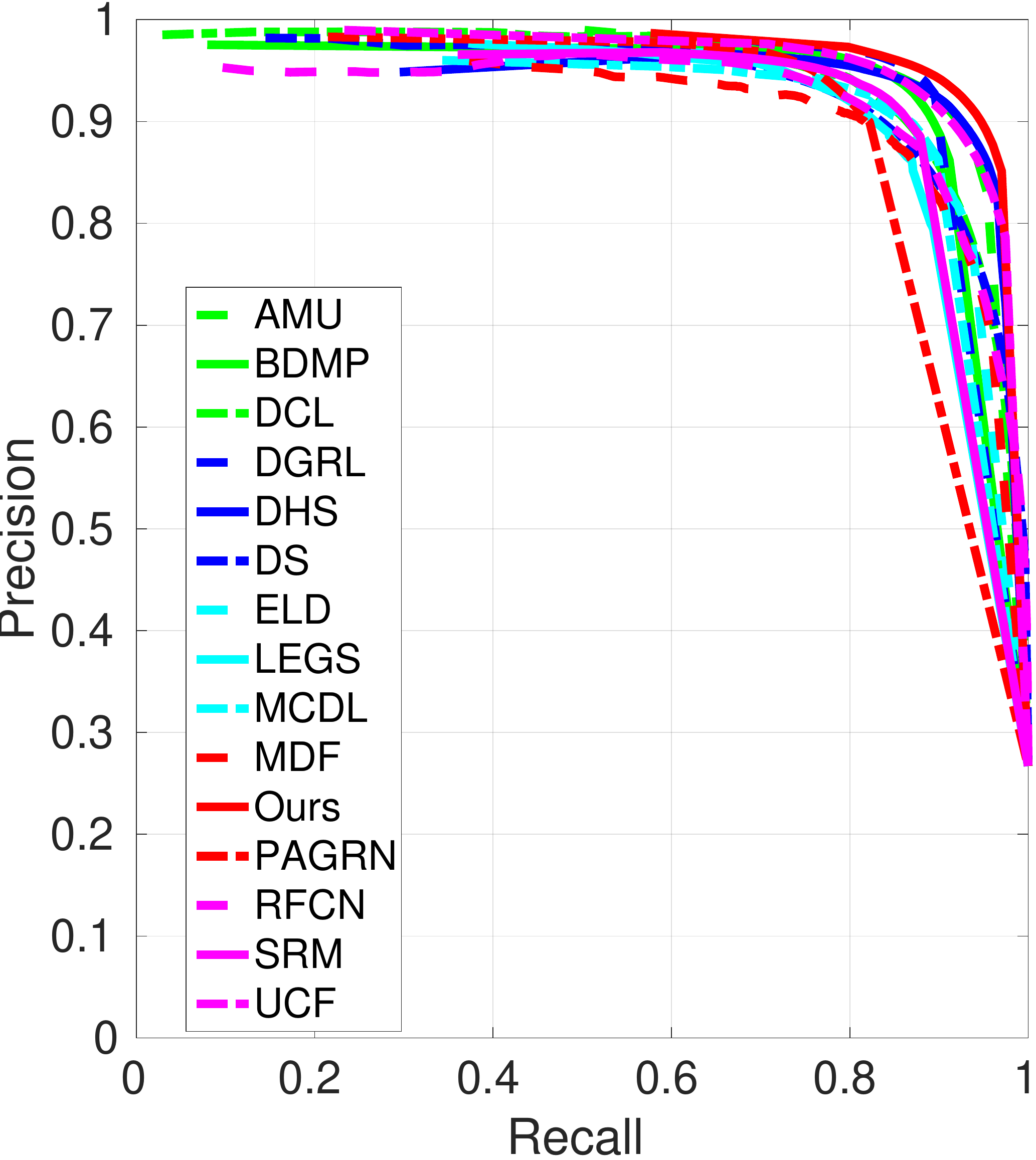} \ &
\includegraphics[width=0.25\linewidth,height=4cm]{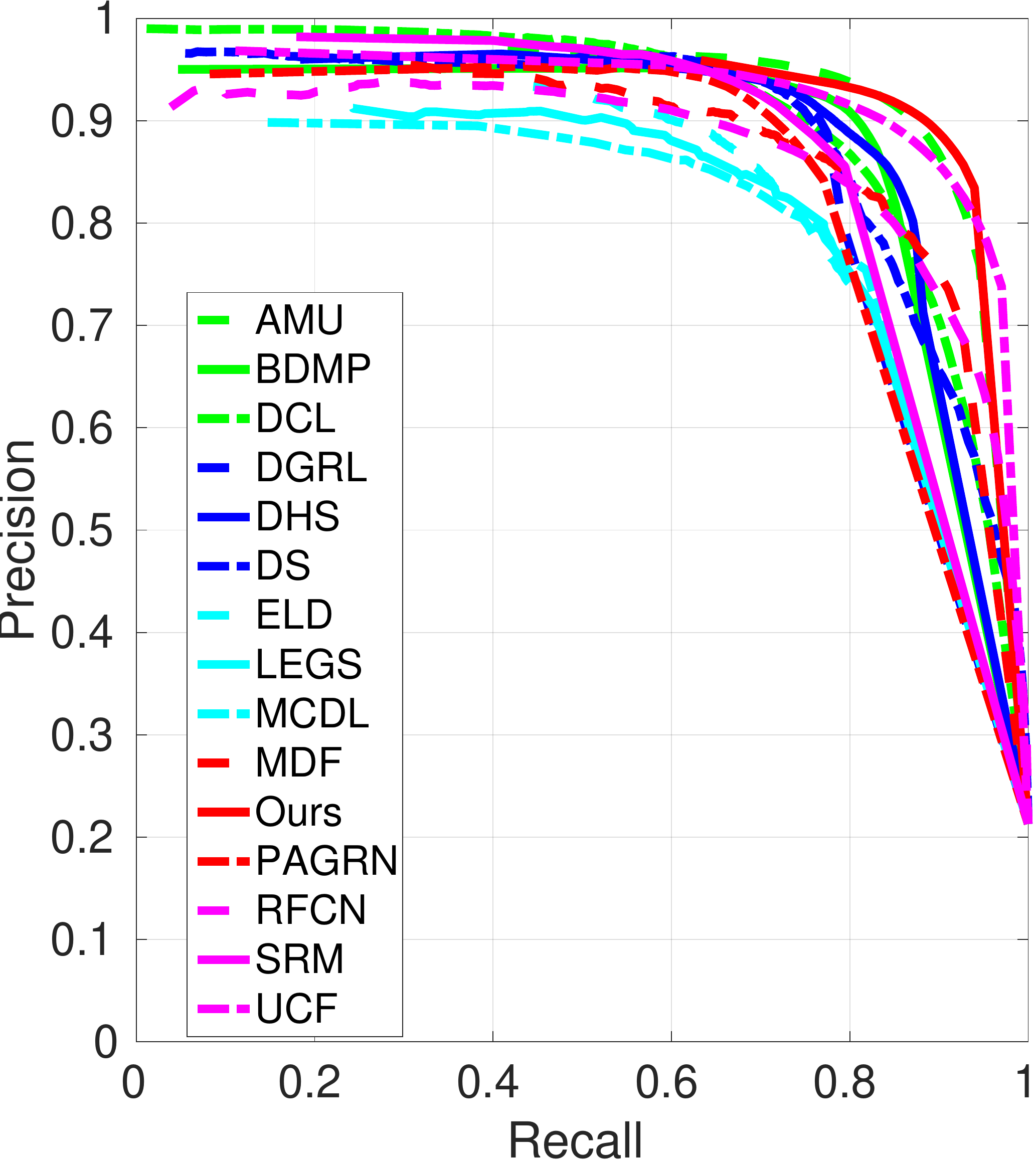} \ &
\includegraphics[width=0.25\linewidth,height=4cm]{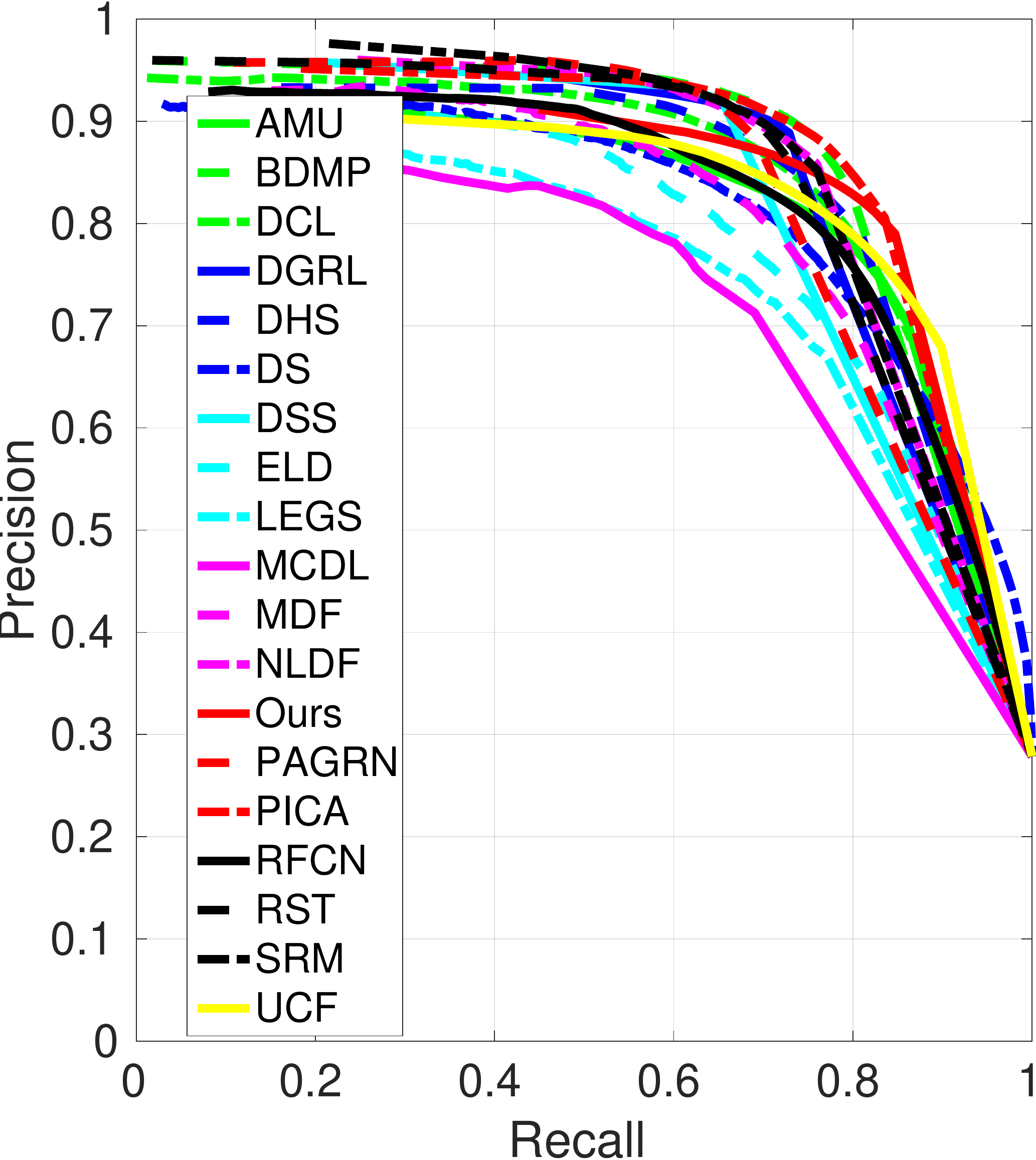} \ \\
{\small(e) \textbf{PASCAL-S}} & {\small(f) \textbf{SED1}} & {\small(g) \textbf{SED2}} & {\small(h) \textbf{SOD}}\\
\\
\end{tabular}
}
\vspace{-4mm}
\caption{The PR curves of the proposed algorithm and other 22 state-of-the-art methods. Our method performs well on these datasets.}
\label{fig:PR-curve}
\end{center}
\vspace{-4mm}
\end{figure*}
\begin{figure*}
\resizebox{0.96\textwidth}{8cm}
{
\begin{tabular}{@{}c@{}c@{}c@{}c@{}c@{}c@{}c@{}c@{}c@{}c@{}c@{}c}
\vspace{-1mm}
(a) &
\includegraphics[width=0.1\linewidth,height=1cm]{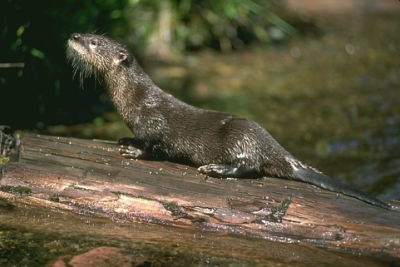}\ &
\includegraphics[width=0.1\linewidth,height=1cm]{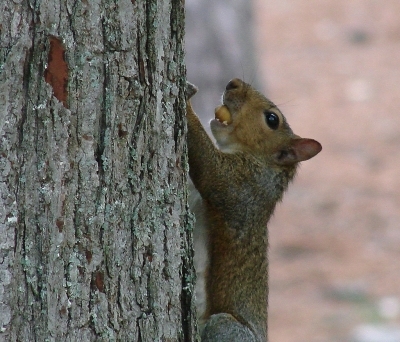}\ &
\includegraphics[width=0.1\linewidth,height=1cm]{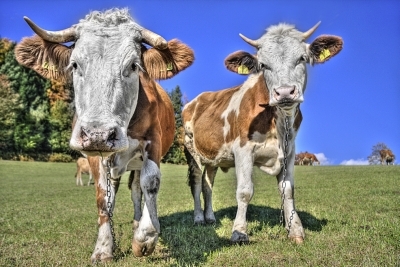}\ &
\includegraphics[width=0.1\linewidth,height=1cm]{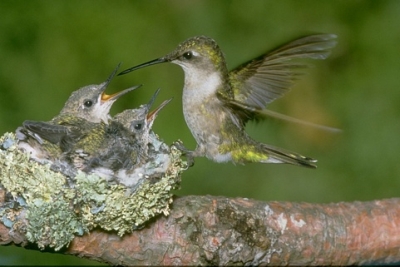}\ &
\includegraphics[width=0.1\linewidth,height=1cm]{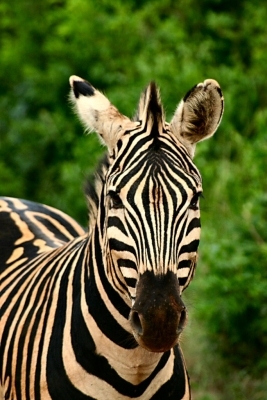}\ &
\includegraphics[width=0.1\linewidth,height=1cm]{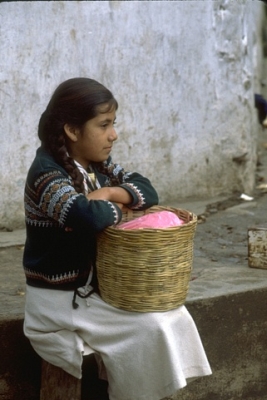}\
\\
\vspace{-1mm}
(b) &
\includegraphics[width=0.1\linewidth,height=1cm]{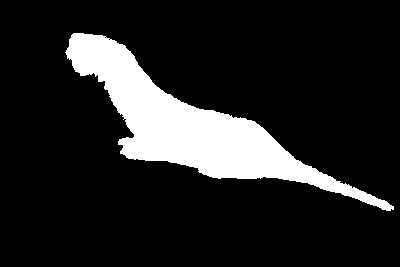}\ &
\includegraphics[width=0.1\linewidth,height=1cm]{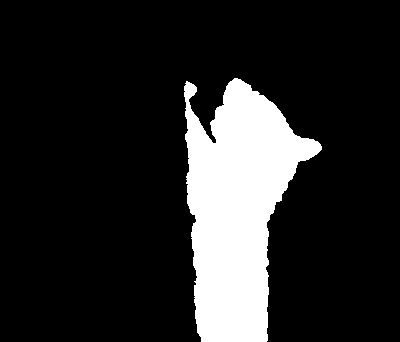}\ &
\includegraphics[width=0.1\linewidth,height=1cm]{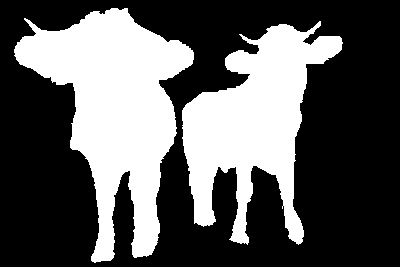}\ &
\includegraphics[width=0.1\linewidth,height=1cm]{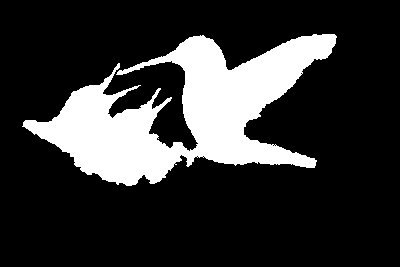}\ &
\includegraphics[width=0.1\linewidth,height=1cm]{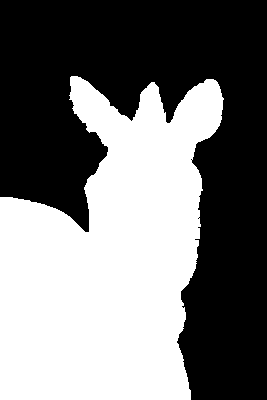}\ &
\includegraphics[width=0.1\linewidth,height=1cm]{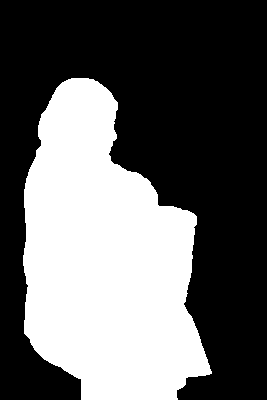}\
\\
\vspace{-1mm}
(c) &
\includegraphics[width=0.1\linewidth,height=1cm]{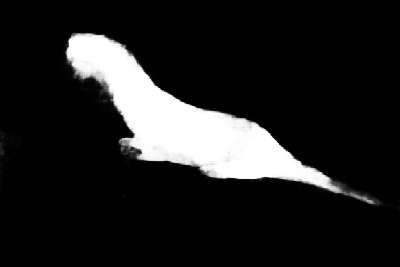}\ &
\includegraphics[width=0.1\linewidth,height=1cm]{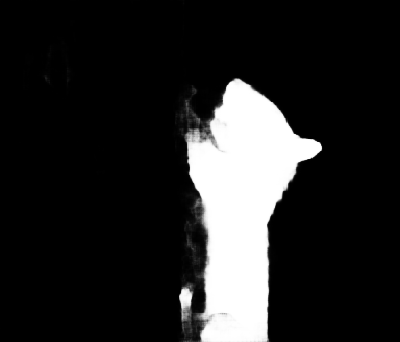}\ &
\includegraphics[width=0.1\linewidth,height=1cm]{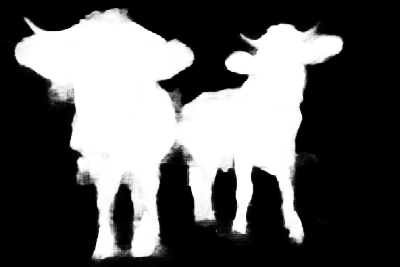}\ &
\includegraphics[width=0.1\linewidth,height=1cm]{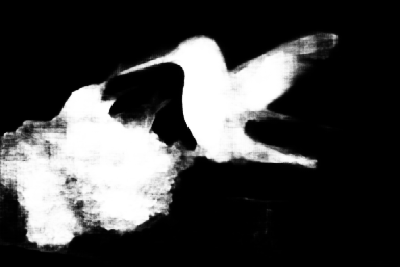}\ &
\includegraphics[width=0.1\linewidth,height=1cm]{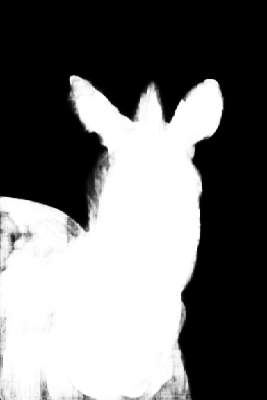}\ &
\includegraphics[width=0.1\linewidth,height=1cm]{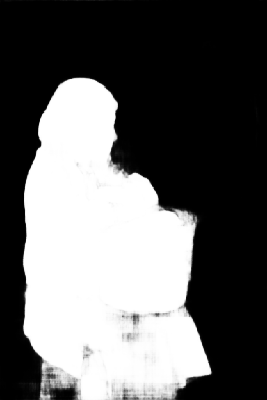}\
\\
\vspace{-1mm}
(d) &
\includegraphics[width=0.1\linewidth,height=1cm]{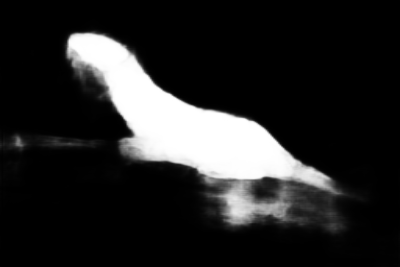}\ &
\includegraphics[width=0.1\linewidth,height=1cm]{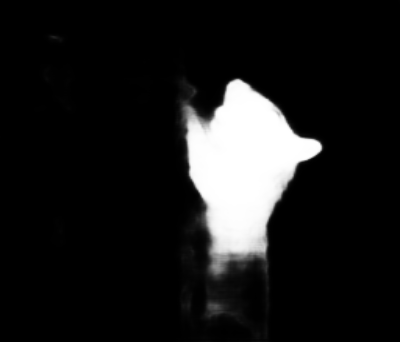}\ &
\includegraphics[width=0.1\linewidth,height=1cm]{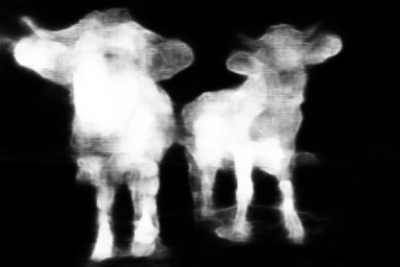}\ &
\includegraphics[width=0.1\linewidth,height=1cm]{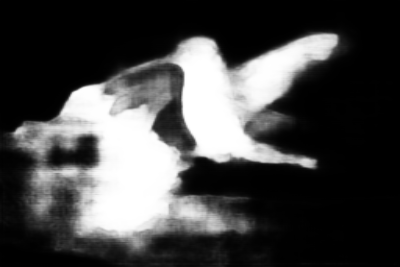}\ &
\includegraphics[width=0.1\linewidth,height=1cm]{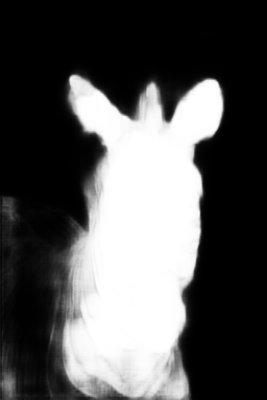}\ &
\includegraphics[width=0.1\linewidth,height=1cm]{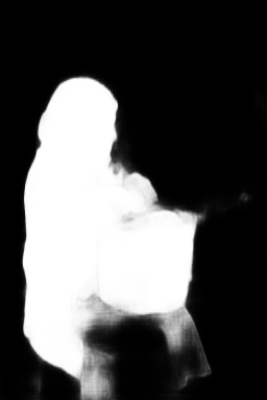}\
\\
\vspace{-1mm}
(e) &
\includegraphics[width=0.1\linewidth,height=1cm]{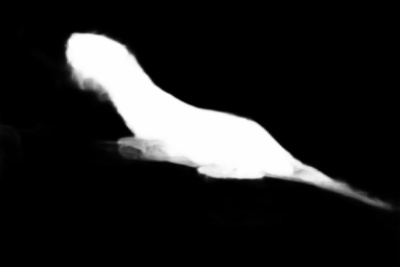}\ &
\includegraphics[width=0.1\linewidth,height=1cm]{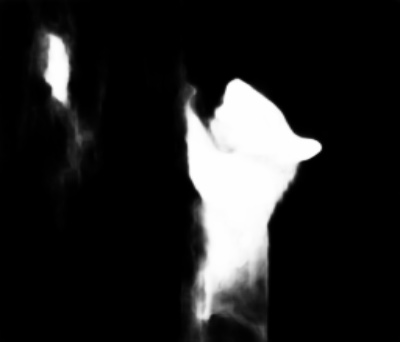}\ &
\includegraphics[width=0.1\linewidth,height=1cm]{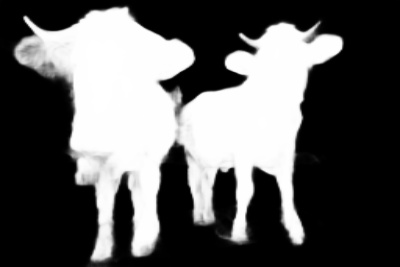}\ &
\includegraphics[width=0.1\linewidth,height=1cm]{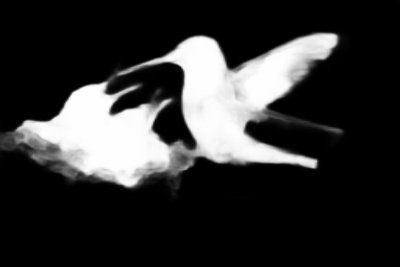}\ &
\includegraphics[width=0.1\linewidth,height=1cm]{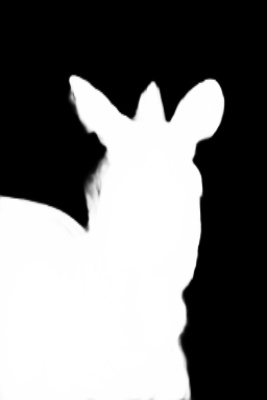}\ &
\includegraphics[width=0.1\linewidth,height=1cm]{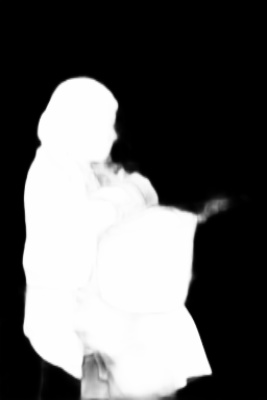}\
\\
\vspace{-1mm}
(f) &
\includegraphics[width=0.1\linewidth,height=1cm]{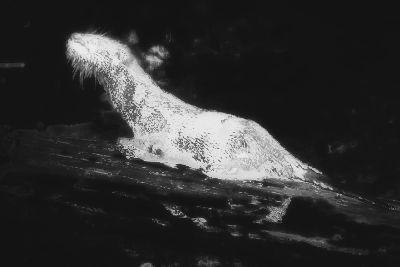}\ &
\includegraphics[width=0.1\linewidth,height=1cm]{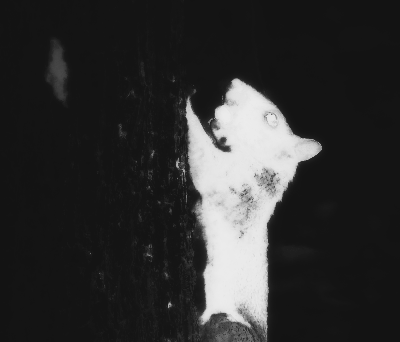}\ &
\includegraphics[width=0.1\linewidth,height=1cm]{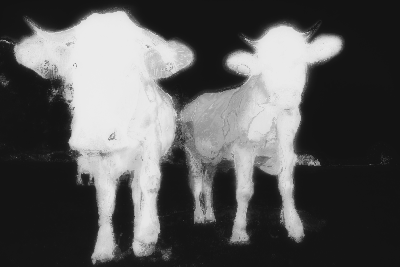}\ &
\includegraphics[width=0.1\linewidth,height=1cm]{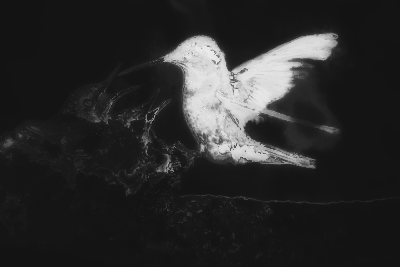}\ &
\includegraphics[width=0.1\linewidth,height=1cm]{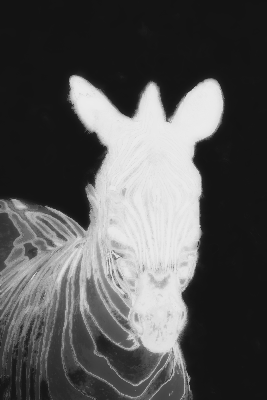}\ &
\includegraphics[width=0.1\linewidth,height=1cm]{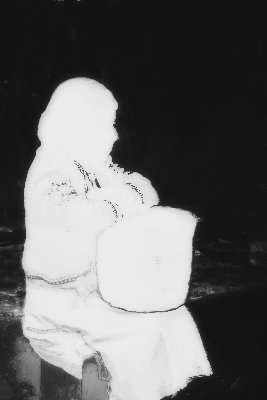}\
\\
\vspace{-1mm}
(g) &
\includegraphics[width=0.1\linewidth,height=1cm]{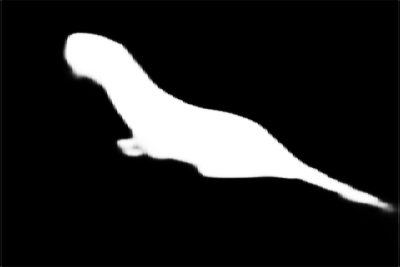}\ &
\includegraphics[width=0.1\linewidth,height=1cm]{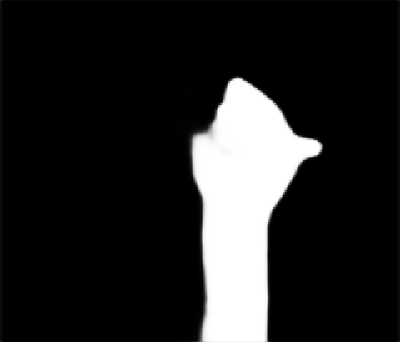}\ &
\includegraphics[width=0.1\linewidth,height=1cm]{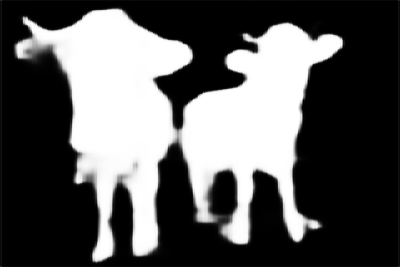}\ &
\includegraphics[width=0.1\linewidth,height=1cm]{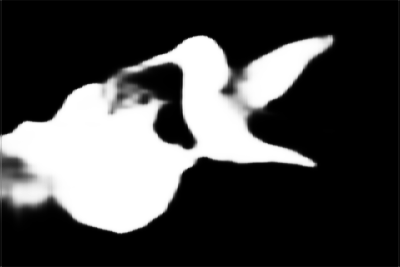}\ &
\includegraphics[width=0.1\linewidth,height=1cm]{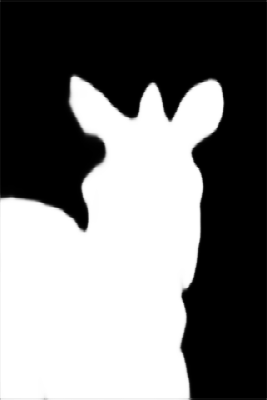}\ &
\includegraphics[width=0.1\linewidth,height=1cm]{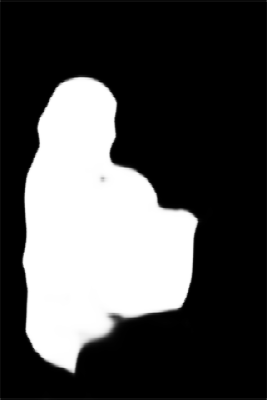}\
\\
\vspace{-1mm}
(h) &
\includegraphics[width=0.1\linewidth,height=1cm]{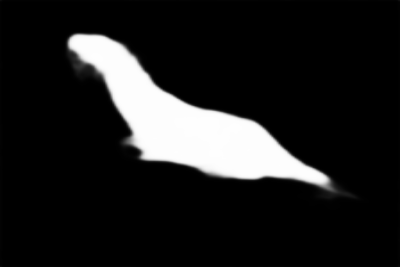}\ &
\includegraphics[width=0.1\linewidth,height=1cm]{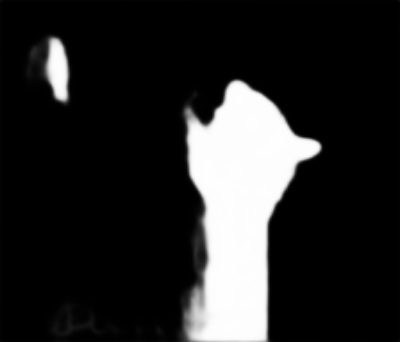}\ &
\includegraphics[width=0.1\linewidth,height=1cm]{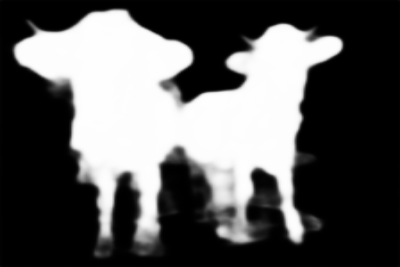}\ &
\includegraphics[width=0.1\linewidth,height=1cm]{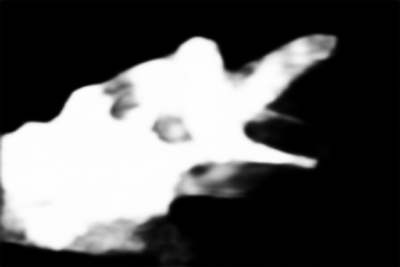}\ &
\includegraphics[width=0.1\linewidth,height=1cm]{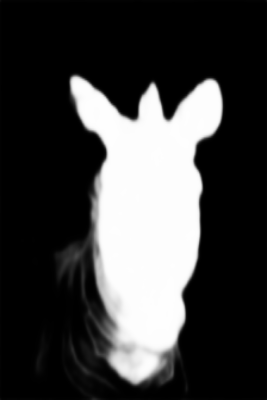}\ &
\includegraphics[width=0.1\linewidth,height=1cm]{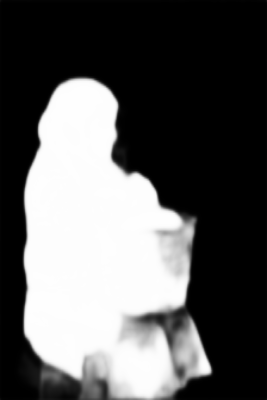}\
\\
\vspace{-1mm}
(i) &
\includegraphics[width=0.1\linewidth,height=1cm]{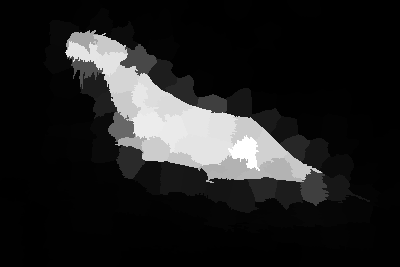}\ &
\includegraphics[width=0.1\linewidth,height=1cm]{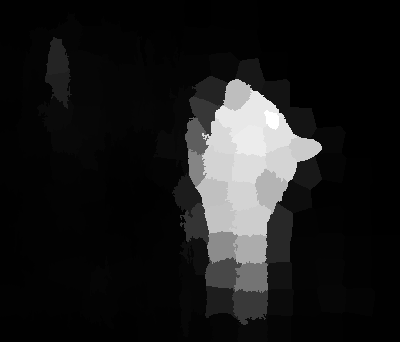}\ &
\includegraphics[width=0.1\linewidth,height=1cm]{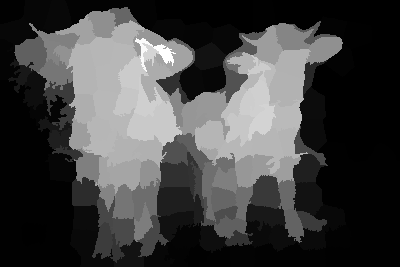}\ &
\includegraphics[width=0.1\linewidth,height=1cm]{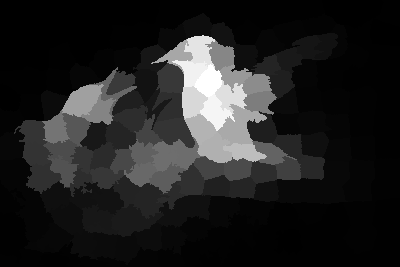}\ &
\includegraphics[width=0.1\linewidth,height=1cm]{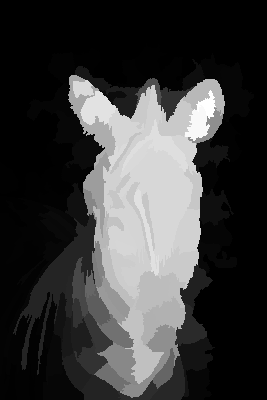}\ &
\includegraphics[width=0.1\linewidth,height=1cm]{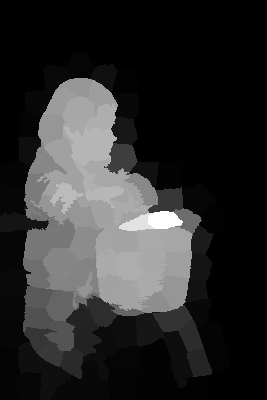}\
\\
\vspace{-1mm}
(j) &
\includegraphics[width=0.1\linewidth,height=1cm]{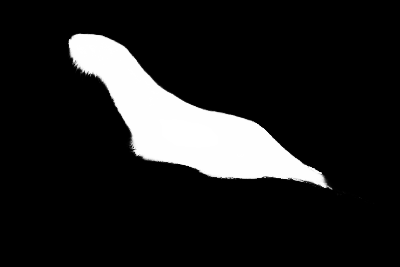}\ &
\includegraphics[width=0.1\linewidth,height=1cm]{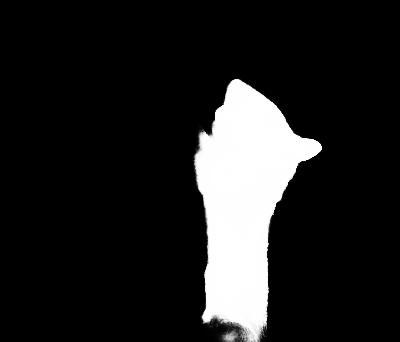}\ &
\includegraphics[width=0.1\linewidth,height=1cm]{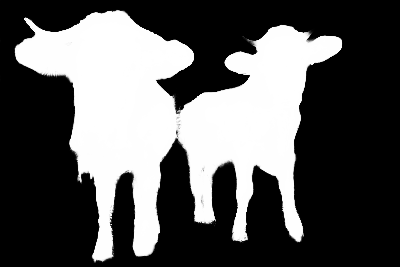}\ &
\includegraphics[width=0.1\linewidth,height=1cm]{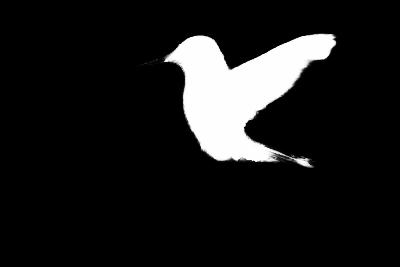}\ &
\includegraphics[width=0.1\linewidth,height=1cm]{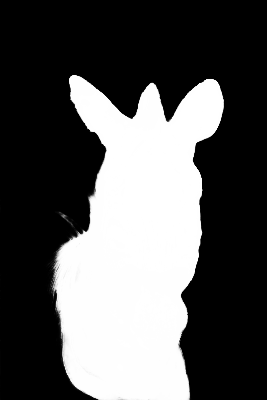}\ &
\includegraphics[width=0.1\linewidth,height=1cm]{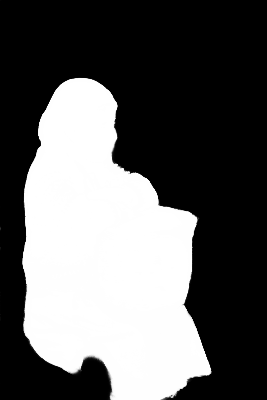}\
\\
\vspace{-1mm}
(k) &
\includegraphics[width=0.1\linewidth,height=1cm]{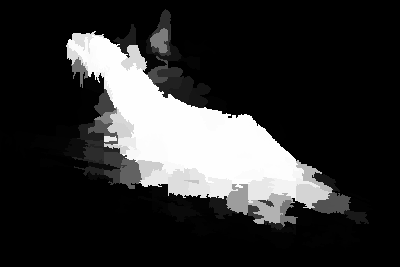}\ &
\includegraphics[width=0.1\linewidth,height=1cm]{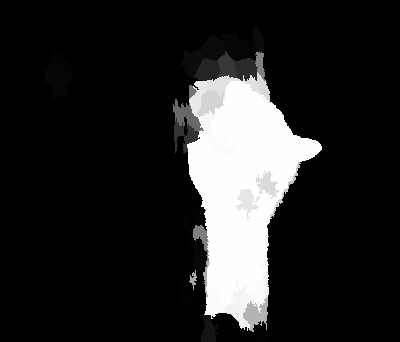}\ &
\includegraphics[width=0.1\linewidth,height=1cm]{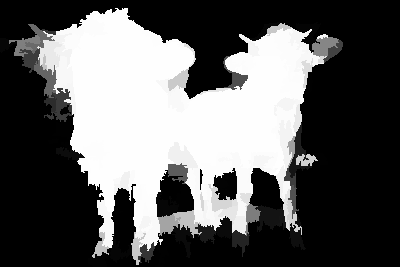}\ &
\includegraphics[width=0.1\linewidth,height=1cm]{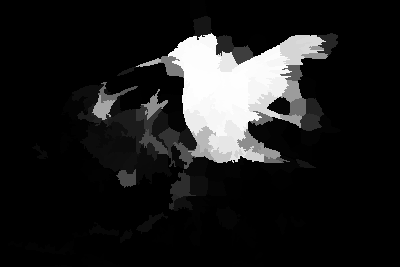}\ &
\includegraphics[width=0.1\linewidth,height=1cm]{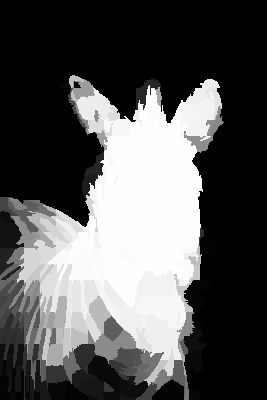}\ &
\includegraphics[width=0.1\linewidth,height=1cm]{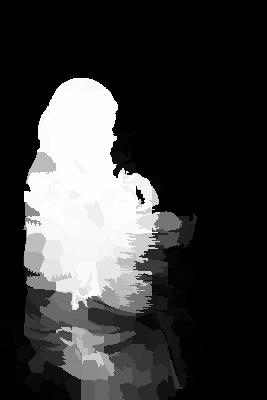}\
\\
\vspace{-1mm}
(l) &
\includegraphics[width=0.1\linewidth,height=1cm]{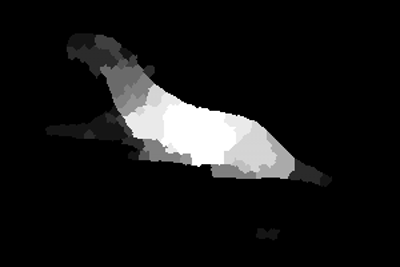}\ &
\includegraphics[width=0.1\linewidth,height=1cm]{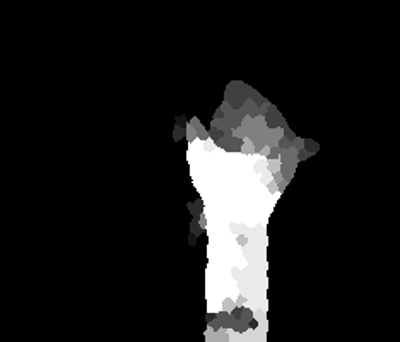}\ &
\includegraphics[width=0.1\linewidth,height=1cm]{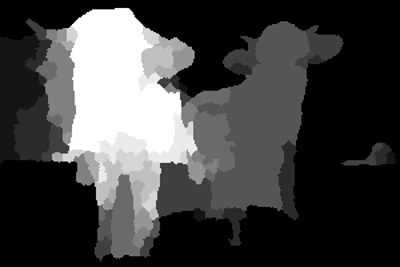}\ &
\includegraphics[width=0.1\linewidth,height=1cm]{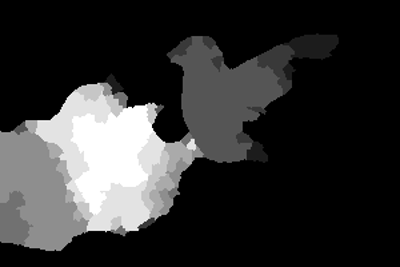}\ &
\includegraphics[width=0.1\linewidth,height=1cm]{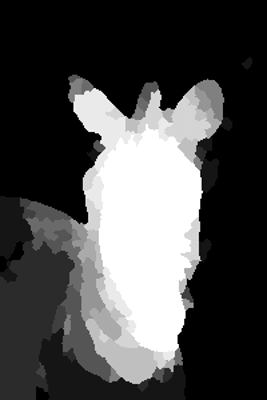}\ &
\includegraphics[width=0.1\linewidth,height=1cm]{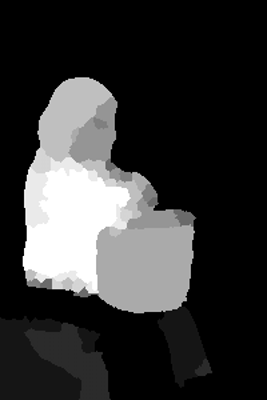}\
\\
\vspace{-1mm}
(m) &
\includegraphics[width=0.1\linewidth,height=1cm]{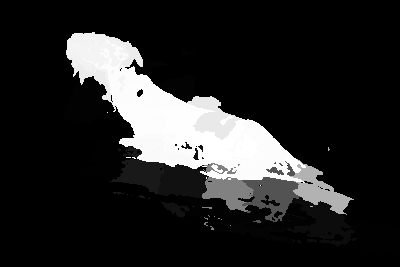}\ &
\includegraphics[width=0.1\linewidth,height=1cm]{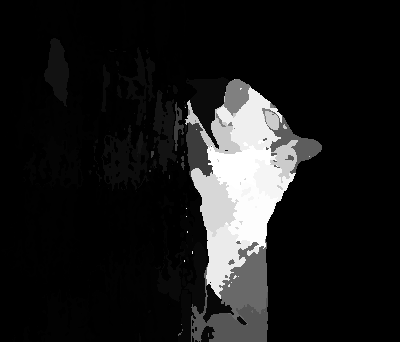}\ &
\includegraphics[width=0.1\linewidth,height=1cm]{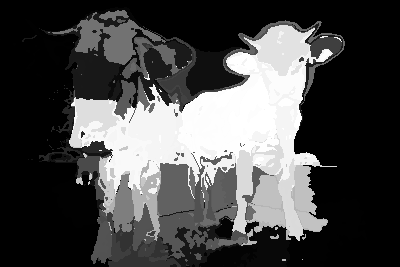}\ &
\includegraphics[width=0.1\linewidth,height=1cm]{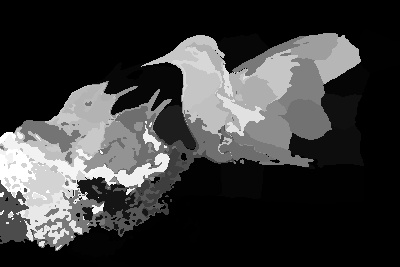}\ &
\includegraphics[width=0.1\linewidth,height=1cm]{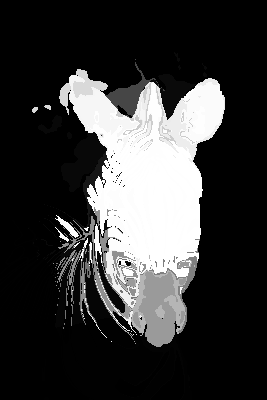}\ &
\includegraphics[width=0.1\linewidth,height=1cm]{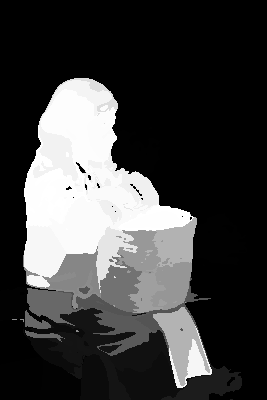}\
\\
\vspace{-1mm}
(n) &
\includegraphics[width=0.1\linewidth,height=1cm]{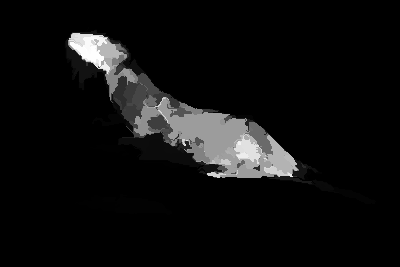}\ &
\includegraphics[width=0.1\linewidth,height=1cm]{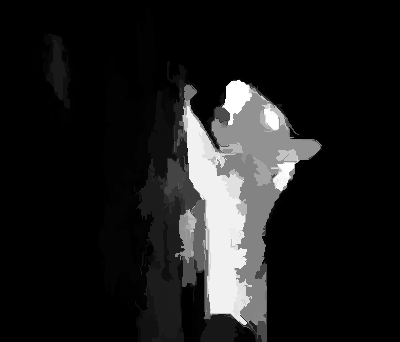}\ &
\includegraphics[width=0.1\linewidth,height=1cm]{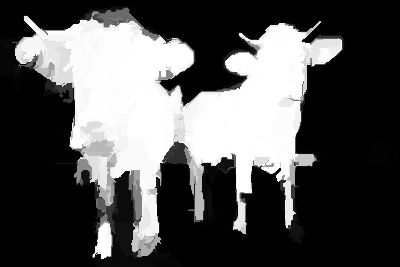}\ &
\includegraphics[width=0.1\linewidth,height=1cm]{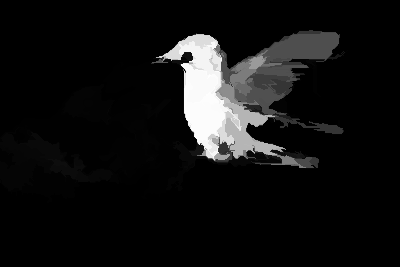}\ &
\includegraphics[width=0.1\linewidth,height=1cm]{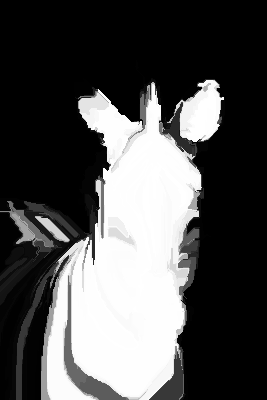}\ &
\includegraphics[width=0.1\linewidth,height=1cm]{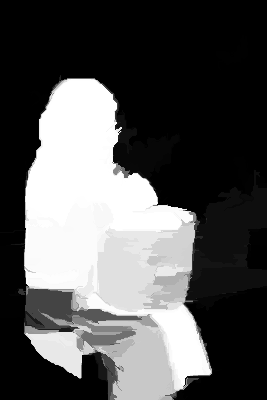}\
\\
\vspace{-1mm}
(o) &
\includegraphics[width=0.1\linewidth,height=1cm]{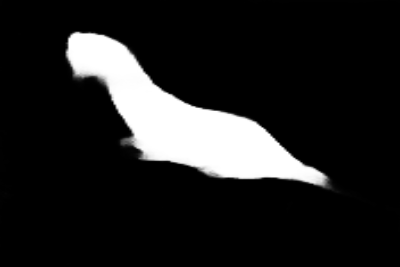}\ &
\includegraphics[width=0.1\linewidth,height=1cm]{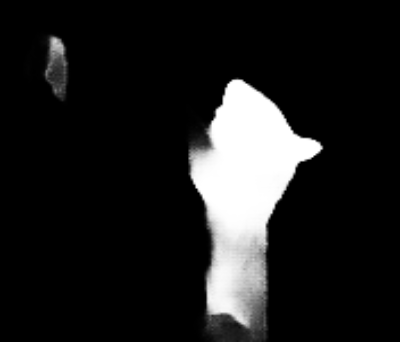}\ &
\includegraphics[width=0.1\linewidth,height=1cm]{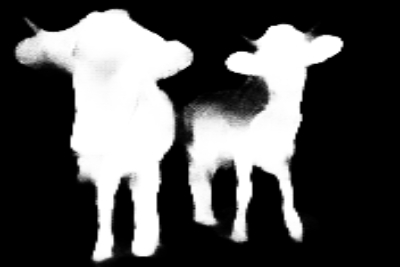}\ &
\includegraphics[width=0.1\linewidth,height=1cm]{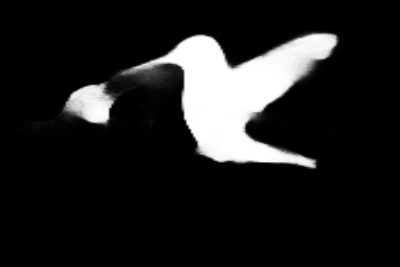}\ &
\includegraphics[width=0.1\linewidth,height=1cm]{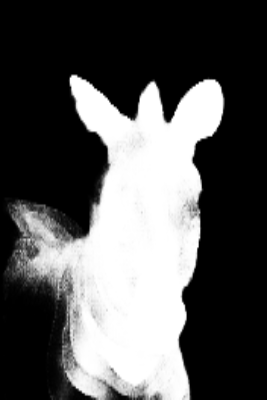}\ &
\includegraphics[width=0.1\linewidth,height=1cm]{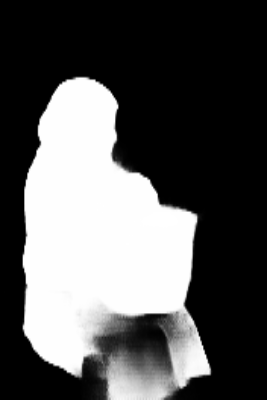}\
\\
\vspace{-1mm}
(p) &
\includegraphics[width=0.1\linewidth,height=1cm]{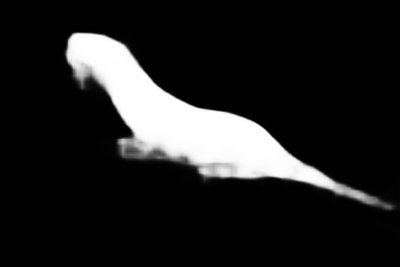}\ &
\includegraphics[width=0.1\linewidth,height=1cm]{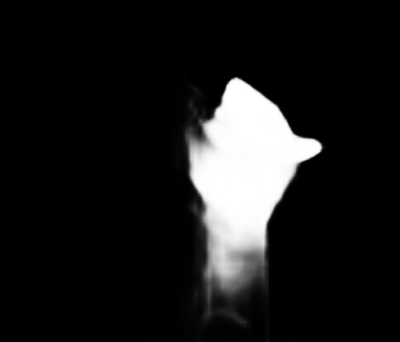}\ &
\includegraphics[width=0.1\linewidth,height=1cm]{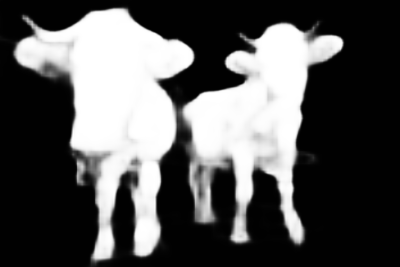}\ &
\includegraphics[width=0.1\linewidth,height=1cm]{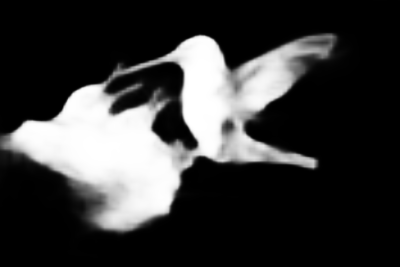}\ &
\includegraphics[width=0.1\linewidth,height=1cm]{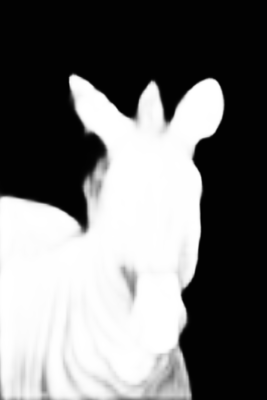}\ &
\includegraphics[width=0.1\linewidth,height=1cm]{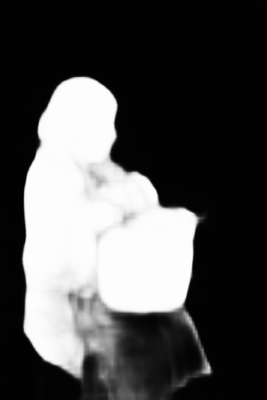}\
\\
\vspace{-1mm}
(q) &
\includegraphics[width=0.1\linewidth,height=1cm]{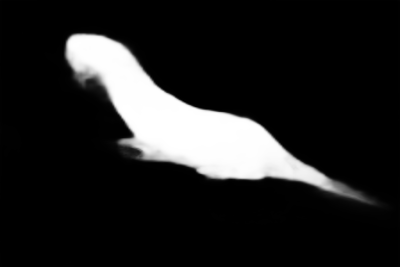}\ &
\includegraphics[width=0.1\linewidth,height=1cm]{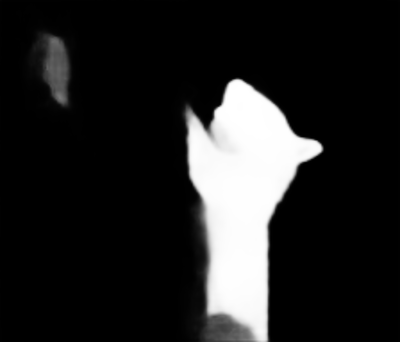}\ &
\includegraphics[width=0.1\linewidth,height=1cm]{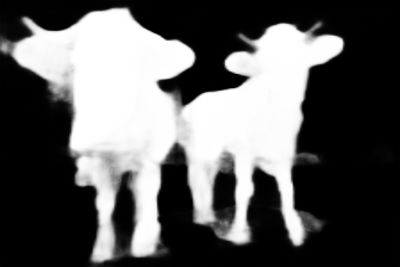}\ &
\includegraphics[width=0.1\linewidth,height=1cm]{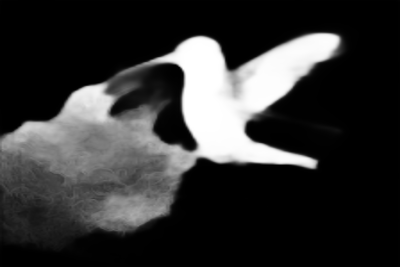}\ &
\includegraphics[width=0.1\linewidth,height=1cm]{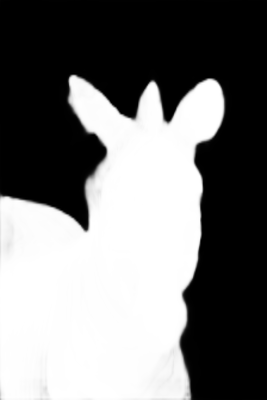}\ &
\includegraphics[width=0.1\linewidth,height=1cm]{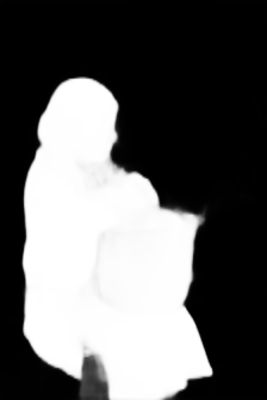}\
\\
\vspace{-1mm}
(r) &
\includegraphics[width=0.1\linewidth,height=1cm]{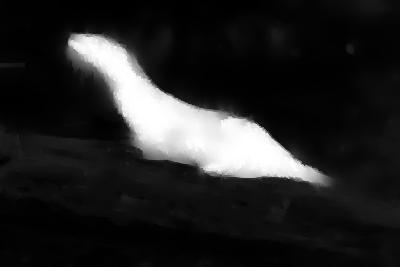}\ &
\includegraphics[width=0.1\linewidth,height=1cm]{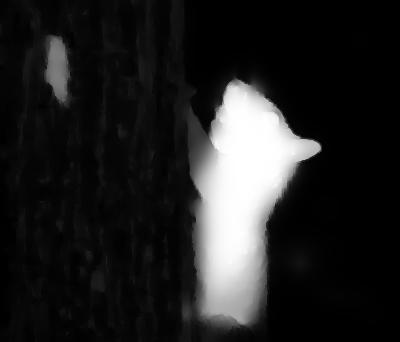}\ &
\includegraphics[width=0.1\linewidth,height=1cm]{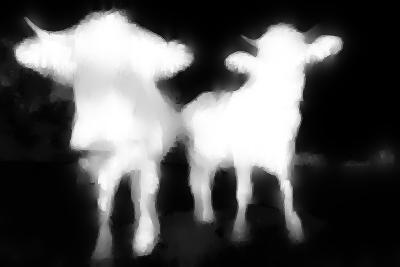}\ &
\includegraphics[width=0.1\linewidth,height=1cm]{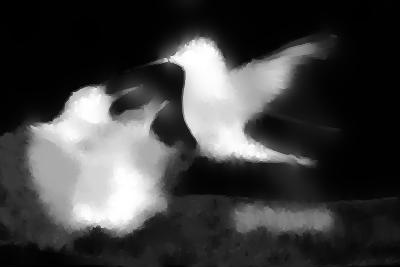}\ &
\includegraphics[width=0.1\linewidth,height=1cm]{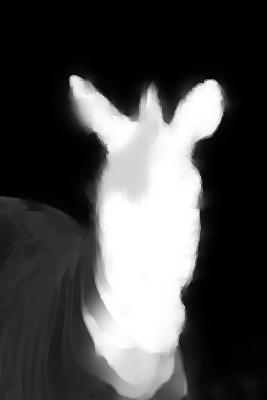}\ &
\includegraphics[width=0.1\linewidth,height=1cm]{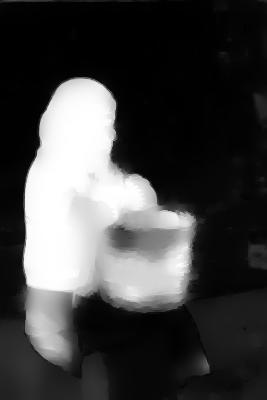}\
\\
\vspace{-1mm}
(s) &
\includegraphics[width=0.1\linewidth,height=1cm]{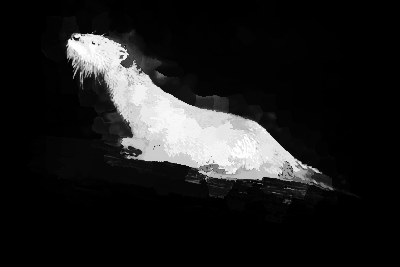}\ &
\includegraphics[width=0.1\linewidth,height=1cm]{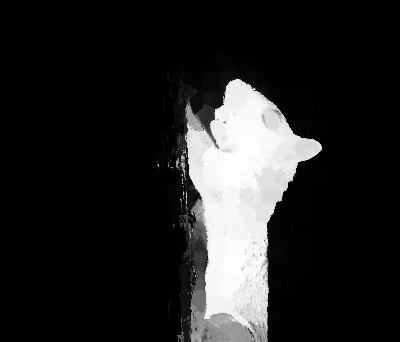}\ &
\includegraphics[width=0.1\linewidth,height=1cm]{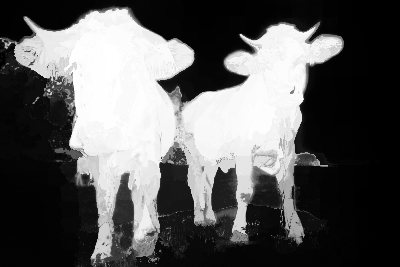}\ &
\includegraphics[width=0.1\linewidth,height=1cm]{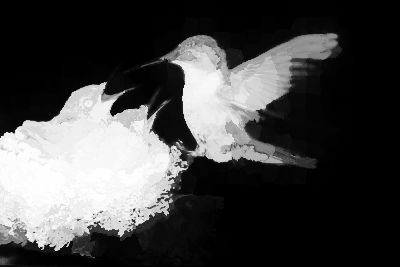}\ &
\includegraphics[width=0.1\linewidth,height=1cm]{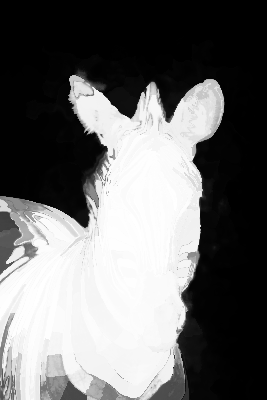}\ &
\includegraphics[width=0.1\linewidth,height=1cm]{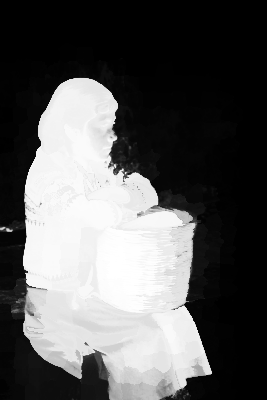}\
\\
\vspace{-1mm}
(t) &
\includegraphics[width=0.1\linewidth,height=1cm]{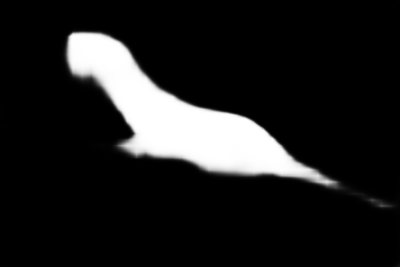}\ &
\includegraphics[width=0.1\linewidth,height=1cm]{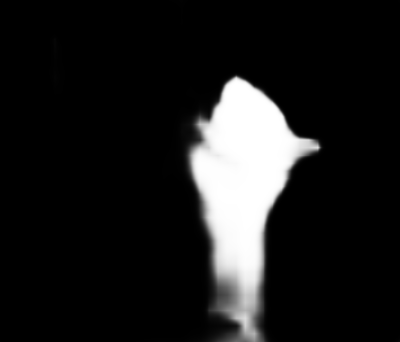}\ &
\includegraphics[width=0.1\linewidth,height=1cm]{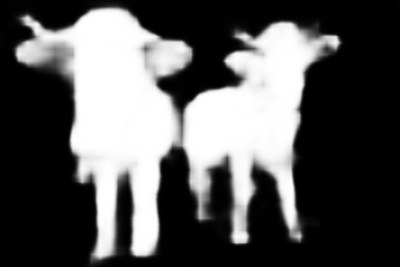}\ &
\includegraphics[width=0.1\linewidth,height=1cm]{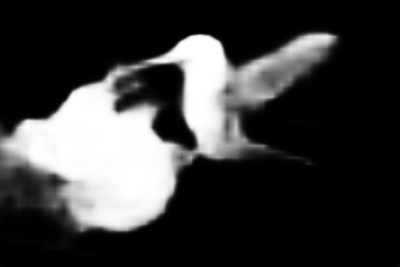}\ &
\includegraphics[width=0.1\linewidth,height=1cm]{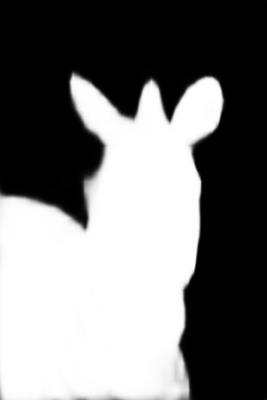}\ &
\includegraphics[width=0.1\linewidth,height=1cm]{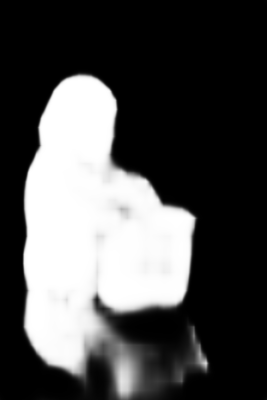}\
\\
\vspace{-1mm}
(u) &
\includegraphics[width=0.1\linewidth,height=1cm]{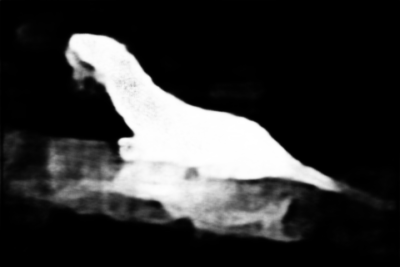}\ &
\includegraphics[width=0.1\linewidth,height=1cm]{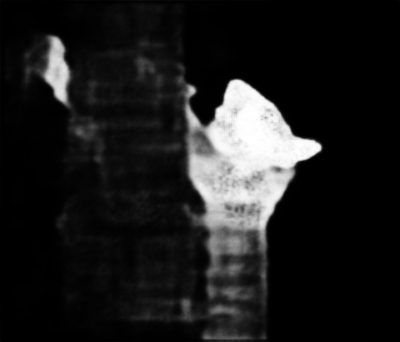}\ &
\includegraphics[width=0.1\linewidth,height=1cm]{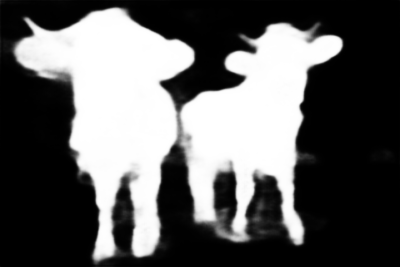}\ &
\includegraphics[width=0.1\linewidth,height=1cm]{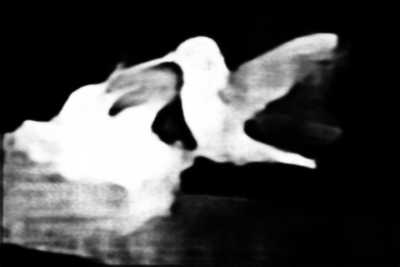}\ &
\includegraphics[width=0.1\linewidth,height=1cm]{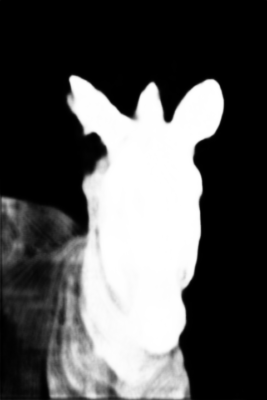}\ &
\includegraphics[width=0.1\linewidth,height=1cm]{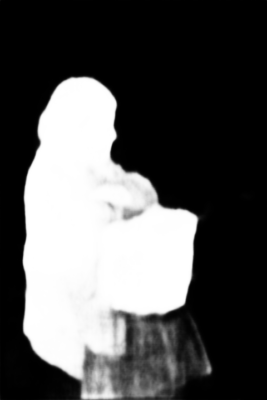}\
\\
\end{tabular}
}
\caption{Comparison of typical saliency maps. From top to bottom: (a) Input images; (b) Ground truth; (c) \textbf{Ours}; (d) \textbf{AMU}~\cite{zhang2017amulet}; (e) \textbf{BDMP}~\cite{zhang2018bi}; (f) \textbf{DCL}~\cite{li2016deep}; (g) \textbf{DGRL}~\cite{wang2018detect}; (h) \textbf{DHS}~\cite{liu2016dhsnet}; (i) \textbf{DS}~\cite{li2016deepsaliency}; (j) \textbf{DSS}~\cite{hou2017deeply}; (k) \textbf{ELD}~\cite{lee2016deep}; (l) \textbf{LEGS}~\cite{wang2015deep}; (m) \textbf{MCDL}~\cite{zhao2015saliency}; (n) \textbf{MDF}~\cite{li2015visual}; (o) \textbf{NLDF}~\cite{luo2017non}; (p) \textbf{PAGRN}~\cite{zhang2018progressive}; (q) \textbf{PICA}~\cite{liu2018picanet};(r) \textbf{RFCN}~\cite{wang2016saliency}; (s) \textbf{RST}~\cite{zhu2017saliency}; (t) \textbf{SRM}~\cite{wang2017stagewise}; (u) \textbf{UCF}~\cite{zhang2017learning}. Due to the limitation of space, we don't show conventional algorithms.
}
\label{fig:map_comparison}
\end{figure*}
\begin{table*}
\begin{center}
\caption{Run time analysis of the compared methods. It is conducted with an i5-6600 CPU and an NVIDIA GeForce GTX 1070 GPU.}
\label{table:times}
\resizebox{1\textwidth}{!}
{
\begin{tabular}{|c|c|c|c|c|c|c|c|c|c|c|c|c|c|c|c|c|c|c|c|c|c|c|c|}
\hline
Models
&Ours
&AMU
&BDMP
&DCL
&DGRL
&DHS
&DS
&DSS
&ELD
&LEGS
&MCDL
&MDF
&NLDF
&PAGRN
&PICA
&RFCN
&RST
&SRM
&UCF
&BL
&BSCA
&DRFI
&DSR
\\
\hline
Times (s)
&0.08
&0.05
&0.06
&0.62
&0.56
&0.05
&0.23
&5.21
&0.93
&2.31
&3.31
&26.45
&2.33
&0.07
&0.236
&5.24
&7.12
&0.16
&0.15
&3.54
&1.42
&47.36
&5.11
\\
\hline
\end{tabular}
}
\end{center}
\vspace{-6mm}
\end{table*}
\vspace{-2mm}
{\flushleft\textbf{Quantitative Evaluation.}}
As illustrated in Tab.~\ref{table:fauc1}, Tab.~\ref{table:fauc2} and Fig.~\ref{fig:PR-curve}, our method outperforms other competing ones across all datasets in terms of near all evaluation metrics.
From these results, we have other notable observations: (1) deep learning based methods consistently outperform traditional methods with a large margin, which further proves the superiority of deep features for SOD.
(2) our method achieves higher S-measure than other methods, especially on complex structure datasets, \emph{e.g.}, the HKU-IS-TE, PASCAL-S, SED and SOD datasets.
We attribute this impressive result to our structural loss.
(3) without segmentation-based pre-training, our method only fine-tuned from the image classification model still achieves better results than the DCL, DS, DSS, and RFCN.
(4) compared to the very recent BDMP, DGRL, DSS and PAGRN, our method is inferior on the DUT-OMRON and DUTS-TE datasets.
However, our method ranks in the second place and is still very comparable.
We note that BDMP, DGRL and PAGRN are trained with the DUTS training dataset.
Thus, it is no wonder they can achieve better performance than other methods.
(5) in contrast to our previous work~\cite{zhang2018salient}, our enhanced method achieves much better performance.

\begin{figure*}
\begin{center}
\resizebox{1\textwidth}{!}
{
\begin{tabular}{@{}c@{}c@{}c@{}c@{}c@{}c@{}c@{}c@{}c@{}c@{}c@{}c@{}c@{}c@{}c@{}c@{}c}
\vspace{-1mm}
\includegraphics[width=0.25\linewidth,height=2.4cm]{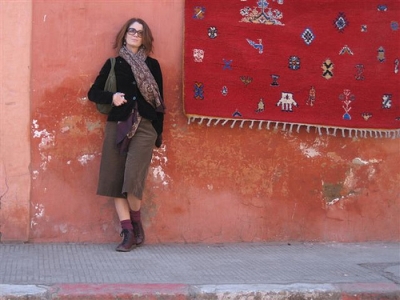} \ &
\includegraphics[width=0.25\linewidth,height=2.4cm]{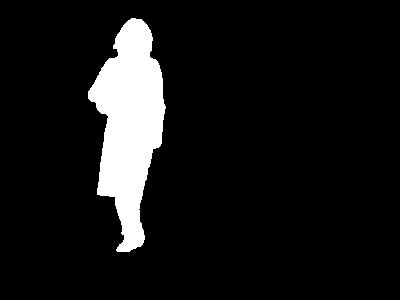} \ &
\includegraphics[width=0.25\linewidth,height=2.4cm]{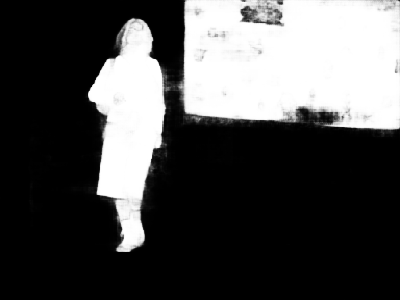} \ &
\includegraphics[width=0.25\linewidth,height=2.4cm]{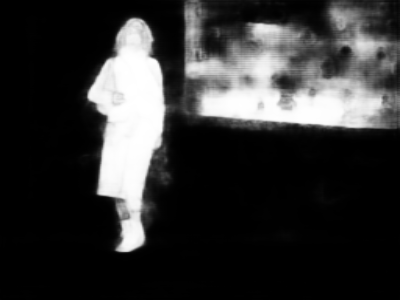} \ &
\includegraphics[width=0.25\linewidth,height=2.4cm]{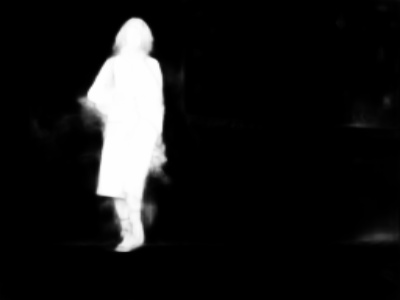} \ &
\includegraphics[width=0.25\linewidth,height=2.4cm]{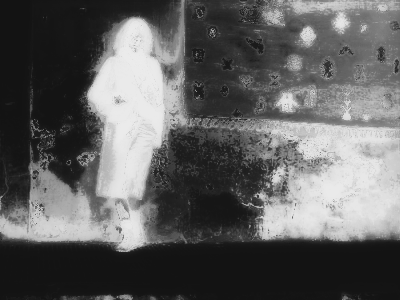} \ &
\includegraphics[width=0.25\linewidth,height=2.4cm]{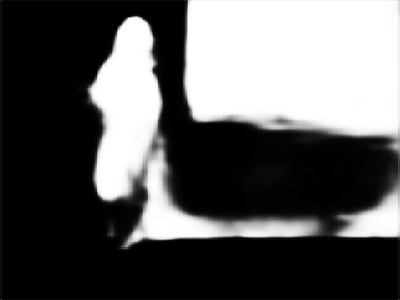} \ &
\includegraphics[width=0.25\linewidth,height=2.4cm]{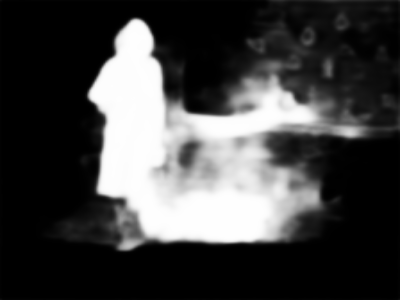} \ &
\includegraphics[width=0.25\linewidth,height=2.4cm]{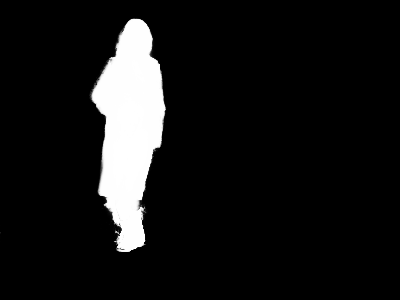} \ &
\includegraphics[width=0.25\linewidth,height=2.4cm]{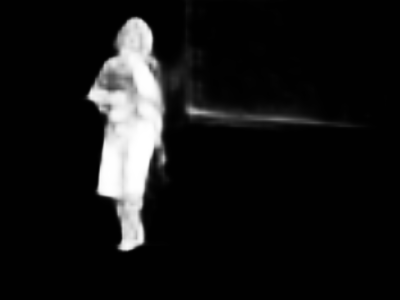} \ &
\includegraphics[width=0.25\linewidth,height=2.4cm]{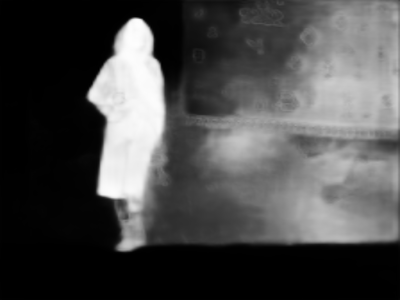} \ &
\includegraphics[width=0.25\linewidth,height=2.4cm]{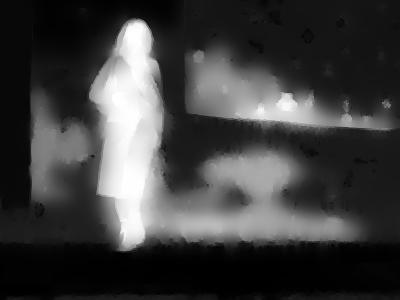} \ &
\includegraphics[width=0.25\linewidth,height=2.4cm]{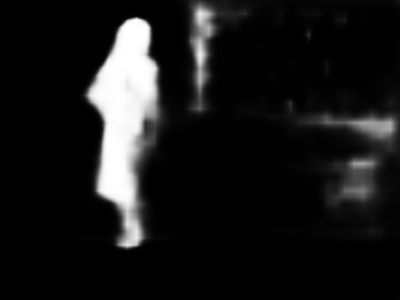} \ &
\includegraphics[width=0.25\linewidth,height=2.4cm]{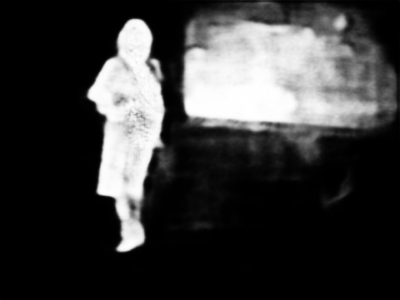} \ \\
\vspace{-1mm}
\includegraphics[width=0.25\linewidth,height=2.4cm]{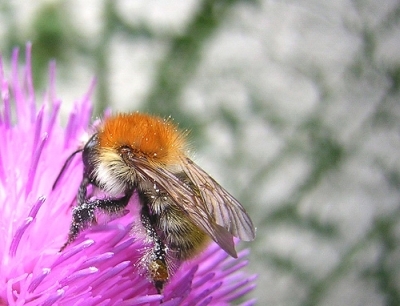} \ &
\includegraphics[width=0.25\linewidth,height=2.4cm]{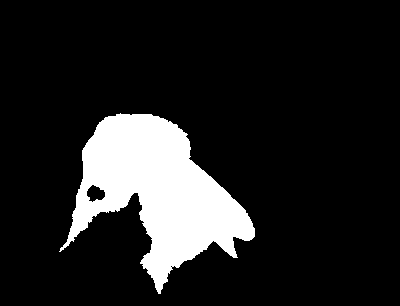} \ &
\includegraphics[width=0.25\linewidth,height=2.4cm]{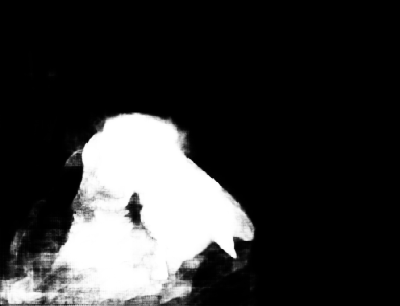} \ &
\includegraphics[width=0.25\linewidth,height=2.4cm]{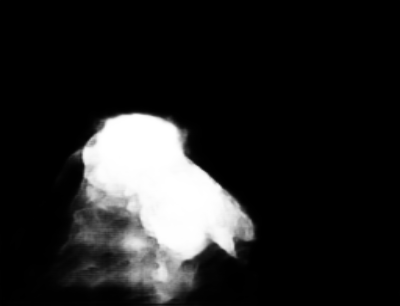} \ &
\includegraphics[width=0.25\linewidth,height=2.4cm]{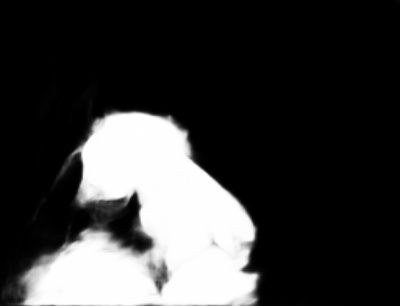} \ &
\includegraphics[width=0.25\linewidth,height=2.4cm]{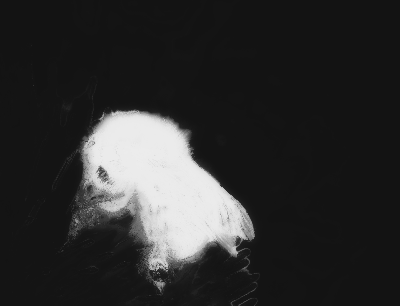} \ &
\includegraphics[width=0.25\linewidth,height=2.4cm]{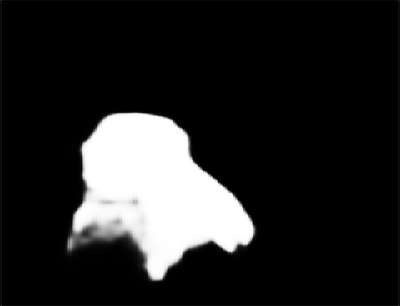} \ &
\includegraphics[width=0.25\linewidth,height=2.4cm]{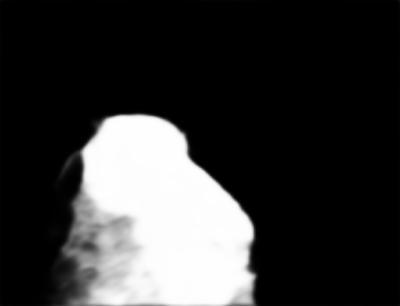} \ &
\includegraphics[width=0.25\linewidth,height=2.4cm]{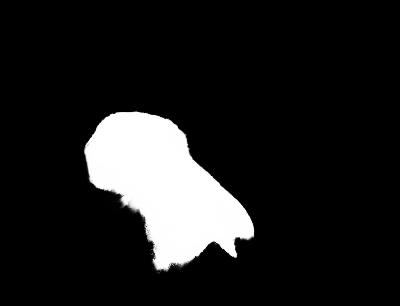} \ &
\includegraphics[width=0.25\linewidth,height=2.4cm]{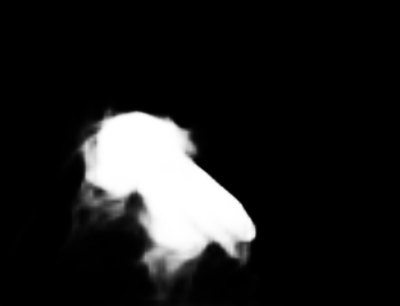} \ &
\includegraphics[width=0.25\linewidth,height=2.4cm]{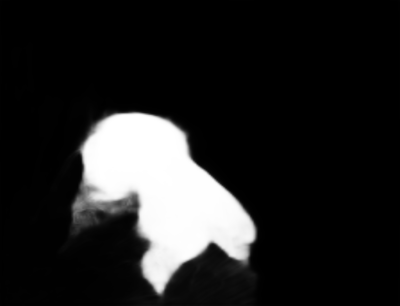} \ &
\includegraphics[width=0.25\linewidth,height=2.4cm]{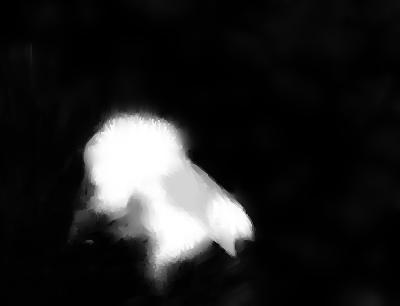} \ &
\includegraphics[width=0.25\linewidth,height=2.4cm]{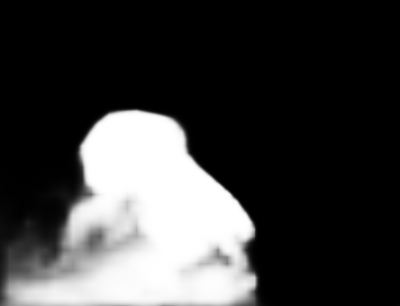} \ &
\includegraphics[width=0.25\linewidth,height=2.4cm]{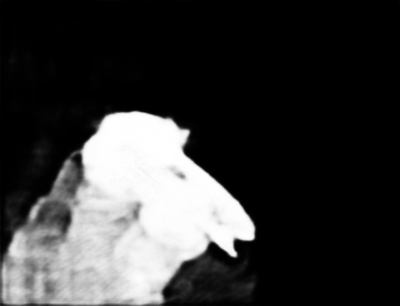} \ \\
\vspace{-1mm}
\includegraphics[width=0.25\linewidth,height=2.4cm]{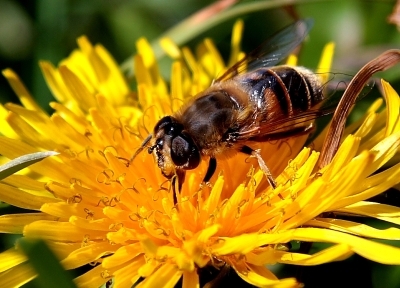} \ &
\includegraphics[width=0.25\linewidth,height=2.4cm]{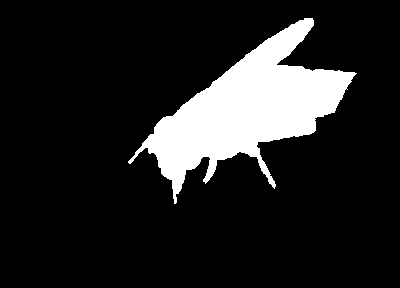} \ &
\includegraphics[width=0.25\linewidth,height=2.4cm]{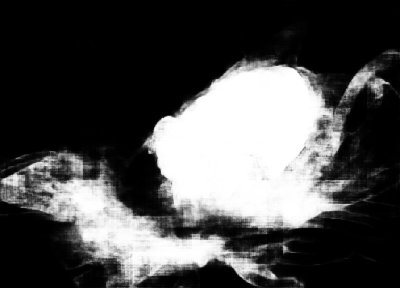} \ &
\includegraphics[width=0.25\linewidth,height=2.4cm]{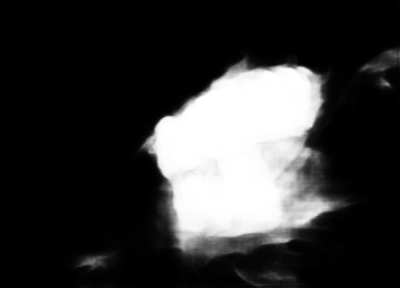} \ &
\includegraphics[width=0.25\linewidth,height=2.4cm]{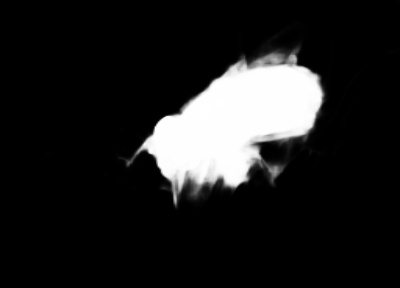} \ &
\includegraphics[width=0.25\linewidth,height=2.4cm]{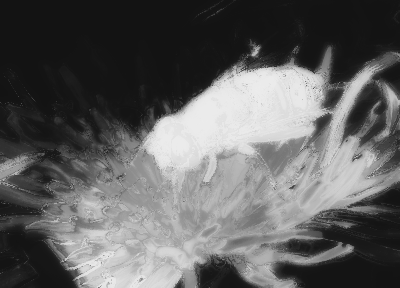} \ &
\includegraphics[width=0.25\linewidth,height=2.4cm]{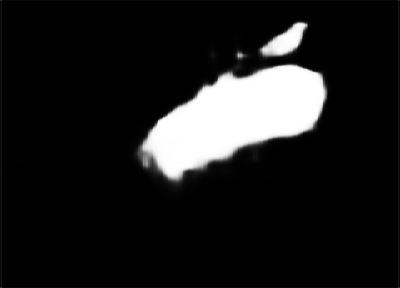} \ &
\includegraphics[width=0.25\linewidth,height=2.4cm]{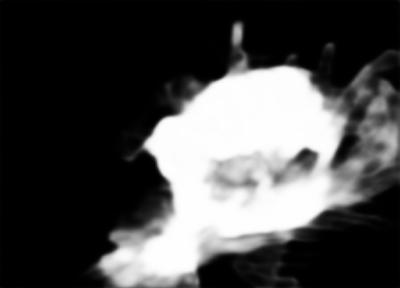} \ &
\includegraphics[width=0.25\linewidth,height=2.4cm]{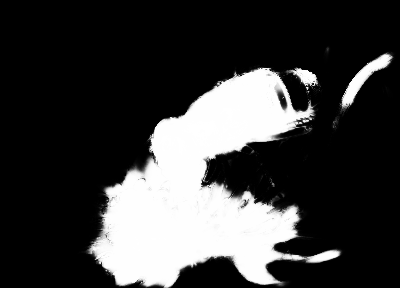} \ &
\includegraphics[width=0.25\linewidth,height=2.4cm]{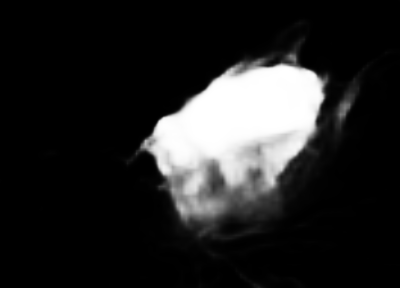} \ &
\includegraphics[width=0.25\linewidth,height=2.4cm]{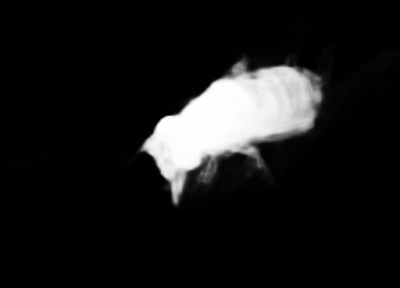} \ &
\includegraphics[width=0.25\linewidth,height=2.4cm]{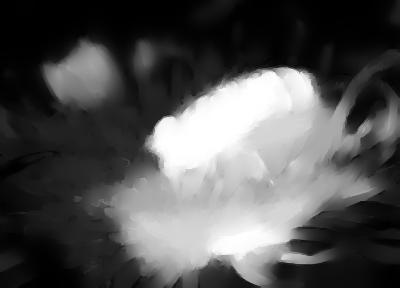} \ &
\includegraphics[width=0.25\linewidth,height=2.4cm]{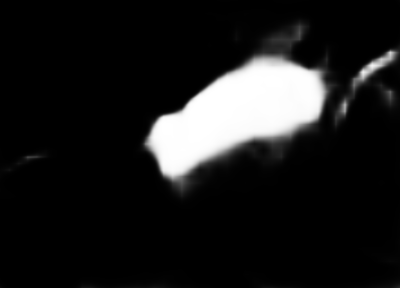} \ &
\includegraphics[width=0.25\linewidth,height=2.4cm]{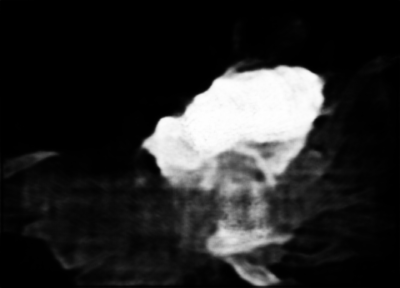} \ \\
\vspace{-1mm}
\includegraphics[width=0.25\linewidth,height=2.4cm]{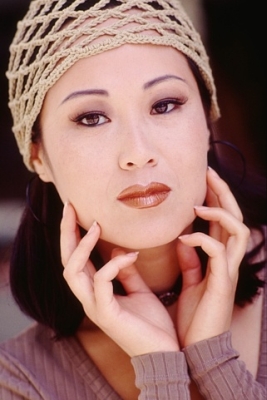} \ &
\includegraphics[width=0.25\linewidth,height=2.4cm]{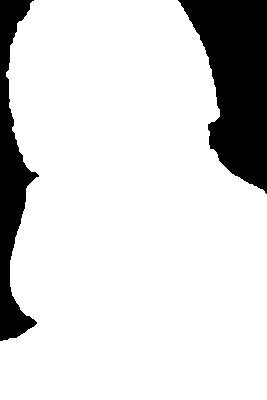} \ &
\includegraphics[width=0.25\linewidth,height=2.4cm]{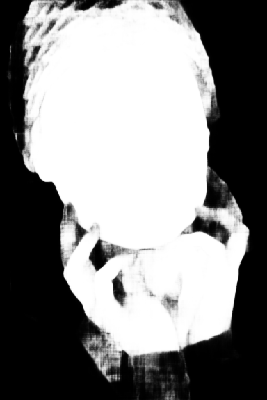} \ &
\includegraphics[width=0.25\linewidth,height=2.4cm]{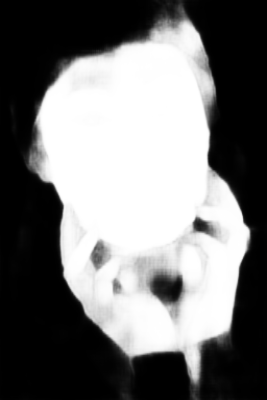} \ &
\includegraphics[width=0.25\linewidth,height=2.4cm]{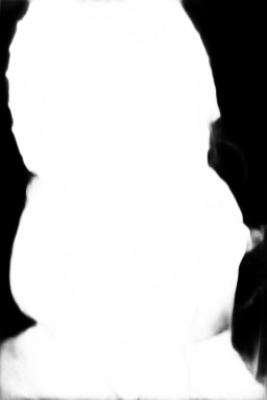} \ &
\includegraphics[width=0.25\linewidth,height=2.4cm]{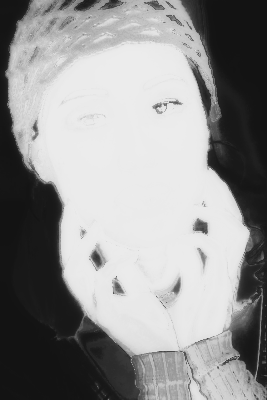} \ &
\includegraphics[width=0.25\linewidth,height=2.4cm]{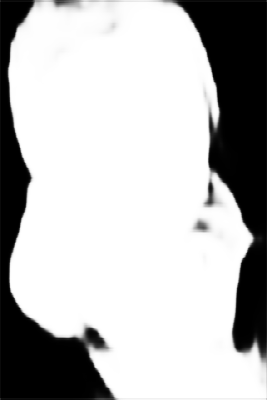} \ &
\includegraphics[width=0.25\linewidth,height=2.4cm]{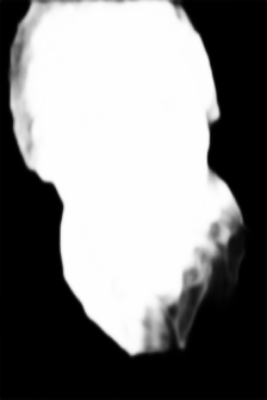} \ &
\includegraphics[width=0.25\linewidth,height=2.4cm]{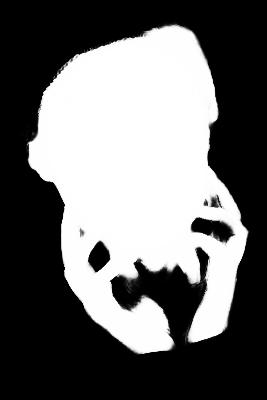} \ &
\includegraphics[width=0.25\linewidth,height=2.4cm]{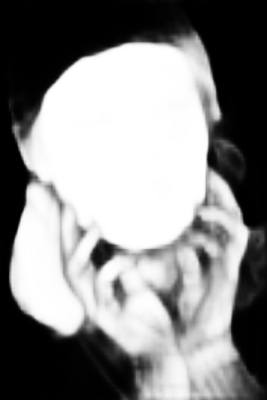} \ &
\includegraphics[width=0.25\linewidth,height=2.4cm]{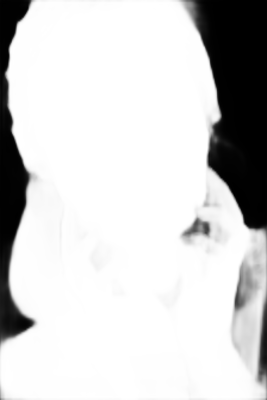} \ &
\includegraphics[width=0.25\linewidth,height=2.4cm]{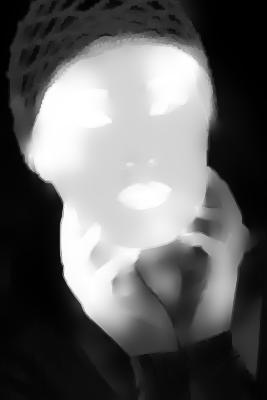} \ &
\includegraphics[width=0.25\linewidth,height=2.4cm]{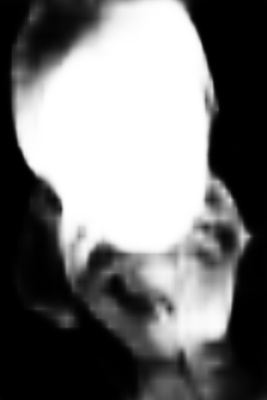} \ &
\includegraphics[width=0.25\linewidth,height=2.4cm]{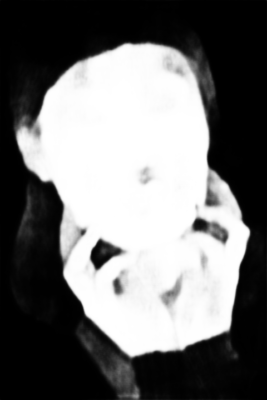} \ \\
\vspace{-1mm}
\includegraphics[width=0.25\linewidth,height=2.4cm]{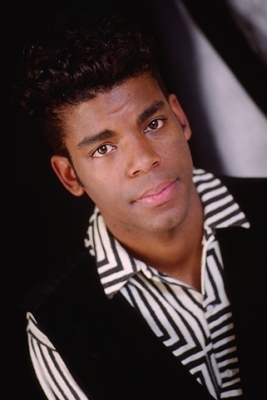} \ &
\includegraphics[width=0.25\linewidth,height=2.4cm]{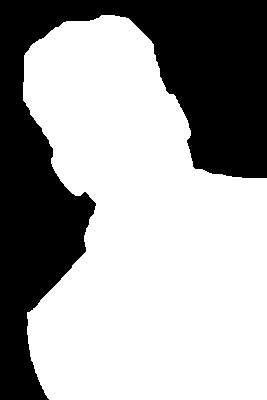} \ &
\includegraphics[width=0.25\linewidth,height=2.4cm]{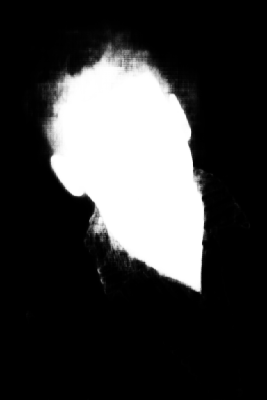} \ &
\includegraphics[width=0.25\linewidth,height=2.4cm]{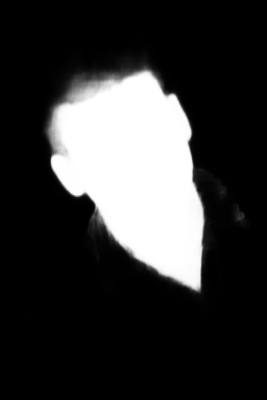} \ &
\includegraphics[width=0.25\linewidth,height=2.4cm]{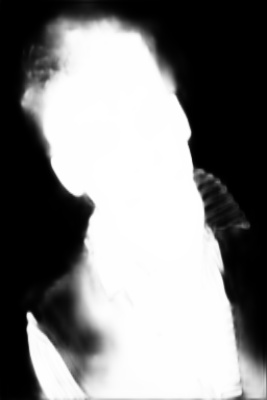} \ &
\includegraphics[width=0.25\linewidth,height=2.4cm]{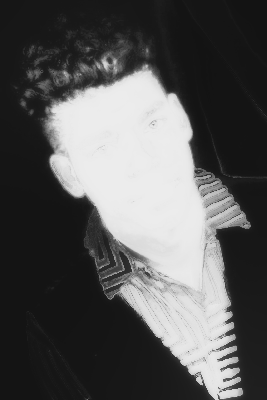} \ &
\includegraphics[width=0.25\linewidth,height=2.4cm]{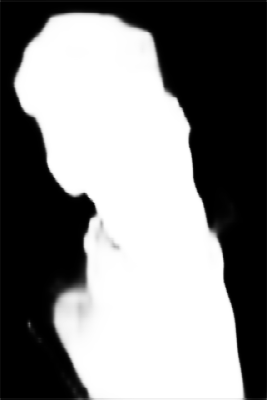} \ &
\includegraphics[width=0.25\linewidth,height=2.4cm]{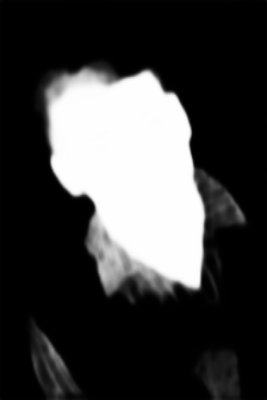} \ &
\includegraphics[width=0.25\linewidth,height=2.4cm]{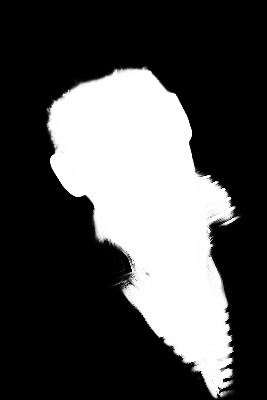} \ &
\includegraphics[width=0.25\linewidth,height=2.4cm]{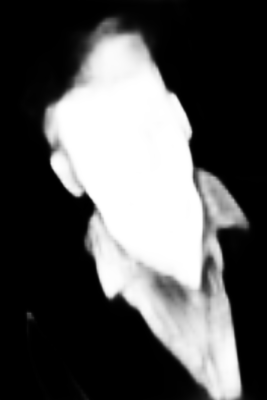} \ &
\includegraphics[width=0.25\linewidth,height=2.4cm]{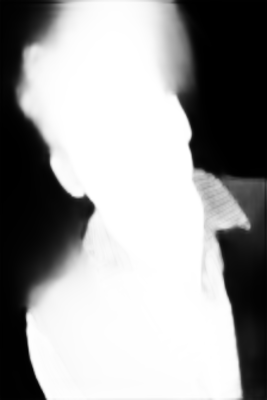} \ &
\includegraphics[width=0.25\linewidth,height=2.4cm]{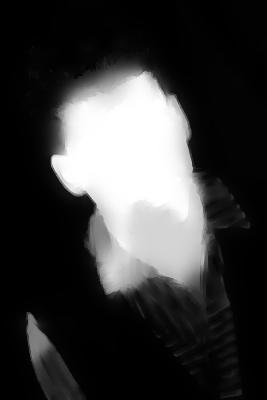} \ &
\includegraphics[width=0.25\linewidth,height=2.4cm]{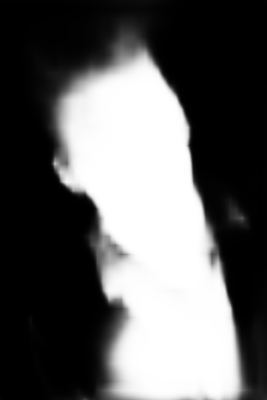} \ &
\includegraphics[width=0.25\linewidth,height=2.4cm]{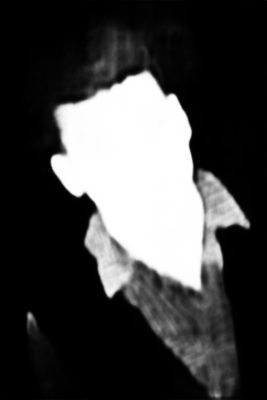} \ \\
\vspace{-1mm}
\includegraphics[width=0.25\linewidth,height=2.4cm]{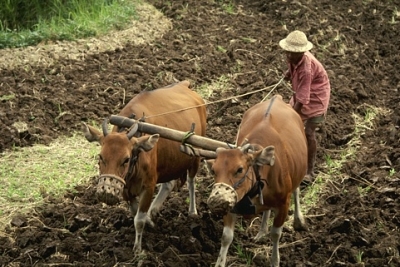} \ &
\includegraphics[width=0.25\linewidth,height=2.4cm]{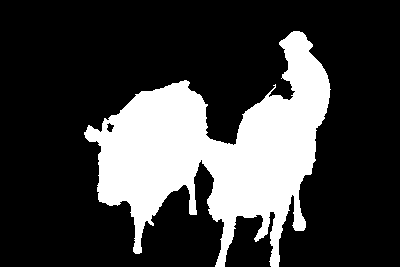} \ &
\includegraphics[width=0.25\linewidth,height=2.4cm]{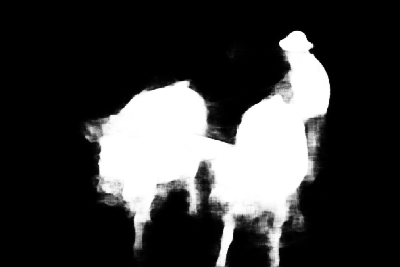} \ &
\includegraphics[width=0.25\linewidth,height=2.4cm]{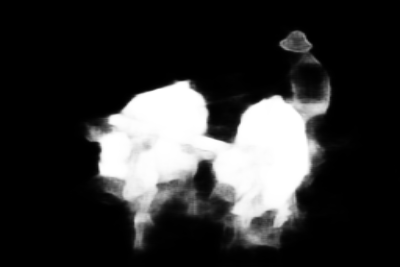} \ &
\includegraphics[width=0.25\linewidth,height=2.4cm]{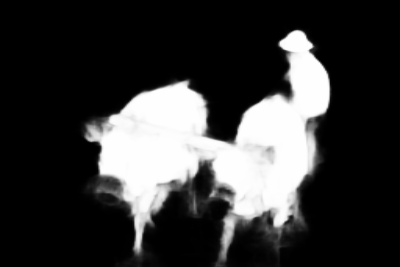} \ &
\includegraphics[width=0.25\linewidth,height=2.4cm]{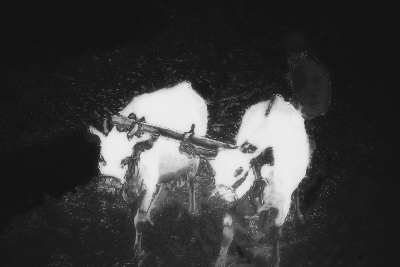} \ &
\includegraphics[width=0.25\linewidth,height=2.4cm]{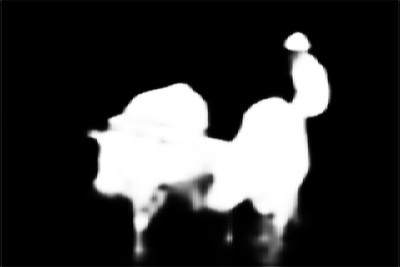} \ &
\includegraphics[width=0.25\linewidth,height=2.4cm]{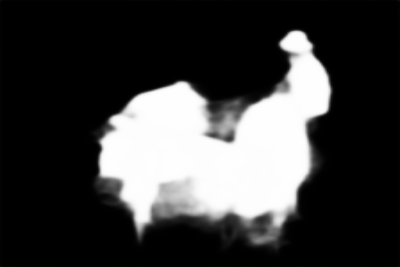} \ &
\includegraphics[width=0.25\linewidth,height=2.4cm]{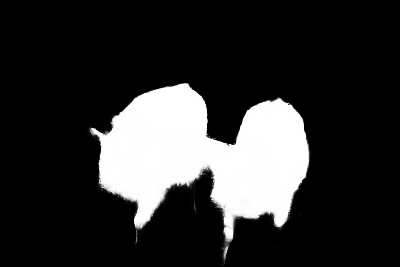} \ &
\includegraphics[width=0.25\linewidth,height=2.4cm]{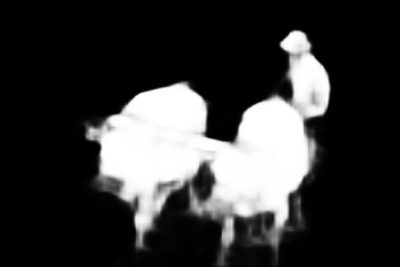} \ &
\includegraphics[width=0.25\linewidth,height=2.4cm]{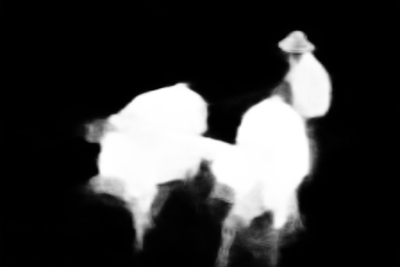} \ &
\includegraphics[width=0.25\linewidth,height=2.4cm]{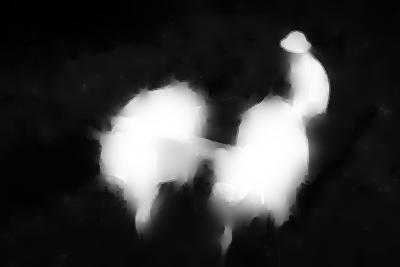} \ &
\includegraphics[width=0.25\linewidth,height=2.4cm]{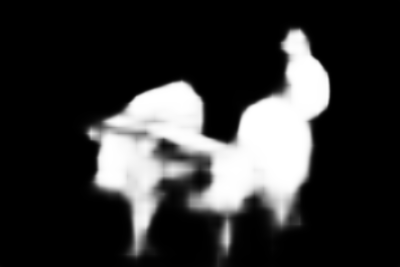} \ &
\includegraphics[width=0.25\linewidth,height=2.4cm]{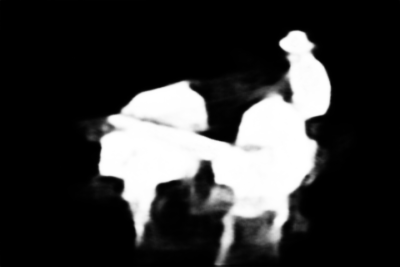} \ \\
{\Huge (a) } & {\Huge (b)} & {\Huge (c)} & {\Huge (d)} & {\Huge (e)} & {\Huge (f)} & {\Huge (g)} & {\Huge (h)}& {\Huge (i)} & {\Huge (j)} & {\Huge (k)}& {\Huge (l)} & {\Huge (m)} & {\Huge (n)} \\
\end{tabular}
}
\caption{Failure examples with different deep learning based approaches. From top to bottom: (a) Input images; (b) Ground truth; (c) \textbf{Ours}; (d) \textbf{AMU}~\cite{zhang2017amulet}; (e) \textbf{BDMP}~\cite{zhang2018bi}; (f) \textbf{DCL}~\cite{li2016deep}; (g) \textbf{DGRL}~\cite{wang2018detect}; (h) \textbf{DHS}~\cite{liu2016dhsnet};(i) \textbf{DSS}~\cite{hou2017deeply}; (j) \textbf{PAGRN}~\cite{zhang2018progressive}; (k) \textbf{PICA}~\cite{liu2018picanet}; (l) \textbf{RFCN}~\cite{wang2016saliency}; (m) \textbf{SRM}~\cite{wang2017stagewise}; (n) \textbf{UCF}~\cite{zhang2017learning}.
}
\label{fig:failure}
\end{center}
\vspace{-4mm}
\end{figure*}

Tab.~\ref{table:times} shows a comparison of running times.
As it can be seen, our method is much faster than most of compared methods.
The testing process only costs 0.08s for each image.
\vspace{-6mm}
{\flushleft\textbf{Qualitative Evaluation.}}
Fig.~\ref{fig:map_comparison} provides several examples in various challenging cases, where our method outperforms other compared methods.
For example, the images in the first two columns are of very low contrast between the objects and backgrounds.
Most of the compared methods fail to capture the whole salient objects, while our method successfully highlights them with sharper edges preserved.
For multiple salient objects (the 3-4 columns), our method still generates more accurate saliency maps.
The images in the 5-6 columns are challenging with complex structures or salient objects near the image boundary.
Most of the compared methods can not predict the whole objects, while our method captures the whole salient regions with preserved structures.
Fig.~\ref{fig:failure} shows some failure examples of the proposed method.
For example, when the salient objects are hard to define (the first three rows), the proposed method detect the backgrounds as salient objects.
While in this case the human visual system can also easily focus on the ambiguous objects.
When the salient objects have very large disconnected regions (the 4-5 rows), the proposed method cannot assign different values to them and distinguish them.
When the salient objects are very small and have similar colors with backgrounds (the last row), it is difficult to correctly detect them with our method.
The other deep learning based approaches almost also fail to detect the salient objects.
\subsection{Ablation Analysis}
We also conduct experiments to evaluate the main components of our model.
All models in this subsection are trained on the augmented MSRA10K dataset and share the same hyper-parameters described in Section IV. B.
Due to the limitation of space, we only show the results on the ECSSD dataset.
Other datasets have the similar performance trend.
For these experiments, if not specified, we use AdaBN~\cite{zhang2018salient} to erase the effect of layer-wise AdaBN.
\begin{table}
\begin{center}
\caption{Results with different input settings on the ECSSD dataset.}
\label{table:input}
\resizebox{0.5\textwidth}{!}
{
\begin{tabular}{|c|c|c|c|c|c|c|}
\hline
Inputs &RGB&O-Input&R-Input&RGB+YCbCr&O-Input+R-Input\\
\hline
$F_\eta\uparrow$         &0.802&0.844& 0.845&0.852& 0.880      \\
\hline
$MAE\downarrow$          &0.1415&0.1023& 0.1002&0.0931& 0.0523     \\
\hline
$S_\lambda\uparrow$      &0.853&0.879& 0.880&0.882& 0.897      \\
\hline
\end{tabular}
}
\end{center}
\vspace{-4mm}
\end{table}
\begin{table}
\begin{center}
\caption{Results with different mean settings on the ECSSD dataset. The best result are shown in bold.}
\label{table:mean}
\resizebox{0.5\textwidth}{!}
{
\begin{tabular}{|c|c|c|c|c|c|c|}
\hline
Means         &ImageNet&MSRA10k&Each Image&Middle Mean&Zero Mean \\
\hline
$F_\eta\uparrow$       &0.880&0.879& 0.875&0.873& 0.871      \\
\hline
$MAE\downarrow$          &0.0523&0.0535& 0.0544&0.0602& 0.0605      \\
\hline
$S_\lambda\uparrow$    &0.897&0.895& 0.886&0.887& 0.889      \\
\hline
$Time (h)\downarrow$     &38.5&38.5& 50.2&38.5&33.7      \\
\hline
\end{tabular}
}
\end{center}
\vspace{-6mm}
\end{table}
\vspace{-2mm}
{\flushleft\textbf{Effect of Complementary Inputs.}}
Our proposed model takes paired images as inputs for complementary information.
One may be curious about the performance only with one input, for example, O-Input or R-Input (no paired input given).
With only one input, our model reduces to a plain multi-level feature fusion model.
Tab.~\ref{table:input} shows the results with different input settings.
From the experimental results, we observe that 1) only with the O-Input (R-Input), our model achieves 0.844 (0.845), 0.102 (0.100) and 0.879 (0.880) in terms of the F-measure, MAE and S-measure metrics on the ECSSD dataset.
These results are very comparable with other state-of-the-art methods and much better than the raw RGB image input.
This indicates that our feature fusion model is very effective.
2) with paired reciprocal inputs, our model achieves about 4\% performance leap of F-measure, around 2.5\% improvement of S-measure, as well as around 50\% decrease in MAE compared with only one input.
To show that the extra model complexity is not the reason of improvements, we also perform an experiment in which RGB and YCbCr images are input into our framework (everything else remains the same).
The paired RGB+YCbCr input can achieve much better performance than raw RGB, and a little better performance than the O-Input or R-Input. The main reason may be the introduction of additional YCbCr information.
Obviously, our reflection features achieve better performance than the one of RGB+YCbCr in all metrics.
From these results, we can see that even the O-Input and R-Input are simple reciprocal in planar reflection, the complementary effect is very expressive.
We attribute this improvement to the lossless feature learning.
\vspace{-2mm}
{\flushleft\textbf{Effect of Adopted Means.}}
As described in the ``Reciprocal Image Input'' part, the mean is not unique and the main reason we chose the mean of ImageNet is reducing the computation burden.
To verify the effect of different means, we also perform experiments with typical means in dataset-level (trained MSRA10K) or image-level (Each Image Mean, Middle Mean and Zero Mean).
The middle mean is set to (128, 128, 128).
Experimental results are shown in Tab.~\ref{table:mean}.
With the pre-computed mean of the MSRA10K dataset, our model shows a little performance decrease compared with using the ImageNet mean.
The main reason may be that the model is initialized from the pre-trained model on the ImageNet classification task.
With the same mean subtraction, the model can minimize the transfer learning gap between the image classification and saliency detection tasks.
With the mean of each input image, we find that it is still feasible to train our model.
The results are still comparable. For example, we got 0.875 (F-measure), 0.054 (MAE) and 0.886 (S-measure) on the ECSSD dataset.
However, because we need to compute the mean of each input image online, the training is relatively slow and slightly unstable.
With the middle mean, the model achieves worse results.
For the zero mean, the model achieves similar results with the middle mean.
However, the training time is reduced because there is no need to compute and subtract the mean online.
Besides, compared to the results of O-Input/R-Input in Tab.~\ref{table:input} (using the ImageNet mean), the O-Input+R-Input can bring about 0.035 improvement of F-measure.
Zero mean brings about 0.070 improvement to raw RGB image input in Tab.~\ref{table:input}.
Meanwhile, using O-Input+R-Input and ImageNet mean can bring about 0.078 improvement to raw RGB image input.
For the performance improvements, the main reason is that the paired inputs can capture more information than single input.
\vspace{-2mm}
{\flushleft\textbf{Effect of Changing the Reflection Parameter.}}
In the proposed reflection, we adopt the multiplicative operator $k$ to measure the reflection scale.
Changing the reflection parameter $k$ will result into different reflection planes.
Tab.~\ref{table:changek} shows experimental results with different values of $k$ on the ECSSD dataset.
When enlarging $k>1$, our model makes improvement on performance.
However, the improvement is rather small and is reasonable. In fact, enlarging $k$ only changes the range of reflection images.
After the AdaBN, the impact of value range of reflection images decreased significantly, resulting to the similar performance.
Besides, when adopting negative values of $k$, the performance has almost no improvement.
The reason is obvious. When using negative values of $k$, the two sibling branches essentially take the same input $X-M$ (even scaled), as shown in Equ.~\ref{equ:equ1}. The complementary of the paired input is very limited.
These results further prove that the feature reflection is the core of performance improvements.
\begin{table}
\begin{center}
\caption{Results with different values of $k$ on the ECSSD dataset.}
\label{table:changek}
\resizebox{0.5\textwidth}{!}
{
\begin{tabular}{|c|c|c|c|c|c|c|c|}
\hline
$k$ &-4&-2&-1&0&1&2&4\\
\hline
$F_\eta\uparrow$         &0.842&0.844& 0.843&0.844&0.880 &0.880&0.881      \\
\hline
$MAE\downarrow$          &0.1025&0.1023& 0.1022&0.1023&0.0523&0.0520& 0.0521     \\
\hline
$S_\lambda\uparrow$      &0.880&0.879& 0.880&0.879&0.897&0.898& 0.898      \\
\hline
\end{tabular}
}
\end{center}
\vspace{-6mm}
\end{table}
\begin{table*}
\begin{center}
\caption{Results with different model settings on ECSSD. The best three results are shown in \textcolor[rgb]{1,0,0}{red},~\textcolor[rgb]{0,1,0}{green} and \textcolor[rgb]{0,0,1}{blue}, respectively.}
\label{table:aggregation}
\resizebox{1\textwidth}{!}
{
\begin{tabular}{|c|c|c|c|c|c|c|c|c|c|c|c|}
\hline
Models
&(a) \tiny{SFCN}-hf+$\mathcal{L}_{bce}$
&(b) \tiny{SFCN}-hf+$\mathcal{L}_{wbce}$
&(c) \tiny{SFCN}-hf+$\mathcal{L}_{wbce}$+$\mathcal{L}_{sc}$
&(d) \tiny{SFCN}-hf+$\mathcal{L}_{wbce}$+$\mathcal{L}_{sc}$+$\mathcal{L}_{s1}$
&(e) \tiny{SFCN}+$\mathcal{L}_{bce}$
&(f) \tiny{SFCN}+$\mathcal{L}_{wbce}$
&(g) \tiny{SFCN}+$\mathcal{L}_{wbce}$+$\mathcal{L}_{sc}$
&(h) \tiny{SFCN}+$\mathcal{L}_{wbce}$+$\mathcal{L}_{s1}$
&Overall \\
\hline
$F_\eta\uparrow$       &0.824&0.841&0.858&0.863&0.848& 0.865& \textcolor[rgb]{0,1,0}{0.873}& \textcolor[rgb]{0,0,1}{0.867}&\textcolor[rgb]{1,0,0}{0.880}        \\
\hline
$MAE\downarrow$          &0.1024&0.1001&0.0923&0.0706&0.0834& 0.0721& \textcolor[rgb]{0,0,1}{0.0613}& \textcolor[rgb]{1,0,0}{0.0491}&\textcolor[rgb]{0,1,0}{0.0523}       \\
\hline
$S_\lambda\uparrow$    &0.833&0.852&0.860&0.869&0.872& 0.864& \textcolor[rgb]{0,0,1}{0.880}& \textcolor[rgb]{0,1,0}{0.882}&\textcolor[rgb]{1,0,0}{0.897}       \\
\hline
\end{tabular}
}
\end{center}
\vspace{-2mm}
\end{table*}
\begin{figure*}
\centering
\resizebox{0.8\textwidth}{!}
{
\begin{tabular}{@{}c@{}c@{}c@{}c@{}c@{}c@{}c@{}c@{}c@{}c@{}c@{}c}
\vspace{-0.5mm}
\hspace{-2mm}
\includegraphics[width=0.2\linewidth,height=2.5cm]{0027.jpg}\hspace{0.1mm}\ &
\includegraphics[width=0.2\linewidth,height=2.5cm]{0027_GT.png}\hspace{0.1mm}\ &
\includegraphics[width=0.2\linewidth,height=2.5cm]{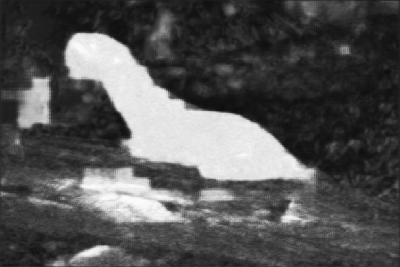}\hspace{0.1mm}\ &
\includegraphics[width=0.2\linewidth,height=2.5cm]{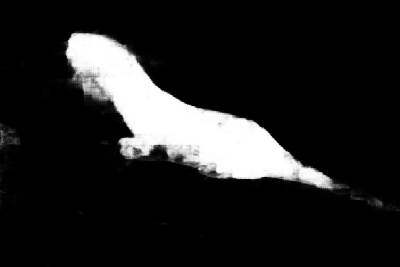}\hspace{0.1mm}\ &
\includegraphics[width=0.2\linewidth,height=2.5cm]{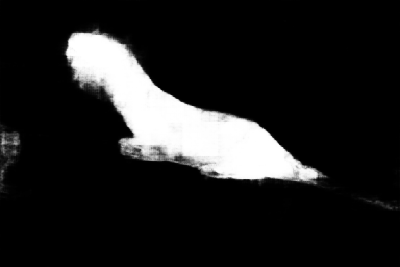}\hspace{0.1mm}\ &
\includegraphics[width=0.2\linewidth,height=2.5cm]{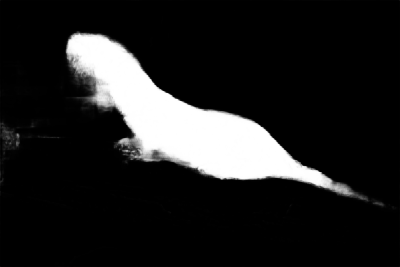}\hspace{0.1mm}\ &
\includegraphics[width=0.2\linewidth,height=2.5cm]{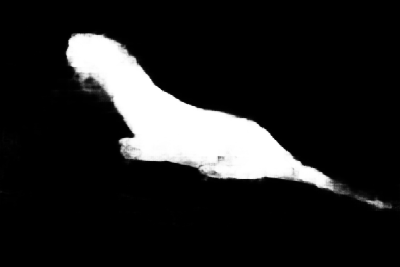}\hspace{0.1mm}\ &
\includegraphics[width=0.2\linewidth,height=2.5cm]{0027_LFR.png}\hspace{0.1mm}\ \\
\vspace{-0.5mm}
\hspace{-2mm}
\includegraphics[width=0.2\linewidth,height=2.5cm]{0023.jpg}\hspace{0.1mm}\ &
\includegraphics[width=0.2\linewidth,height=2.5cm]{0023_GT.png}\hspace{0.1mm}\ &
\includegraphics[width=0.2\linewidth,height=2.5cm]{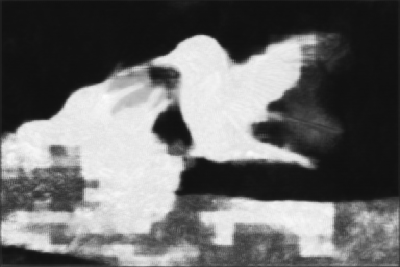}\hspace{0.1mm}\ &
\includegraphics[width=0.2\linewidth,height=2.5cm]{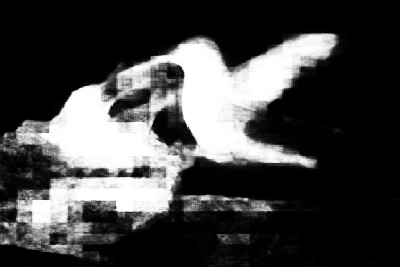}\hspace{0.1mm}\ &
\includegraphics[width=0.2\linewidth,height=2.5cm]{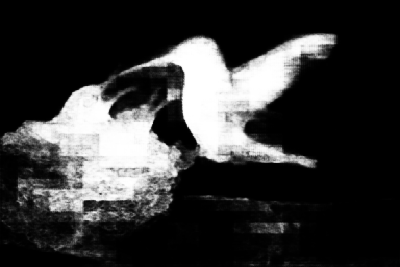}\hspace{0.1mm}\ &
\includegraphics[width=0.2\linewidth,height=2.5cm]{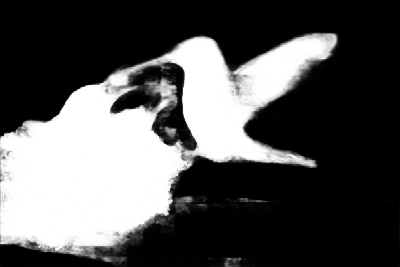}\hspace{0.1mm}\ &
\includegraphics[width=0.2\linewidth,height=2.5cm]{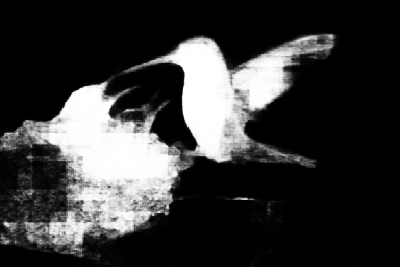}\hspace{0.1mm}\ &
\includegraphics[width=0.2\linewidth,height=2.5cm]{0023_LFR.png}\hspace{0.1mm}\ \\
\vspace{-0.5mm}
\hspace{-2mm}
\includegraphics[width=0.2\linewidth,height=2.5cm]{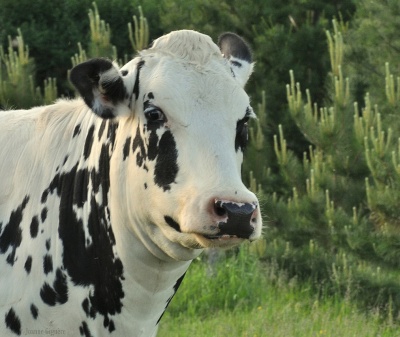}\hspace{0.1mm}\ &
\includegraphics[width=0.2\linewidth,height=2.5cm]{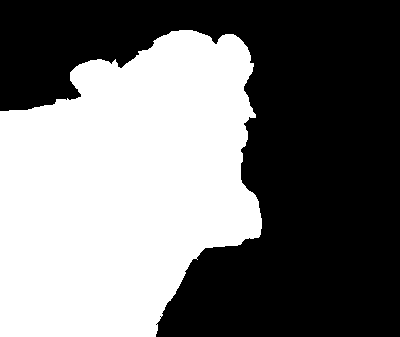}\hspace{0.1mm}\ &
\includegraphics[width=0.2\linewidth,height=2.5cm]{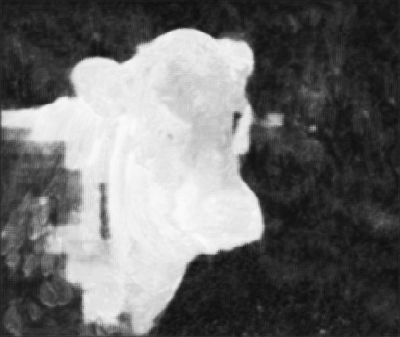}\hspace{0.1mm}\ &
\includegraphics[width=0.2\linewidth,height=2.5cm]{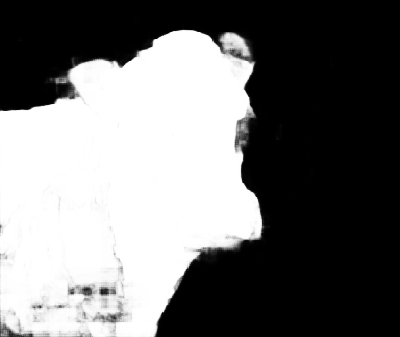}\hspace{0.1mm}\ &
\includegraphics[width=0.2\linewidth,height=2.5cm]{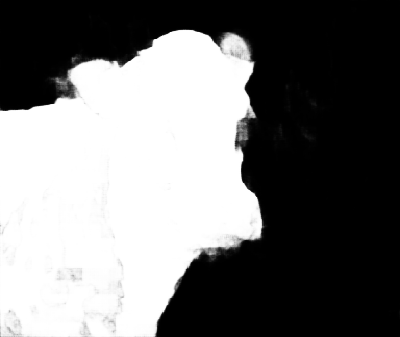}\hspace{0.1mm}\ &
\includegraphics[width=0.2\linewidth,height=2.5cm]{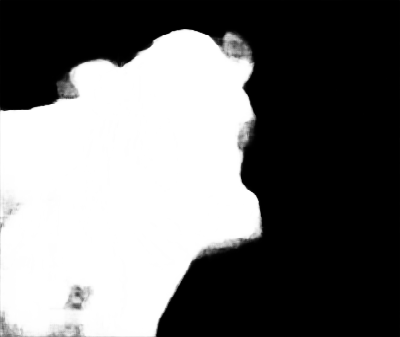}\hspace{0.1mm}\ &
\includegraphics[width=0.2\linewidth,height=2.5cm]{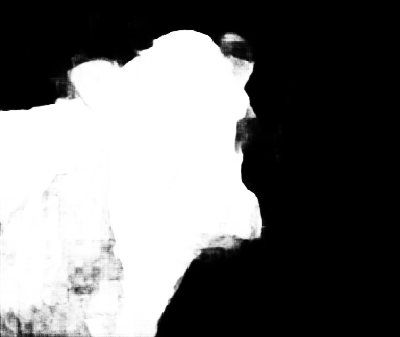}\hspace{0.1mm}\ &
\includegraphics[width=0.2\linewidth,height=2.5cm]{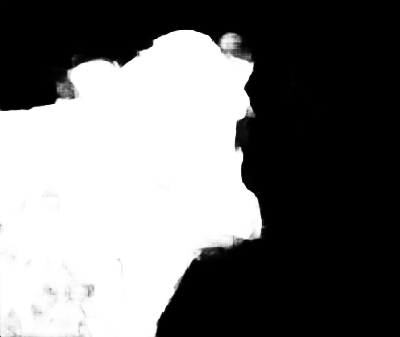}\hspace{0.1mm}\ \\
{\small (a)} & {\small(b)} & {\small(c)} & {\small(d)} & {\small(e)}& {\small(f)}& {\small(g)} & {\small(h)}\ \\
\end{tabular}
}
\vspace{-2mm}
\caption{Visual comparison of typical saliency maps with different model settings. From left to right: (a) Input images; (b) Ground truth; Results of (c) \small{SFCN}-hf+$\mathcal{L}_{bce}$; (d) \small{SFCN}+$\mathcal{L}_{bce}$; (e) \small{SFCN}+$\mathcal{L}_{wbce}$; (f) \small{SFCN}+$\mathcal{L}_{wbce}$+$\mathcal{L}_{sc}$; (g) \small{SFCN}+$\mathcal{L}_{wbce}$+$\mathcal{L}_{s1}$; (h) The overall model \small{SFCN}+$\mathcal{L}_{wbce}$+$\mathcal{L}_{sc}$+$\mathcal{L}_{s1}$.
}
\vspace{-2mm}
\label{fig:comparison}
\end{figure*}
\vspace{-2mm}
{\flushleft\textbf{Effect of Hierarchical Fusion.}}
The results in Tab.~\ref{table:aggregation} show the effect of hierarchical fusion module.
We can see that the SFCN only using the channel concatenation operator without hierarchical fusion (model (a)) has achieved comparable performance to most deep learning based methods.
This further confirms the effectiveness of learned reflection features.
With the hierarchical fusion, the resulting SFCN (model (e)) improves the performance by about 2\% leap.
The main reason is that the fusion method introduces more contextual information from high layers to low layers, which helps to locate the salient objects with large receptive fields.
We also try to use summation for fusing the reflection information.
We found that the resulting model can reduce the feature dimensionality and parameters.
Besides, the model converges faster than our proposed SFCN.
However, the performance decreases too much.
For example, the F-measure of the ECSSD dataset decreases from 0.848 to 0757.
The main reason may be that summing the features weakens interaction or fusion, leading to the degeneration problem.
With concatenation, the features are still kept and can be recaptured for the robust prediction.
\vspace{-2mm}
{\flushleft\textbf{Effect of Diverse Losses.}}
The proposed weighted structural loss also plays a key role in our model.
Tab.~\ref{table:aggregation} shows the effect of different losses.
Because the essential imbalance of salient/no-salient classes, it's no wonder that training with the class-weighted loss $\mathcal{L}_{wbce}$ (model (b) and model (f)) achieves better results than the plain cross-entropy loss $\mathcal{L}_{bce}$ (model (a) and model (e)), about 2\% improvement in three main evaluation metrics.
With other two losses $\mathcal{L}_{sc}$ and $\mathcal{L}_{s1}$, the model achieves better performance in terms of MAE and S-measure.
Specifically, the semantic content loss $\mathcal{L}_{sc}$ improves the S-measure with a large margin.
Thus, our model can capture spatially consistent saliency maps.
The edge-preserving loss $\mathcal{L}_{s1}$ significantly decreases the MAE, providing clear object boundaries.
These results demonstrate that individual loss components complement each other.
When taking them together, the overall model, \emph{i.e.}, SFCN+$\mathcal{L}_{wbce}$+$\mathcal{L}_{sc}$+$\mathcal{L}_{s1}$, achieves best results in all evaluation metrics.
In addition, our model can be trained with these losses in the end-to-end manner.
This advantage makes our saliency inference more efficient, without any bells and whistles, such as result fusing~\cite{wang2015deep,li2016deepsaliency,zhang2017amulet} and condition random field (CRF) post-processing~\cite{li2016deep,wang2016saliency}.
For better understanding, we also provide several visual comparisons with different losses in Fig.~\ref{fig:comparison}.
\begin{table}
\begin{center}
\caption{Results with/without sharing weights on the ECSSD dataset.}
\label{table:sharing weights}
\resizebox{0.35\textwidth}{1cm}
{
\begin{tabular}{|c|c|c|c|c|c|c|}
\hline
Models &w/o &w/ \\
\hline
$F_\eta\uparrow$         &0.862&0.880      \\
\hline
$MAE\downarrow$          &0.0569&0.0523     \\
\hline
$S_\lambda\uparrow$      &0.883&0.897     \\
\hline
Model size (MB)$\downarrow$      &386.1&194.4      \\
\hline
\end{tabular}
}
\end{center}
\vspace{-6mm}
\end{table}
\vspace{-2mm}
{\flushleft\textbf{Effect of Sharing Weights.}}
To achieve the lossless reflection features, the two sibling branches of our SFCN are designed to share weights in each convolutional layer.
In order to verify the effect of sharing weights, we also perform experiments with models of independently learnable weights.
Note that if the SFCN doesn't share weights between two sibling branches, it has about $2\times$ parameters.
To some extent, the model is over-fitting.
We find the performance decreases. The results are shown in Tab.~\ref{table:sharing weights}.
From the results, we can see that with shared weights the model can significantly improve the performance.
Thus, it is beneficial for each branch to share weights and keep its own BN statistics in each layer.
\vspace{-2mm}
{\flushleft\textbf{Effect of Layer-wise AdaBN.}}
In our SFCN, we adopt layer-wise AdaBN operators in each convolutional layer for capturing each domain statistic.
The results in Tab.~\ref{table:fauc1} and Tab.~\ref{table:fauc2} show that our proposed layer-wise AdaBN generally outperforms our previous plain AdaBN~\cite{zhang2018salient}.
To further verify the effectiveness, we replace the proposed layer-wise AdaBN with the regular BN~\cite{ioffe2015batch} in our SFCN model.
Tab.~\ref{table:adabn} shows that when adopting the regular BN, the performance decreases significantly when testing.
As described in Section III. A, the main reason is that there is a large domain shift for the testing in the different domain.
Besides, the size of minibatch also affects our proposed layer-wise AdaBN and regular BN.
For comparison, we have also tried to use different batch size, \emph{e.g.}, 1, 2, 4, 8, 12, 16, 20, 24, etc.
We find that 1) with the small batch size ($<4$), the training is unstable and the performance of our model may decrease; 2) with the relatively large batch size ($>=$16), the influence of batch size is very limited to performance, and larger batch size will cost more computation resources.
Thus, we adopt the batch size of 12 to achieve both effectiveness and efficiency.
\begin{table}
\begin{center}
\doublerulesep=0.1pt
\caption{Results with different BN settings on the ECSSD dataset.}
\label{table:adabn}
\resizebox{0.45\textwidth}{!}
{
\begin{tabular}{|c|c|c|c|c|c|c|c|c|c|c|c|c|c|c|c|c|c|c|c|c|c|c|c|c|||c|c|c|c|c|c|c|c|||}
\hline
\multicolumn{4}{|c|}{}
&\multicolumn{6}{|c|}{BN~\cite{ioffe2015batch}}
&\multicolumn{6}{|c|}{Layer-wise AdaBN}
\\
\hline
\hline
\multicolumn{4}{|c|}{Batchsize}
&\multicolumn{2}{|c|}{$F_\eta\uparrow$}&\multicolumn{2}{|c|}{$MAE\downarrow$}&\multicolumn{2}{|c|}{$S_\lambda\uparrow$}
&\multicolumn{2}{|c|}{$F_\eta\uparrow$}&\multicolumn{2}{|c|}{$MAE\downarrow$}&\multicolumn{2}{|c|}{$S_\lambda\uparrow$}
\\
\hline
\hline
\multicolumn{4}{|c|}{1}
&\multicolumn{2}{|c|}{0.859}&\multicolumn{2}{|c|}{0.0722}&\multicolumn{2}{|c|}{0.874}
&\multicolumn{2}{|c|}{0.876}&\multicolumn{2}{|c|}{0.0624}&\multicolumn{2}{|c|}{0.879}
\\
\multicolumn{4}{|c|}{2}
&\multicolumn{2}{|c|}{0.863}&\multicolumn{2}{|c|}{0.0701}&\multicolumn{2}{|c|}{0.877}
&\multicolumn{2}{|c|}{0.891}&\multicolumn{2}{|c|}{0.0604}&\multicolumn{2}{|c|}{0.895}
\\
\multicolumn{4}{|c|}{4}
&\multicolumn{2}{|c|}{0.866}&\multicolumn{2}{|c|}{0.0601}&\multicolumn{2}{|c|}{0.876}
&\multicolumn{2}{|c|}{0.882}&\multicolumn{2}{|c|}{0.0583}&\multicolumn{2}{|c|}{0.899}
\\
\multicolumn{4}{|c|}{8}
&\multicolumn{2}{|c|}{0.867}&\multicolumn{2}{|c|}{0.0604}&\multicolumn{2}{|c|}{0.874}
&\multicolumn{2}{|c|}{0.903}&\multicolumn{2}{|c|}{0.0511}&\multicolumn{2}{|c|}{0.910}
\\
\multicolumn{4}{|c|}{12}
&\multicolumn{2}{|c|}{0.866}&\multicolumn{2}{|c|}{0.0607}&\multicolumn{2}{|c|}{0.874}
&\multicolumn{2}{|c|}{0.911}&\multicolumn{2}{|c|}{0.0421}&\multicolumn{2}{|c|}{0.916}
\\
\multicolumn{4}{|c|}{16}
&\multicolumn{2}{|c|}{0.864}&\multicolumn{2}{|c|}{0.0605}&\multicolumn{2}{|c|}{0.876}
&\multicolumn{2}{|c|}{0.911}&\multicolumn{2}{|c|}{0.0420}&\multicolumn{2}{|c|}{0.917}
\\
\multicolumn{4}{|c|}{20}
&\multicolumn{2}{|c|}{0.867}&\multicolumn{2}{|c|}{0.0605}&\multicolumn{2}{|c|}{0.873}
&\multicolumn{2}{|c|}{0.915}&\multicolumn{2}{|c|}{0.0421}&\multicolumn{2}{|c|}{0.910}
\\
\multicolumn{4}{|c|}{24}
&\multicolumn{2}{|c|}{0.870}&\multicolumn{2}{|c|}{0.0606}&\multicolumn{2}{|c|}{0.872}
&\multicolumn{2}{|c|}{0.913}&\multicolumn{2}{|c|}{0.0422}&\multicolumn{2}{|c|}{0.916}
\\
\hline
\hline
\end{tabular}
}
\vspace{-8mm}
\end{center}
\end{table}
\section{Conclusion}
In this work, we propose a novel end-to-end feature learning framework for salient object detection.
Our method devises a symmetrical FCN to learn complementary visual features under the guidance of lossless feature reflection.
The proposed symmetrical FCN architecture contains three branches.
Two of the branches are used to generate the features of the paired reciprocal images, and the other is adopted for hierarchical feature fusion.
For training, we also propose a new weighted structural loss that integrates the location, semantic and contextual information of salient objects.
Thus, the new structural loss function can lead to clear object boundaries and spatially consistent saliency to boost the detection performance.
Extensive experiments on seven large-scale saliency datasets demonstrate that the proposed method achieves significant improvement over the baseline and performs better than other state-of-the-art methods.
\ifCLASSOPTIONcaptionsoff
  \newpage
\fi



\bibliographystyle{IEEEtran}
\bibliography{IEEEabrv,refs}
\ignore{
\begin{IEEEbiography}{}

\end{IEEEbiography}

\begin{IEEEbiographynophoto}{John Doe}

\end{IEEEbiographynophoto}


\begin{IEEEbiographynophoto}{Jane Doe}
Biography text here.
\end{IEEEbiographynophoto}

}



\end{document}